%% file: manuscript.tex
\documentclass[journal]{IEEEtran}
\usepackage{amsmath,amsfonts}
\usepackage{amsthm,cases}
\usepackage{algorithmic}
\usepackage{algorithm,bbding,pifont}
\usepackage{array}
\usepackage{makecell,threeparttable}
\usepackage{textcomp}
\usepackage{stfloats}
\usepackage{url}
\usepackage{color,xcolor}
\usepackage{verbatim}
\usepackage{graphicx}
\usepackage{cite}
\usepackage{multirow}
\input{imZoomConfig.tex}
\newtheorem{lemma}{Lemma}
\newtheorem{definition}{Definition}
\newtheorem{assumption}{Assumption}

\newtheorem{theorem}{Theorem}

\newtheorem{remark}{Remark}

\newcommand{\rr}[1]{\textcolor{red}{#1}}

\newcommand{\rb}[1]{\textcolor{blue}{#1}}
\newcommand{\y}{\color{black}}
\newcommand{\yy}{\color{black}}
\newcommand{\h}{\color{black}}

\makeatletter
\renewcommand{\@thesubfigure}{\hskip\subfiglabelskip}
\makeatother

\begin{document}

\title{Edge-guided Low-light Image Enhancement with Inertial Bregman Alternating Linearized Minimization}
\author{Chaoyan Huang$^\dag$, Zhongming Wu$^\dag$, Tieyong Zeng$^*$
\thanks{This work was supported in part by the National Natural Science Foundation
of China Grants 12471291 and 12001286, the Natural Science Foundation of
Jiangsu Province Grant BK20241899, the China Postdoctoral Science Foundation Grant 2022M711672, the National Key R\&D Program of China under Grant 2021YFE0203700, Grant NSFC/RGC N\_CUHK 415/19, Grant ITF MHP/038/20, Grant RGC 14300219, 14302920, 14301121, and CUHK Direct Grant for Research.}
\thanks{$^\dag$: Contribute equally to this work and are co-first authors.}
\thanks{$^*$: Corresponding author.}
\thanks{Chaoyan Huang and Tieyong Zeng are with the Department of Mathematics, The Chinese University of Hong Kong, Shatin, Hong Kong (e-mail: {cyhuang@math.cuhk.edu.hk}, e-mail: {zeng@math.cuhk.edu.hk}).}
\thanks{Zhongming Wu is with the School of Management Science and Engineering, Nanjing University of Information Science and Technology, Nanjing, China (e-mail: {wuzm@nuist.edu.cn}).}
}
\maketitle

\markboth{Journal of \LaTeX\ Class Files, %~Vol.~14, No.~8, 
January~2024}%
{Shell \MakeLowercase{\textit{et al.}}: A Sample Article Using IEEEtran.cls for IEEE Journals}
% \IEEEpubid{0000--0000/00\$00.00~\copyright~2023 IEEE}

\begin{abstract}

Prior-based methods for low-light image enhancement often face challenges in extracting available prior information from dim images. To overcome this limitation,  we introduce a simple yet effective Retinex model with the proposed edge extraction prior. More specifically, we design an edge extraction network to capture the fine edge features from the low-light image directly. Building upon the Retinex theory, we decompose the low-light image into its illumination and reflectance components and introduce an edge-guided Retinex model for enhancing the low-light image. To solve the proposed model, we propose a novel inertial Bregman alternating linearized minimization algorithm. This algorithm addresses the optimization problem associated with the edge-guided Retinex model, enabling effective enhancement of the low-light image. Through rigorous theoretical analysis, we establish the convergence properties of the algorithm. Besides, we prove that the proposed algorithm converges to a stationary point of the problem through nonconvex optimization theory. Furthermore, extensive experiments are conducted on multiple syntheses and real-world low-light image datasets to demonstrate the efficiency and superiority of the proposed scheme.
\end{abstract}


\section{Introduction}
\IEEEPARstart{D}ue to lighting and environmental conditions, images captured by the camera may be low-light, which negatively impacts post-processing tasks like segmentation, classification, and recognition. Therefore, it is crucial to improve the quality of the low-light image.
Note that the Retinex theory describes the low-light image as the reflectance and the illumination part \cite{land1977retinex}. By enhancing the reflectance and adjusting the {\y illumination} at the same time, the norm-light image can be obtained \cite{li2020luminance}. %,liu2021underexposed}. 
{\y Generally, the regularization term like the edge prior \cite{yuan2012performance} can be utilized to enhance the low-light image. 
Furthermore, {\yy the deep edge extraction module} \cite{xu2023low} can also be applied for low-light image enhancement. 
However, extracting detailed edge information from low-light images is challenging.
As shown in Fig. \ref{Fig1}, one can hardly recognize the content of the observed low-contrast image (a). Besides, the edge structure of (e) in the dark area is not clear. 
To handle this problem, we propose to learn an edge extractor to} {obtain the detailed edge structure directly from the low-light image.
In comparison to the edge produced by the SMG \cite{xu2023low} in (g), the edge restored by our proposed edge extractor more closely resembles the ground-truth (GT) edge (h).  
With the assistance of the proposed edge extractor, we develop a novel Retinex model, which results in our enhancement outcome (c) demonstrating superior performance.

\begin{figure}[t]
\setlength{\abovecaptionskip}{0in}
\subfigcapskip=-0.1in
\hspace{-0.2in}
\subfigure[\tiny{(a) Input (10.74/50.87)
}]{
\zoomincludgraphic{0.83in}{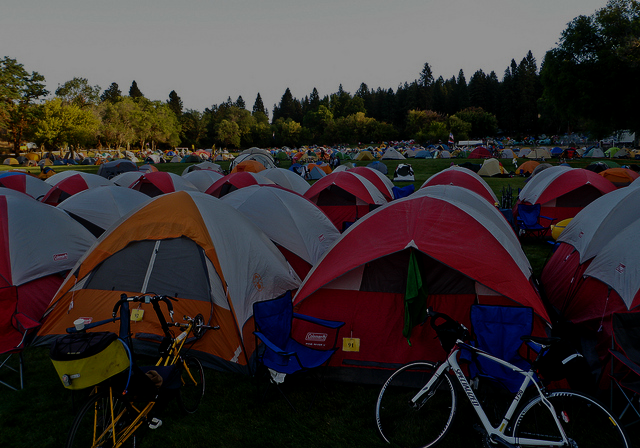}{0.68}{0.06}{0.85}{0.23}{2.5}{help_grid_off}{up_left}{line_connection_off}{2}{orange}{1}{red}}\hspace{-0.28in}
\subfigure[\tiny{(b) SMG (19.55/84.10)}]{
\zoomincludgraphic{0.83in}{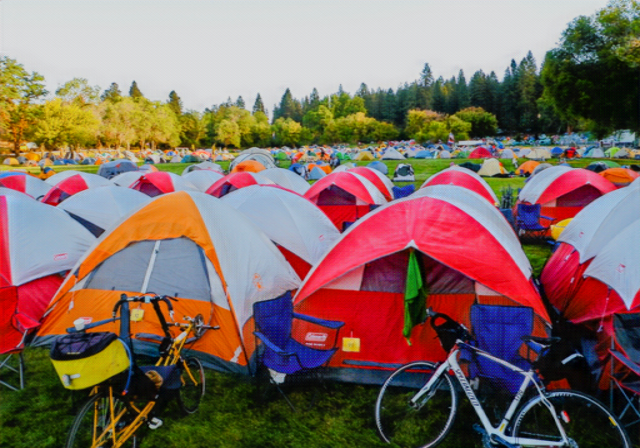}{0.68}{0.06}{0.85}{0.23}{2.5}{help_grid_off}{up_left}{line_connection_off}{2}{orange}{1}{red} }
\hspace{-0.38in}
\subfigure[\tiny{(c) Ours (24.18/92.45)}]{
\zoomincludgraphic{0.83in}{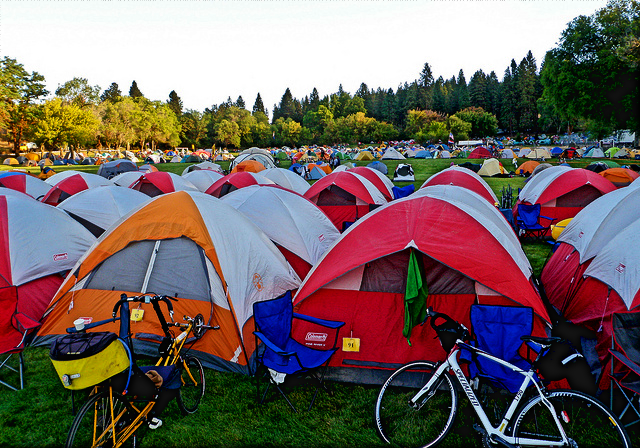}{0.68}{0.06}{0.85}{0.23}{2.5}{help_grid_off}{up_left}{line_connection_off}{2.5}{orange}{1}{red} }
\hspace{-0.38in}
\subfigure[\tiny{(d) GT}]{
\zoomincludgraphic{0.83in}{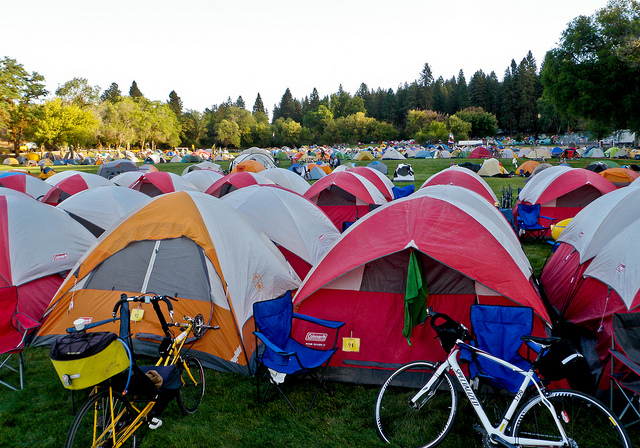}{0.68}{0.06}{0.85}{0.23}{2.5}{help_grid_off}{up_left}{line_connection_off}{2}{orange}{1}{red} }

\hspace{-0.2in}
\subfigure[\tiny{(e) edge of (a)}]{
\zoomincludgraphic{0.83in}{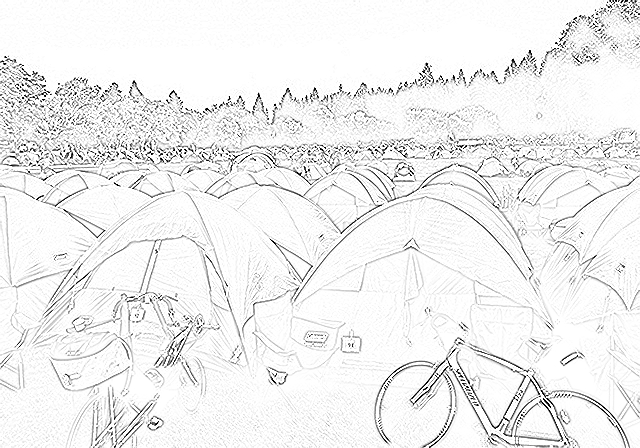}{0.68}{0.06}{0.85}{0.23}{2.5}{help_grid_off}{up_left}{line_connection_off}{2}{orange}{1}{red} }\hspace{-0.33in}
\subfigure[\tiny{(f) edge of SMG
}]{
\zoomincludgraphic{0.83in}{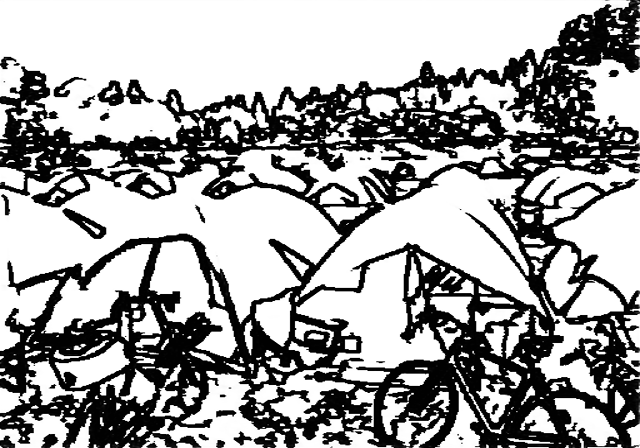}{0.68}{0.06}{0.85}{0.23}{2.5}{help_grid_off}{up_left}{line_connection_off}{2}{orange}{1}{red} }\hspace{-0.33in}
\subfigure[\tiny{(g) edge of Ours
}]{
\zoomincludgraphic{0.83in}{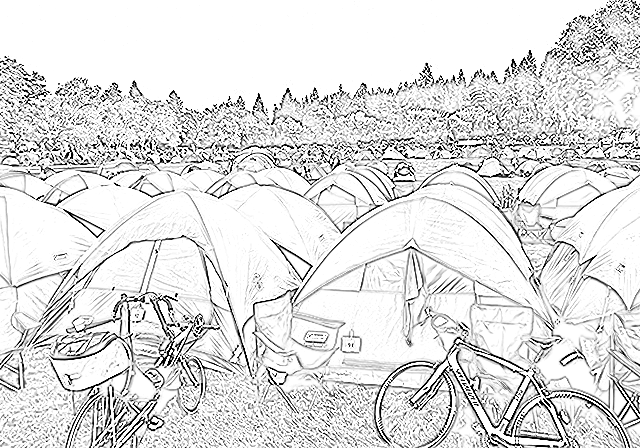}{0.68}{0.06}{0.85}{0.23}{2.5}{help_grid_off}{up_left}{line_connection_off}{2}{orange}{1}{red} }\hspace{-0.33in}
\subfigure[\tiny{(h) GT}]{
\zoomincludgraphic{0.83in}{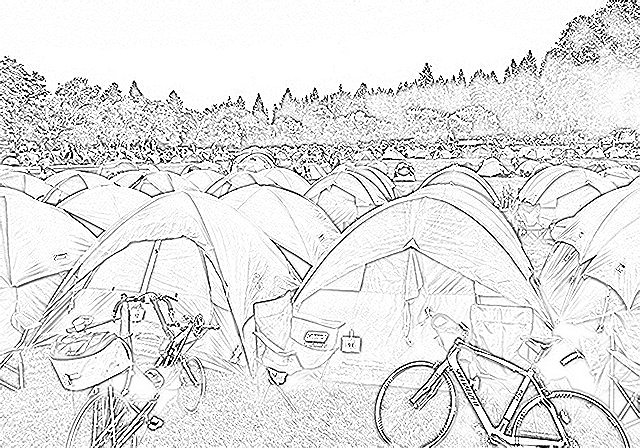}{0.68}{0.06}{0.85}{0.23}{2.5}{help_grid_off}{up_left}{line_connection_off}{2}{orange}{1}{red} }\hspace{-0.3in}
\caption{\y Image enhancement results with index PSNR (dB) and SSIM (\%). (a) is the input low-light image from dataset SYD \cite{lv2021attention}; (b) and (c) are enhancement results by SMG \cite{xu2023low} and ours; (d) is the ground truth; {\yy(e) is the edge of (a) extracted by the Laplace operator;} (f) and (g) are corresponding edges generated by SMG \cite{xu2023low} and ours; (h) is the ground truth.}\label{Fig1}
\end{figure}

Based on the Retinex theory, a low-light image $S$ can be decomposed as the reflectance $R$ and illumination part $L$, i.e., $S=R\circ L$, where $R\in (0,1)$, $L\in (0,\infty)$, {\y and $\circ$ is the element-wise production operator}. Directly obtaining $R$ and $L$ to reconstruct the desired image $\hat S$ is challenged for both $R$ and $L$ are unknown. Therefore, a minimization model has been proposed as follows:
\begin{equation}\label{eq1}
    \min_{R,L} \frac{\beta}{2}\Vert S- R\circ L\Vert^2 + \frac{\alpha}{2}\Phi(R)+\Psi(L),
\end{equation}
where the first term is the data fidelity term,  $\Phi(\cdot)$ and $\Psi(\cdot)$ are the prior regularizers for $R$ and $L$, respectively,  $\beta$ and $\alpha$ are positive parameters. {\y Therefore, we only need to find $R$ and $L$ in the model and hence enhance the intensity of $S$. }

{\y 
For model \eqref{eq1}, many researchers are devoted to estimating a fine reflectance map. 
For example, a dual reflectance estimation method was proposed in \cite{jia2024low} %with convergence analysis 
to improve the performance of the Retinex model. Following this line, the weighted fractional-order total variation and dark channel prior were applied in \cite{ma2023low} to enhance the reflectance part. Wu et al. \cite{wu2023retinex} replaced the traditional prior with a convolutional neural network-based implicit regularizer to restore the reflectance map. Furthermore, Liu et al. \cite{liu2023lowlight} applied two implicit regularizers to adaptively fit both reflectance and illumination priors. A similar idea was presented in \cite{wu2022uretinex}. 
These works have competitive results in enhancing low-light images, however, most of them lack the theoretical guarantee.
}

To better preserve fine details, Gu et al. \cite{gu2019novel} utilized the modified total variational model to fractional order for low-light image enhancement. 
{\y The alternating direction method of multipliers (ADMM) with convergence guarantee was proposed by adding auxiliary variables to solve the model.}
However, when their model faces the non-uniformed light image, the enhanced image may be unsatisfied. To overcome this disadvantage, Ma et al. \cite{ma2022retinex} improved the model with a weighted map without theoretical analysis. 
By extending the fractional-order gradient total variation, small-magnitude details are maintained in the reflectance and the illumination is more likely piecewise smooth. 
Since the edge of an image refers to the curves changing sharply, and gradient can represent that ``change'' in the image \cite{catte1992image}. Most edge detection methods rely on the computation of image gradients. 
However, lower contrast often indicates smaller gradient magnitudes. 
Accordingly, the edge information cannot be obtained directly from the observed image. 
The authors in \cite{9056796} attempted to manipulate the gradient magnitudes of the reflectance to boost the contrast. By multiplying the gradient of the reflectance part with a constant, more clear edge information can be collected. This strategy has also been employed in \cite{li2018structure,wu2023retinex} for edge extraction in the context of low-light image enhancement. 
Drawing from these exemplary studies, it is evident that edge information is essential in low-light image enhancement.

\begin{table}
\centering
\resizebox{.92\hsize}{!}{
\begin{threeparttable}
\caption{\scriptsize SUMMARIZING OF ESSENTIAL COMPARISON ASPECTS OF Retinex-based METHODS\tnote{1}.}
    \label{tab:table}
    \begin{tabular}{cccccc}
    \hline
         Methods  & Fide. & Reg. & Learn. & Alg. &Conv. \\\hline
         \cite{gu2019novel}& \ding{51} & \ding{51} & \ding{55} & \ding{55} & \ding{51} \\
          \cite{ma2022retinex}& \ding{51} & \ding{51} & \ding{55} & \ding{55} & \ding{55} \\
         \cite{ng2011total}& \ding{51} & \ding{51} & \ding{55} & \ding{55} & \ding{51} \\
         \cite{chang2015retinex}& \ding{51} & \ding{51} & \ding{55} & \ding{55} & \ding{55} \\      
         \cite{9056796}& \ding{51} & \ding{51} & \ding{55} & \ding{55} & \ding{55} \\
         \cite{jia2024low}&\ding{51} & \ding{51}& \ding{55} & \ding{55} & \ding{51}\\
         \cite{jia2024variational}&\ding{51} & \ding{51}& \ding{55} & \ding{51} & \ding{51}\\
         \cite{wu2023retinex}& \ding{51} & \ding{51} & \ding{51} & \ding{55} & \ding{55} \\
         \cite{Chen2018Retinex}& \ding{51} & \ding{55} & \ding{51} & \ding{55} & \ding{55} \\
         \cite{liu2023lowlight}& \ding{51} & \ding{51} & \ding{51} & \ding{55} & \ding{55} \\
         \cite{wu2022uretinex}& \ding{51} & \ding{51} & \ding{51} & \ding{55} & \ding{55} \\
         \hline
         Ours& \ding{51} & \ding{51} & \ding{51} & \ding{51} &\ding{51}\\\hline
        \end{tabular}
   \begin{tablenotes}
    \tiny
        \item[1] {`Fide.' means that the data fidelity term holds; `Reg.' represents preserving the regularizer; `Learn.' denotes involving the deep learning-based prior; `Alg.' implies proposing the new algorithm; `Conv.' indicates analyzing the convergence of the algorithm.}
   \end{tablenotes}
\end{threeparttable}}
\end{table}

As previously mentioned, directly extracting edge information from the observed low-light image often yields unsatisfactory results. Deep learning-based methods, on the other hand, have shown exceptional performance in image processing. We are wondering whether these deep learning-based methods can assist us in enhancing the results of the variational Retinex model.
Note that the authors in \cite{fang2020soft} applied the deep learning-based module to generate an edge prediction network with a Laplace loss function to help the image super-resolution task. 
With such an edge prediction network, the edge information of the low-resolution image can be extracted easily. A similar idea was explored in \cite{fang2020multilevel} and \cite{fang2020learning} for image denoising and in \cite{yang2023fast} for MRI reconstruction. In the realm of low-light image enhancement,  the authors in \cite{xu2023low} proposed an end-to-end neural network that incorporates an edge detection module to guide low-light image enhancement. Similarly, an edge enhancement module was introduced within an edge-enhanced multi-exposure fusion network \cite{zhu2020eemefn} to enhance low-light images. In \cite{rana2021edge}, a low-light image enhancement neural network was built with an edge module. 
From these edge-guided deep learning methods, it is apparent that the edge information is significant to image processing. However, the lack of interpretation is still a notable limitation. 

{\y On the other hand, the computational complexity of model \eqref{eq1} is expensive for the product form of two unknown variables. 
}
To tackle this problem, the logarithmic domain can be considered. Let $s=\log S$, $l=\log L$, and $r'=\log R$, then we have $s=l+r'$. Note that $R\in (0,1)$, we let $r=-r'>0$, then $l=s+r$. Hence Eq. \eqref{eq1} can be reformulated as 
\begin{equation}\label{eq2}
    \min_{r,l} \frac{\beta}{2}\Vert  l - s - r\Vert^2 + \frac{\alpha}{2}\Phi(r)+\Psi(l).
\end{equation}
Based on the model in the logarithmic domain, a total variation model was proposed in \cite{ng2011total}. They let $\Psi(l)=\Vert\nabla l\Vert^2+\frac{\mu}{2}\Vert l\Vert^2$ and $\Psi(r) = \Vert\nabla r\Vert_1$ to analyze the convergence of the proposed image enhancement model.  
Later, a dictionary learning-based model was proposed in \cite{chang2015retinex} to improve Eq. \eqref{eq2} by saving more details through sparse representation. While discussing the Retinex model in the logarithmic domain, the edge information still cannot be fully used. Thus, how to get a more clear edge from the low-light image becomes a bottleneck.

Inspired by these works, we aim to explore the deep edge prediction network to directly extract the edge feature from the low-light image and integrate the learned edge extractor into the Retinex model.  
For simplicity, we will focus on the Retinex model in the logarithmic domain. To solve our low-light image enhancement model, we propose a novel inertial Bregman alternating linearized minimization algorithm. The details of the proposed algorithm will be introduced in Section \ref{sec:alg}. For a clear 
perception, we summarize the essential comparison aspects in Table \ref{tab:table}. The contributions of the proposed low-light image enhancement scheme can be summarized as follows. 
\begin{itemize}
    \item We propose an edge extractor for low-light image enhancement, where the latent fine edge structure can be exacted directly from the observed low-light image.
    \item With the help of the latent edge structure, we integrate the proposed edge extractor into the Retinex model in the logarithmic domain for low-light image enhancement. 
    \item To solve the proposed model, an effective inertial Bregman alternating linearized minimization algorithm is proposed. Besides, we establish the convergence properties of the proposed algorithm.
    \item To illustrate the robustness and superiority of the proposed scheme over several state-of-the-art methods, experiments in real-world low-light data are conducted. 
\end{itemize}

The remainder of this paper is organized as follows. Section II reviews the basic
concept of the edge regularizer and some preliminaries for the proposed algorithm. The proposed edge extraction, image enhancement model, and algorithm are displayed in Section III. In Section IV, the convergence properties of the proposed algorithm are established. Experimental results on different datasets are in Section V. Finally, the conclusions follow in Section VI.

\section{Related Works}
In this section, we will briefly recall the edge regularizer and some preliminaries for the proposed algorithm.

\subsection{Edge regularizer}
The restoration of the most detailed information in an image relies on a careful extraction of its edge information \cite{liu2021generalized}. However, for low-contrast images, conventional edge detection often fails to extract satisfactory edges (see Fig. \ref{Fig1}(e)). {\h In this paper, %rather than extracting a binary edge map, 
we aim to derive a deep edge exactor that reflects the local edge strength.}

{\h The traditional variational method to extract edge information involves modifying the image gradient $\nabla S$. Following the strategy in \cite{fang2020soft}, we first compute the normalized gradients
\begin{equation}\label{normalgrad}
    S_x = \frac{\nabla_x S}{\sqrt{1+\left|\nabla S\right|^2}},\quad
    S_y = \frac{\nabla_y S}{\sqrt{1+\left|\nabla S\right|^2}},
\end{equation}
where the $\nabla_x S$ and $\nabla_y S$ are horizontal and vertical gradients of the image $S$. 
Such normalization \eqref{normalgrad} suppresses low contrast variations and enhances prominent edges. %Based on these normalized components, 
The continuous edge indicator function $G$ is defined with standard divergence operation $\operatorname{div}(\cdot)$ as
\begin{equation}\label{eq3}
    G=\operatorname{div}(S_x,S_y).
\end{equation}
%The resulting edge map $G$ has values close to zero in flat regions of $S$ and non-zero at edges, but gradual transitions or textured areas may also yield non-zero responses.
%The continuous values of \(G\) denote the strength of edges across the image to better guide the restoration process. 
The detailed design of our edge extractor can be found in Section \ref{Edge extraction}. 
In discrete form, $G$ is approximated by the finite-difference expression
%The discretization of $G$ is given by
\[
G(i,j)=\frac{S_x(i+1,j)-S_x(i,j)}{\Delta x}+\frac{S_y(i,j+1)-S_y(i,j)}{\Delta y},
\]
where $\Delta x$ and $\Delta y$ are the grid spacings in the $x$- and $y$-directions, respectively.
%The definition of $G(i,j)$ follows the standard finite-difference approximation of the divergence operator. 
%Although \(G\) is computed as the divergence of the normalized gradients, it is not a binary edge map. Instead, the continuous values of \(G\) denote the strength of edges across the image. In practical applications, a thresholding process can be applied to obtain a binary map, but here we exploit the continuous nature of $G$ to better guide the restoration process. }
%
}

{\h While an idealized 45-degree edge (e.g., \( S(x,y) = x + y \)) in a continuous domain would theoretically yield \( G = 0 \) due to uniform normalized gradients, this scenario assumes infinite contrast, perfect linearity, and noise-free conditions, which are rarely met in practice. Real-world edges exhibit finite contrast, discretization effects, and irregularities (e.g., noise or curvature), which disrupt gradient uniformity and ensure non-zero divergence.} %For instance, discretization introduces spatial variations in normalized gradients even for linear edges, while finite contrast amplifies nonlinear dependencies through the normalization in \eqref{normalgrad}. Consequently, \( G \) robustly captures multi-oriented edges, including 45-degree structures, under practical imaging conditions, as validated in our experiments.  

The total variation regularizer, pioneered by Rudin, Osher, and Fatemi \cite{rudin1992nonlinear}, known as the ROF regularizer, was proposed to preserve the edge of an image, which is defined by
\begin{equation}\label{rof}
    \|\nabla S\|_1=\sum_{i, j}\left|(\nabla S)_{i, j}\right|_2=\sum_{i, j}\sqrt{\left(\nabla_x S\right)_{i, j}^2+\left(\nabla_y S\right)_{i, j}^2}.
\end{equation}
This isotropic version of the ROF regularizer is known as the TV regularizer. The anisotropic version, as described in \cite{esedoglu2004decomposition}, is given by
\begin{equation}\label{anis}\hspace{-0.1in}
    \|\nabla S\|_1=\sum_{i, j}\left|(\nabla S)_{i, j}\right|_1=\sum_{i, j}{\vert\left(\nabla_x S\right)_{i, j}\vert+\vert\left(\nabla_y S\right)_{i, j}}\vert.
\end{equation}
There are many variants of the ROF model to represent the edge of the image, such as the change of norm, weighted gradient, generalized variation, ROF model in different spaces, etc. More specifically, the ROF model was used in \cite{feng2021local} as an additional regular term to obtain the detailed texture in hyperspectral anomaly detection. The ROF model with $\ell_0$-norm $\Vert \nabla S \Vert_0$ was utilized to enforce the sparsity of image gradient to help the image restoration in \cite{lv2021blind}. More structural information was collected by applying a weighted ROF regularizer $\Vert T\nabla S \Vert_1$ for image segmentation \cite{wu2021adaptive}. Total
generalized variation (TGV) contributed to help maintain sharp edges in \cite{wu2023vdip}. Total variation in HSV color space was applied to help image enhancement \cite{wang2023saturation}. Nonlocal total variation was applied to data speckle reduction in \cite{nie2016nonlocal}. 

{\y Based on the aforementioned literature, edge information is beneficial for a variety of tasks. In this paper, we employ a convolutional neural network to learn the edge extractor, aiding in the image enhancement task.
}

\begin{figure*}
    \centering
    \includegraphics[width=7in]{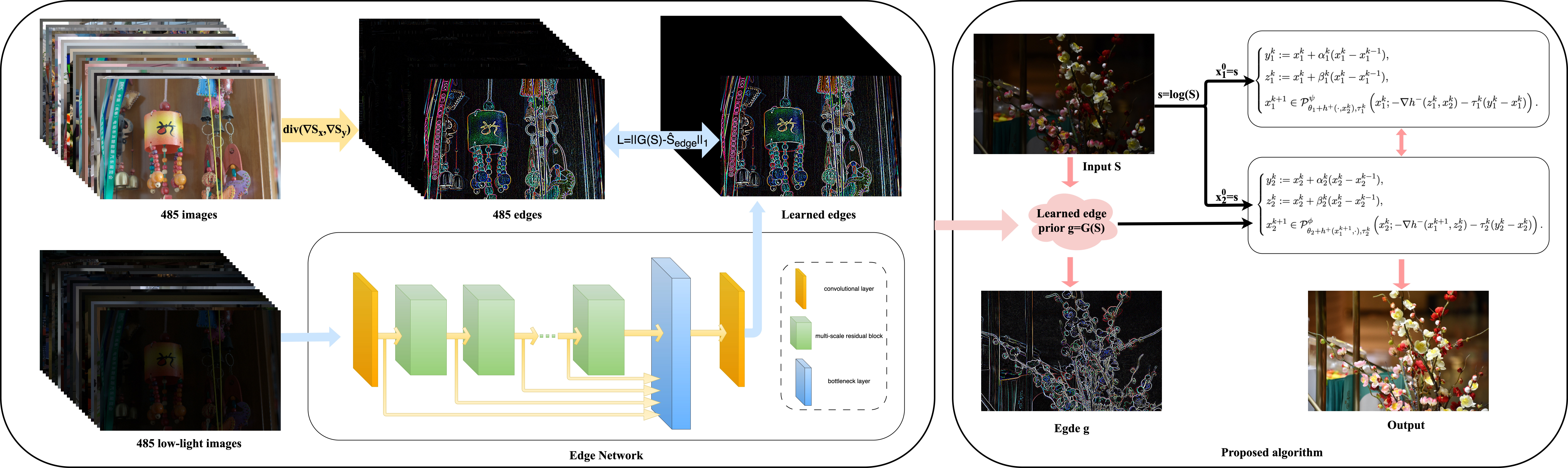}
    \caption{The flowchart of the proposed scheme. We use 485 paired images from the LOL dataset as the training set. The edge information can be directly obtained by using the proposed Edge-Net. 
    The yellow, green, and blue blocks are the convolution layer, multi-scale residual block, and bottleneck layer, respectively. {\y Then we use the learned edge extractor to help with the image enhancement task.} Through the proposed convergent framework, the low-light image can be well enhanced.}
    \label{fig:retinex}
\end{figure*}

\subsection{Preliminaries}\label{sec:alg}
In this subsection, we review some preliminaries about subdifferential and Kurdyka-{\L}ojasiewicz (KL) property in nonconvex optimization, which will be used in our algorithm design and convergence analysis.

\begin{definition} \label{def2.1}{\rm\cite{attouch2013convergence, bolte2014alternating}} \rm(Subdifferentials)
Let $f:\mathbb {R}^{n}\rightarrow (-\infty,+\infty]$ be a proper and lower semicontinuous function.
\begin{itemize}
\item[(i)] For a given $x\in {\rm dom} f$, the Fr\'{e}chet subdifferential of $f$ at $x$, written by $\widehat{\partial}f(x)$, is the set of all vectors $u\in \mathbb{R}^n$ satisfying
$$\liminf_{y\neq x,y\rightarrow x}\frac{f(y)-f(x)-\langle u,y-x\rangle}{\|y-x\|}\geq0,$$
and we set $\widehat{\partial}f(x) = \emptyset$ when $x\notin {\rm dom}f$.

\item[(ii)] The limiting-subdifferential, or simply the subdifferential, of $f$ at $x$, written by $\partial f(x)$, is defined by
\begin{equation}\label{pf}
\begin{aligned}
\partial f(x):=\{u\in\mathbb{R}^n\; | \; \exists \ x^k\rightarrow x, ~{\rm s.t.}~f(x^k)\rightarrow f(x)\\
 ~{\rm and}~ \widehat{\partial}f(x^k) \ni u^k \rightarrow u \}. 
\end{aligned}
\end{equation}

\item[(iii)] A point $x^*$ is called (limiting-)critical point or stationary point of $f$ if it satisfies $0\in\partial f(x^*)$, and the set of critical points of $f$ is denoted by ${\rm crit} f$.
\end{itemize}
\end{definition}

We use the notation ${\rm dom} (\partial f):=\{x\in\mathbb{R}^n\; | \;  \partial f(x) \neq \emptyset\}$. 
If $g$ is a continuously differentiable function, it holds that $\partial(f+g)=\partial f+\nabla g$. 

Next, we recall the KL property \cite{attouch2010proximal, bolte2014alternating, kurdyka1998}, which plays a central role in the convergence analysis.

\begin{definition}\label{def2.2}(KL property and KL function)
Let $f:\mathbb{R}^{n}\rightarrow (-\infty,+\infty]$ be a proper and lower semicontinuous function.
\begin{itemize}
\item[$(a)$]    $f$ is said to satisfy KL property at $x^*\in{\rm dom}(\partial f)$ if there exist a constant $\eta\in(0,+\infty]$, a continuous concave function $\phi:[0,\eta)\rightarrow \mathbb{R}_+$ and a neighborhood $U$ of $x^*$, such that
\begin{itemize}
\item[\rm(i)] $\phi(0)=0$ and $\phi$ is continuously differentiable on $(0,\eta)$ with $\phi'>0$;

\item[\rm(ii)] for all $x\in U\cap\{z \in \mathbb{R}^n\; | \; f(x^*) < f(z) < f(x^*)+\eta\}$, the following inequality holds:
\begin{equation} \label{kl}
\phi'(f(x)-f(x^*)){\rm dist}(0,\partial f(x))\geq1.
\end{equation}
\end{itemize}

\item[$(b)$]  $f$ is called a KL function if $f$ satisfies the KL property at each point of dom$(\partial f)$.
\end{itemize}
\end{definition}

Denote $\Phi_{\eta}$ the set of functions $\phi$ which satisfy the involved conditions in Definition \ref{def2.2}(a). 

\begin{definition}
(Bregman distance) Let $\psi(\cdot): \mathbb{R}^n \rightarrow(-\infty,+\infty]$ be a strongly convex and continuously differentiable function over $\rm{dom}(\partial \psi)$. The Bregman distance associated with the kernel $\psi(\cdot)$ is the function $D_\psi(\cdot, \cdot): \rm{dom}(\psi) \times \rm{dom}(\partial \psi) \rightarrow \mathbb{R}$ given by
$$
\begin{aligned}
D_\psi(x, y)=\psi(x)-\psi(y)-\langle\nabla \psi(y), x-y\rangle, \\
\qquad\qquad\forall x \in \operatorname{dom}(\psi), y \in \operatorname{int}(\operatorname{dom}(\psi)) .
\end{aligned}
$$
\end{definition}

Note that the Bregman distance includes the standard Euclidean distance as its special case, i.e., $D_\psi(x, y)=\frac{1}{2}\|x-y\|^2$ if the kernel function is chosen by
$\psi(\cdot)=\frac{1}{2}\|\cdot\|^2$.
We refer to \cite{ahookhosh2021bregman,bolte2018first,cai2022developments,he2022unified,wu2021inertial} and references therein for the Bregman distance.

\section{Edge-guided model with algorithm}
\subsection{Edge extraction}\label{Edge extraction}
Different from the traditional edge extraction methods, in this paper, we propose to obtain the edge from a simple but efficiently designed neural network. More specifically, we use the multi-scale residual block (MSRB) \cite{li2018multi} as the backbone of the proposed edge network. Due to the bypass network structure, the MERB can detect more features of images and the information of different passes can be shared with each other. Giving the observed image $S$ and the corresponding label $\widehat{S}$, the proposed edge detection network is trained. 
The loss function of the proposed network is 
\begin{equation}
    \mathcal{L}=\Vert G(S)-\widehat{S}_{edge}\Vert_1,
\end{equation}
where $G(\cdot)$ represents the proposed edge exactor, $G(S)$ denotes the {\y learning-based} edge, and $\widehat{S}_{edge}$ is the edge label obtained from Eq. \eqref{eq3}. {\h Note that the learned image edge $G(S)$ has the same dimension as the input image $S$.} As shown in Fig.\ref{fig:retinex}, we capture labels from the corresponding bright-light images.
By minimizing the loss function, we train the edge network that can generate the edge directly from the low-light image. The detailed training setting is described in Section \ref{experiment}.

\subsection{The proposed Retinex image enhancement scheme}
Following the setting of Eq. \eqref{eq2} in the logarithmic domain, we propose to minimize the following energy functional based on Retinex theory for low-light image enhancement: 
\begin{equation}\label{model}
E(r, l)= \frac{\beta}{2}\Vert l - s -r \Vert^2+\frac{\alpha}{2}\Phi(r,g)+\Psi(l), 
%E(r, l)= \Vert\nabla l\Vert_1 + \frac{\alpha}{2}\Vert \nabla r - \nabla g\Vert^2 + \frac{\beta}{2}\Vert l - s -r \Vert^2, 
\end{equation}
where $g=G(S)$ is the learned edge extractor, $\Phi(r,g)=\Vert \nabla r - \nabla g\Vert^2$, $\Psi(l)$ is a regularizer, $\alpha$ and $\beta$ are the positive parameters. 
To solve the proposed model \eqref{model}, we consider a more general class of structured optimization problems that minimize the following function: 
{\small
\begin{equation}\label{problem}
\Phi(x_1, x_2):=\theta_1(x_1)+\underbrace{h^{+}(x_1, x_2)-h^{-}(x_1, x_2)}_{h(x_1, x_2)}+\theta_2(x_2),
\end{equation}}where $x_1$ and $x_2$ are decision variables, $\theta_1$ and $\theta_2$ are assumed to be general nonsmooth and nonconvex functions, and $h(x_1, x_2)$ is continuously differentiable for both $x_1$ and $x_2$ component-wise, which is also assumed to be decomposable into two parts $h^{+}(x_1, x_2)$ and $h^{-}(x_1, x_2)$. Define
\begin{equation}\label{bprof}
\mathcal{P}_{g,\mu}^\psi(u;v):= \arg\min_{\hskip -0.4cm x\in\mathbb{R}^n}\left\{g(x)+\langle x,v\rangle+\mu D_\psi(x,u)\right\}
\end{equation}
 with certain strongly convex function $\psi$ and the Bregman divergence $D_\psi(\cdot,\cdot)$.

We propose an inertial Bregman alternating linearized minimization algorithm to solve the general optimization problem that minimizes the function given in \eqref{problem}. The algorithm framework is summarized in Algorithm \ref{algo1}, where the kernel functions $\psi_k$ and $\varphi_k$, for all $k$, are selected to be strongly convex with the smallest moduli $\rho_1$ and $\rho_2$, respectively.

{
\begin{remark}
Compared with existing methods, Algorithm \ref{algo1} is novel for solving problem \eqref{problem}, which integrates the general inertial acceleration technique with Bregman regularization. The most closely related works include \cite{gao2020gauss, guo2023two, mukkamala2020convex, he2022unified, pock2016inertial}. However, the inertial technique is not incorporated in \cite{he2022unified}, and the methods proposed in \cite{gao2020gauss, guo2023two, mukkamala2020convex, he2022unified, pock2016inertial} are tailored to tackle particular instances of problem \eqref{problem}.
\end{remark}}

Next, we apply the proposed inertial Bregman alternating minimization algorithm, i.e., Algorithm \ref{algo1}, to solve \eqref{model}. Note that the model 
\eqref{model} falls into the form of \eqref{problem} with $(x_1,x_2):=(l,r)$, $\theta_1(x_1):=\Psi(l)$, $\theta_2(x_2):=\frac{\alpha}{2}\Phi(r,g)$, $h^{+}(x_1, x_2):=0$, and $h^{-}(x_1, x_2):=-\frac{\beta}{2}\Vert l - s -r \Vert^2$. By choosing $\psi_k\equiv\psi:=\frac{1}{2}\|\cdot\|^2$ and $\varphi_k:=\frac{1}{2}\|\cdot\|_{M_k}^2$ with $M_k:=\gamma_k I -\frac{\alpha}{\tau_2^k}\nabla^\top\nabla$, where $\gamma_k>\frac{\alpha}{\tau_2^k}\lambda_{\max}(\nabla^\top\nabla)$, % and applying Algorithm \ref{algo1} to solve \eqref{model}, 
we obtain the following $l$- and $r$-subproblems:
\begin{equation}\label{l-subproblem}
\begin{aligned}
    &l^{k+1}=\underset{l}{\arg\min}~\Big\{\Psi(l)+\frac{\tau_1^k}{2}\Vert l-l^k\Vert^2
    \\&\qquad\qquad +\langle l,\beta(r^k-z_1^k-s)-\tau_1^k(y_1^k-l^k)\rangle\Big\},
    \end{aligned}
\end{equation}
 % and
 \begin{equation}\label{r-subproblem}
\begin{aligned}   
    &r^{k+1}= \underset{r}{\arg\min}~\Big\{\frac{\alpha}{2}\Vert\nabla r-\nabla g\Vert^2+\frac{\tau_2^k}{2}\Vert r-r^k\Vert_{M_k}^2\\&\qquad\qquad +\langle r,\beta(l^{k+1}-z_2^k+s)-\tau_2^k(y_2^k-r^k)\rangle\Big\}.
\end{aligned}
\end{equation}

Notice that the $l$-subproblem \eqref{l-subproblem} is a convex optimization problem that can be efficiently solved by classic ADMM algorithm, while the $r$-suproblem \eqref{r-subproblem} possesses the closed-form solution according to the definition of $M_k$ as follows:
{\small
\begin{equation}\label{r-solution}\hspace{-0.2in}
r^{k+1} =\frac{\alpha \nabla^T\nabla g+\beta(z_2^k-l^{k+1}-s)+\tau_2^k(y_2^k+M_kr^k-r^k)}{\gamma^k}.
\end{equation}}
Therefore, the detailed iterative scheme of Algorithm \ref{algo1} for solving the proposed model can be read as follows: 
\begin{equation}\label{modelalgorithm}
  \left\{
   \begin{aligned}
y_1^k = l^k+\alpha_1^k(l^k-l^{k-1}), \\
z_1^k = l^k+\beta_1^k(l^k-l^{k-1}),\\
{\rm Compute}~ l^{k+1} ~{\rm by}~ \eqref{l-subproblem},
% \left\{
%    \begin{aligned}
%    w^{k+1} &= \mathcal{S}_\frac{1}{\eta}(\nabla x_1^{k+1}+\frac{\xi^k}{\eta}),\\
%    x_1^{k+1} &= \frac{\tau_1^ky_1^k+\beta(s+z_1^k-x_2^k)+\eta\nabla^T (w^k - \frac{\xi^k}{\eta})}{\tau_1^k+\eta\nabla^T\nabla},\\
% \xi^{k+1} &= \xi^k+\nabla x_2^{k+1}-w^{k+1}, 
%    \end{aligned}\right.
   \\
y_2^k = r^k+\alpha_2^k(r^k-r^{k-1}), \\
z_2^k = r^k+\beta_2^k(r^k-r^{k-1}),\\
{\rm Compute}~ r^{k+1} ~{\rm by}~ \eqref{r-solution}.
\end{aligned}\right.
  \end{equation}

In the implementation, the parameter $\gamma_k$ is not difficult to choose since the maximum eigenvalue $\lambda_{\max}(\nabla^\top\nabla)\leq 8$ \cite{chambolle2004algorithm}.
{\y Drawing from \cite{ng2011total}, the image produced by the Retinex model usually tends to be over-enhanced. Consequently, a Gamma correction is incorporated into $\hat{L} = \exp(l^{k+1})$. Thus, the final result is given by $\hat{S} = \hat{L}^{\frac{1}{2.2}} \circ \hat{R}$, where $\hat{R} = \exp(-r^{k+1})$. 
}

\begin{algorithm}[t!]
\caption{Inertial Bregman alternating linearized minimization algorithm}
\label{algo1}
\begin{algorithmic}
\STATE{Choose the parameters $\alpha_i^k$, $\beta_i^k$ and $\tau_i^k$, $i=1,2$. For given $x^0\in\mathbb{R}^n$, $x^{-1}:=x^0$, and $k:=0$.}
\WHILE{the stopping criterion is not satisfied,}
\STATE{
1. Compute $y_1^k,z_1^k$ and $x_1^{k+1}$ by
\begin{subequations}
\begin{numcases}{}
y_1^k:= x_1^k+\alpha_1^k(x_1^k-x_1^{k-1}),\label{ama1:1}\\ 
z_1^k:=x_1^k+\beta_1^k(x_1^k-x_1^{k-1}),\label{ama1:2}\\
x_1^{k+1}\in {\bf\mathcal{P}}_{\theta_1+h^+(\cdot,x_2^k),\tau_1^k}^{\psi_k}\left(x_1^k;t_1^k\right),\label{ama1:3}
\end{numcases}
\end{subequations}
with $t_1^k=-\nabla h^-(z_1^k,x_2^k)-\tau_1^k(y_1^k-x_1^k)$.

2. Compute $y_2^k,z_2^k$ and $x_2^{k+1}$ by
\begin{subequations}\begin{numcases}{} 
y_2^k:= x_2^k+\alpha_2^k(x_2^k-x_2^{k-1}),\label{ama2:1}\\[0.2cm]
z_2^k:=x_2^k+\beta_2^k(x_2^k-x_2^{k-1}),\label{ama2:2}\\[0.2cm]
x_2^{k+1}\in {\bf\mathcal{P}}_{\theta_2+h^+(x_1^{k+1},\cdot),\tau_2^k}^{\varphi_k}\left(x_2^k;t_2^k\right),\label{ama2:3}
 \end{numcases}
\end{subequations}
with $t_2^k=-\nabla h^-(x_1^{k+1},z_2^k)-\tau_2^k(y_2^k-x_2^k)$.
}
\ENDWHILE
\end{algorithmic}
\end{algorithm}

\section{Convergence results}
Now we establish the convergence of Algorithm \ref{algo1}. We start by giving some standard assumptions for \eqref{problem}.
\begin{assumption}\label{ass1}
 The components $\theta_i (i=1,2)$ are proper lower semicontinuous functions.   
  The objective function $ \Phi(x_1, x_2)$ is bounded from below for all $x_1,x_2$.
\end{assumption} 
\begin{assumption}\label{ass2}
The function $h(x_1, x_2)=h^{+}(x_1, x_2)-h^{-}(x_1, x_2)$ satisfies the following conditions.
\begin{itemize}
\item[(i)] For any fixed $x_2$, the function $h^{-}(\cdot, x_2)$ is $C_{L_1^{-}(x_2)}^{1,1}$, i.e., the partial gradient $\nabla_{x_1} h^{-}(\cdot, x_2)$ is globally Lipschitz with module $L_1^{-}(x_2)$, that is for $\forall u, v \in \mathbb{R}^n$,
{\small
$$
\left\|\nabla_{x_1} h^{-}\left(u, x_2\right)-\nabla_{x_1} h^{-}\left(v, x_2\right)\right\| \leq L_1^{-}(x_2)\left\|u-v\right\|.
$$}Likewise, for any fixed $x_1$, the function $h^{-}(x_1, \cdot)$ is assumed to be $C_{L_2^{-}(x_1)}^{1,1}$.
\item[(ii)] For $i=1,2$, there exist $\lambda_i^{-}$ and $\lambda_i^{+}>0$ such that
$\inf \left\{L_1^{-}\left(x_2^k\right): k \in \mathbb{N}\right\} \geq \lambda_1^{-}$,
$ \inf \left\{L_2^{-}\left(x_1^k\right): k \in \mathbb{N}\right\} \geq \lambda_2^{-}$,
$\sup \left\{L_1^{-}\left(x_2^k\right): k \in \mathbb{N}\right\} \leq \lambda_1^{+}$,
and $\sup \left\{L_2^{-}\left(x_1^k\right): k \in \mathbb{N}\right\} \leq \lambda_2^{+}.$
\item[(iii)] $h $ is a continuously differentiable function and its gradient $\nabla h(x_1, x_2)=\nabla h^{+}(x_1, x_2)-\nabla h^{-}(x_1, x_2)$ is Lipschitz continuous on bounded subsets of $\mathbb{R}^n \times \mathbb{R}^m$. That is, for each bounded subsets $\mathbb{B}_1 \times \mathbb{B}_2$ of $\mathbb{R}^n \times \mathbb{R}^m$, there exists $L>0$ (or $L^{+}$and $L^{-}$satisfying $L=L^{+}+L^{-}$) such that
$$
\begin{aligned}
&\left\|\left(\begin{array}{l}
\nabla_{x_1} h\left(x_1, y_1\right) \\
\nabla_{x_2} h\left(x_1, y_1\right)
\end{array}\right)-\left(\begin{array}{c}
\nabla_{x_1} h\left(x_2, y_2\right) \\
\nabla_{x_2} h\left(x_2, y_2\right)
\end{array}\right)\right\| \\
&\leq L\left\|\left(\begin{array}{l}
x_1 \\
y_1
\end{array}\right)-\left(\begin{array}{l}
x_2 \\
y_2
\end{array}\right)\right\|
\end{aligned}
$$
for all $\left(x_i, y_i\right) \in \mathbb{B}_1 \times \mathbb{B}_2~(i=1,2)$.
\end{itemize}\end{assumption}

\begin{assumption}\label{ass3}
The parameters $\{\alpha_i^k\} $, $\{\beta_i^k\} $, and $\{\tau_i^k\} $, $i=1,2$, for all $k\in\mathbb{N}$ should satisfy the following conditions:
\begin{itemize}
\item[(i)]$\exists\ 0 < \bar\alpha_i < \frac{(1-\varepsilon)\rho_i}{2}$ such that $\alpha_i^k\in[0,\bar\alpha_i]$, $i=1,2$. In addition, $\beta_i^k\in[0,\bar\beta_i]$ for some $\bar\beta_i>0$, $i=1,2$.
\item[(ii)] it holds that
\begin{equation}\label{tauk}
\begin{aligned}
&\tau_1^k=\frac{(1+\varepsilon)\delta_1+(1+\beta_1^k)L_1^-(x_2^k)}{\rho_1-\alpha_1^k},\\
&\tau_2^k=\frac{(1+\varepsilon)\delta_2+(1+\beta_2^k)L_2^-(x_1^{k+1})}{\rho_2-\alpha_2^k},  
\end{aligned}
\end{equation}
where
\begin{equation}\label{delta}\hspace{-0.25in}
\delta_1=\frac{\bar\alpha_1+\bar\beta_1\rho_1}{(1-\varepsilon)\rho_1-2\bar\alpha_1}\lambda_1^+,~~
\delta_2=\frac{\bar\alpha_2+\bar\beta_2\rho_2}{(1-\varepsilon)\rho_2-2\bar\alpha_2} \lambda_2^+. 
\end{equation}
\end{itemize}
\end{assumption}

\begin{remark}
According to Assumption \ref{ass3}, for all $k\in\mathbb{N}$, $\tau_i^k,i=1,2$ is bounded from above, that is, $\tau_i^k\leq \tau_i^+, i=1,2,$ with 
\begin{equation}\begin{aligned}
  &\tau_1^+=\frac{(1+\varepsilon)\delta_1+(1+\bar\beta_1)\lambda_1^+}{\rho_1-\bar\alpha_1},\\
&\tau_2^+=\frac{(1+\varepsilon)\delta_2+(1+\bar\beta_2)\lambda_2^+}{\rho_2-\bar\alpha_2}.   
\end{aligned}\end{equation}
\end{remark}

Before the convergence analysis, for notational simplicity, we define an auxiliary function that enjoys the property of yielding a sequence of decreasing function values. Let $\Theta: \mathbb{R}^{n_1 \times n_2} \times \mathbb{R}^{n_1 \times n_2} \rightarrow(-\infty, \infty]$ be the auxiliary function 
\begin{equation}\label{Theta}\hspace{-0.1in}
 \Theta_{\delta_1, \delta_2}(\mathbf{w}):=\Phi\left(w_1\right)+\frac{\delta_1}{2}\left\|w_{11}-w_{21}\right\|^2+\frac{\delta_2}{2}\left\|w_{12}-w_{22}\right\|^2,
\end{equation}
where $\Phi$ is defined in \eqref{problem}, 
$ w_1=\left(w_{11}, w_{12}\right) \in \mathbb{R}^{n_1} \times \mathbb{R}^{n_2}, w_2=\left(w_{21}, w_{22}\right) \in \mathbb{R}^{n_1} \times \mathbb{R}^{n_2}$, and $\mathbf{w}=\left(w_1, w_2\right)$.

\begin{lemma}\label{lemma1}
 Let $\left\{\left(x_1^k, x_2^k\right)\right\}_{k \in \mathbb{N}}$ be a sequence generated by Algorithm \ref{algo1}, for all $k \in \mathbb{N}$, $w_1^k=$ $\left(x_1^k, x_2^k\right), w_2^k=\left(x_1^{k-1}, x_2^{k-1}\right)$, and $\mathbf{w}^k=\left(w_1^k, w_2^k\right)$. Suppose that Assumptions \ref{ass1}, \ref{ass2} and \ref{ass3} hold. 
 Then, the sequence $\{\Theta_{\delta_1,\delta_2}({\bf w}^k)\}_{k\in\mathbb{N}}$ is nonincreasing. In particular, for all $k\in\mathbb{N}$ and $\varepsilon>0$, it holds that
\begin{equation}\label{descent}
\Theta_{\delta_1,\delta_2}\left(\mathbf{w}^{k+1}\right)-\Theta_{\delta_1,\delta_2}\left(\mathbf{w}^k\right)
\leq-\delta\left\|\mathbf{w}^{k+1}-\mathbf{w}^k\right\|^2,   
\end{equation}
where
$\delta_1,~\delta_2$ are defined in \eqref{delta},
and $\delta = \frac{\varepsilon}{2}\min\{\delta_1,\delta_2\}>0$.
 \end{lemma}
%The proof of Lemma \ref{lemma1} can be found in Appendix \ref{Plemma1}. 
%\section{Proof of Lemma \ref{lemma1}}\label{Plemma1}

To prove Lemma \ref{lemma1}, we first recall the descent property of the smooth function in the following lemma.

\begin{lemma}\label{Lem2.2}\hspace{-0.1cm}{\rm\cite{bolte2014alternating}}
 Let $f:~\mathbb{R}^{n}\rightarrow \mathbb{R}$ be a continuously differentiable function with gradient $\nabla f$ assumed $L_f$-Lipschitz continuous. Then, for any $u,v\in \mathbb{R}^{n}$, we have
 \begin{equation}
 \left|f(u)-f(v)-\langle u-v,\nabla f(v)\rangle\right|\leq \frac{L_f}{2}\|u-v\|^2.
 \end{equation}
 \end{lemma}

In the following, we prove Lemma \ref{lemma1}.
\begin{proof}
Using the updating schemes of $x_1^{k+1}$ and $x_2^{k+1}$ in \eqref{ama1:3} and \eqref{ama2:3}, respectively, we have
$$
\begin{aligned}
&\theta_1\left(x_1^{k+1}\right)+h^{+}\left(x_1^{k+1}, x_2^k\right)+\tau_1^k D_{\psi_k}\left(x_1^{k+1}, x_1^k\right) \\
&-\left\langle x_1^{k+1}\right. \left.-x_1^k, \nabla_{x_1} h^{-}\left(z_1^k, x_2^k\right)+\tau_1^k\left(y_1^k-x_1^k\right)\right\rangle \\ & \hspace{-0.2in} \leq \theta_1\left(x_1^k\right)+h^{+}\left(x_1^k, x_2^k\right),
\end{aligned}
$$
and
$$
\begin{aligned}
&\theta_2\left(x_2^{k+1}\right)+h^{+}\left(x_1^{k+1}, x_2^{k+1}\right) +\tau_2^k D_{{\varphi_k}}\left(x_2^{k+1}, x_2^k\right) \\&
-\left\langle x_2^{k+1}-x_2^k, \nabla_{x_2} h^{-}\left(x_1^{k+1}, z_2^k\right)+\tau_2^k\left(y_2^k-x_2^k\right)\right\rangle \\ & \hspace{-0.2in} \leq \theta_2\left(x_2^k\right)+h^{+}\left(x_1^{k+1}, x_2^k\right) .
\end{aligned}
$$
Summing both inequalities and using the definition of $\Phi(\cdot,\cdot)$ in \eqref{problem} yield
\begin{equation}\label{proof_eq1}
\begin{aligned}
&\Phi\left(x_1^{k+1},x_2^{k+1}\right)\\ \leq &\Phi\left(x_1^k,x_2^k\right) -h^-\left(x_1^{k+1},x_2^{k+1}\right)+h^-\left(x_1^k,x_2^k\right)\\
&+\left\langle x_1^{k+1}\right. \left.-x_1^k, \nabla_{x_1} h^{-}\left(z_1^k, x_2^k\right)+\tau_1^k\left(y_1^k-x_1^k\right)\right\rangle 
\\
&+\left\langle x_2^{k+1}-x_2^k, \nabla_{x_2} h^{-}\left(x_1^{k+1}, z_2^k\right)+\tau_2^k\left(y_2^k-x_2^k\right)\right\rangle \\&
-\tau_1^kD_{\psi_k}\left(x_1^{k+1}, x_1^k\right)
-\tau_2^kD_{{\varphi_k}}\left(x_2^{k+1}, x_2^k\right).
\end{aligned}    %\nonumber
\end{equation}
%According to Assumption \ref{ass2}(i), it follows from Lemma \ref{Lem2.2} that
From Assumption \ref{ass2}(i) and Lemma \ref{Lem2.2}, we know that
$$
\begin{aligned}
-h^{-}\left(x_1^{k+1}, x_2^k\right) \leq&\frac{L_1^{-}\left(x_2^k\right)}{2}\left\|x_1^{k+1}-x_1^k\right\|^2-h^{-}\left(x_1^k, x_2^k\right)\\&-\left\langle x_1^{k+1}-x_1^k, \nabla_{x_1} h^{-}\left(x_1^k, x_2^k\right)\right\rangle,
\end{aligned}
$$
where $L_1^{-}\left(x_2^k\right)$ is the Lipschitz constant of $\nabla_{x_1} h^{-}\left(x_1, x_2^k\right)$ at $x_1^k$. Similarly, we have
$$\begin{aligned}
&-h^{-}\left(x_1^{k+1}, x_2^{k+1}\right)\\\leq&\frac{L_2^{-}\left(x_1^{k+1}\right)}{2}\left\|x_2^{k+1}-x_2^k\right\|^2-h^{-}\left(x_1^{k+1}, x_2^k\right)\\&-\left\langle x_2^{k+1}-x_2^k, \nabla_{x_2} h^{-}\left(x_1^{k+1}, x_2^k\right)\right\rangle,
\end{aligned}$$
where $L_2^{-}\left(x_1^{k+1}\right)$ is the Lipschitz constant of $\nabla_{x_2} h^{-}\left(x_1^{k+1}, x_2\right)$ at $x_2^k$.
Summing the above two inequalities and plugging into \eqref{proof_eq1}, we get 
\begin{equation}\label{proof_eq2}
\begin{aligned}
& \Phi\left(x_1^{k+1}, x_2^{k+1}\right) \\
\leq & \Phi\left(x_1^k, x_2^k\right)-\tau_1^k D_{\psi_k}\left(x_1^{k+1}, x_1^k\right)-\tau_2^k D_{{\varphi_k}}\left(x_2^{k+1}, x_2^k\right) \\
& +\left\langle x_1^{k+1}-x_1^k, \nabla_{x_1} h^{-}\left(z_1^k, x_2^k\right)-\nabla_{x_1} h^{-}\left(x_1^k, x_2^k\right)\right\rangle \\
& +\left\langle x_2^{k+1}-x_2^k, \nabla_{x_2} h^{-}\left(x_1^{k+1}, z_2^k\right)-\nabla_{x_2} h^{-}\left(x_1^{k+1}, x_2^k\right)\right\rangle \\
& +\tau_1^k\left(y_1^k-x_1^k\right)+\tau_2^k\left(y_2^k-x_2^k\right) \\
& +\frac{L_1^{-}\left(x_2^k\right)}{2}\left\|x_1^{k+1}-x_1^k\right\|^2+\frac{L_2^{-}\left(x_1^{k+1}\right)}{2}\left\|x_2^{k+1}-x_2^k\right\|^2 .
\end{aligned}
\end{equation}
Note that 
\begin{equation} \begin{aligned}
&\left\langle x_1^{k+1}\right. \left.-x_1^k, \nabla_{x_1} h^{-}\left(z_1^k, x_2^k\right)-\nabla_{x_1} h^{-}\left(x_1^k, x_2^k\right)\right\rangle  \\
&\leq\frac{L_1^-(x_2^k)\beta_1^k}{2}\left\| x_1^{k+1}-x_1^k\right\|^2
\\&\quad+\frac{1}{2L_1^-(x_2^k)\beta_1^k} \left\|\nabla_{x_1} h^{-}\left(z_1^k, x_2^k\right)-\nabla_{x_1} h^{-}\left(x_1^k, x_2^k\right)\right\|^2 \\
&\leq \frac{L_1^-(x_2^k)\beta_1^k}{2}\left\|x_1^{k+1}-x_1^k\right\|^2+\frac{L_1^-(x_2^k)}{2\beta_1^k}\left\| z_1^k-x_1^k\right\|^2.
\end{aligned}
\end{equation}
This together with \eqref{ama1:1} and \eqref{ama1:2}, we have
\begin{equation}\label{proof_eq3}\begin{aligned}
&\left\langle x_1^{k+1}\right. \left.-x_1^k, \nabla_{x_1} h^{-}\left(z_1^k, x_2^k\right)-\nabla_{x_1} h^{-}\left(x_1^k, x_2^k\right)+\tau_1^k\left(y_1^k-x_1^k\right)\right\rangle  \\
&\leq \left(\frac{L_1^-(x_2^k)\beta_1^k}{2}+\frac{\alpha_1^k\tau_1^k}{2}\right)\left\|x_1^{k+1}-x_1^k\right\|^2\\&\quad+\left(\frac{L_1^-(x_2^k)\beta_1^k}{2}+\frac{\alpha_1^k\tau_1^k}{2}\right)\left\| x_1^k-x_1^{k-1}\right\|^2.
\end{aligned}
\end{equation}
Similarly,
{\footnotesize\begin{equation}\label{proof_eq4}
\begin{aligned}
&\left\langle x_2^{k+1}-x_2^k, \nabla_{x_2} h^{-}\left(x_1^{k+1}, z_2^k\right)-\nabla_{x_2} h^{-}\left(x_1^{k+1}, x_2^k\right)+\tau_2^k\left(y_2^k-x_2^k\right)\right\rangle  \\
&\leq \left(\frac{L_2^-(x_1^{k+1})\beta_2^k}{2}+\frac{\alpha_2^k\tau_2^k}{2}\right)\left\|x_2^{k+1}-x_2^k\right\|^2\\&\quad+\left(\frac{L_2^-(x_1^{k+1})\beta_2^k}{2}+\frac{\alpha_2^k\tau_2^k}{2}\right)\left\| x_2^k-x_2^{k-1}\right\|^2.
\end{aligned}
\end{equation}}Recalling the strong convexity of ${\psi_k}(\cdot)$ and ${\varphi_k}(\cdot)$, we accordingly chose ${\psi_k}(\cdot)$ and ${\varphi_k}(\cdot)$ with strongly convex modulus $\rho_1$ and $\rho_2$, respectively, so that  
\begin{equation}\label{proof_eq5}
\begin{aligned}
&D_{{\psi_k}}\left(x_1^{k+1}, x_1^k\right) \geq \frac{\rho_1}{2}\left\|x_1^{k+1}-x_1^k\right\|^2,\\
&D_{\varphi_k}\left(x_2^{k+1}, x_2^k\right) \geq \frac{\rho_2}{2}\left\|x_2^{k+1}-x_2^k\right\|^2.
\end{aligned}
\end{equation}
Substituting \eqref{proof_eq3}, \eqref{proof_eq4} and \eqref{proof_eq5} into \eqref{proof_eq2}, we obtain
{\small
\begin{equation} \label{proof_eq6}
\begin{aligned}
&\Phi\left(x_1^{k+1},x_2^{k+1}\right) - \Phi\left(x_1^k,x_2^k\right)\\
&\leq 
-\left(\frac{\rho_1\tau_1^k}{2}-\frac{L_1^{-}\left(x_2^k\right)}{2}-\frac{L_1^-(x_2^k)\beta_1^k}{2}-\frac{\alpha_1^k\tau_1^k}{2}\right)\left\|x_1^{k+1}-x_1^k\right\|^2\\
&\quad+\left(\frac{L_1^-(x_2^k)\beta_1^k}{2}+\frac{\alpha_1^k\tau_1^k}{2}\right)\left\| x_1^k-x_1^{k-1}\right\|^2\\
&\quad-\left(\frac{\rho_2\tau_2^k}{2}-\frac{L_2^{-}\left(x_1^{k+1}\right)}{2}-\frac{L_2^-(x_1^{k+1})\beta_2^k}{2}-\frac{\alpha_2^k\tau_2^k}{2}\right)\left\|x_2^{k+1}-x_2^k\right\|^2\\
&\quad+\left(\frac{L_2^-(x_1^{k+1})\beta_2^k}{2}+\frac{\alpha_2^k\tau_2^k}{2}\right)\left\| x_2^k-x_2^{k-1}\right\|^2.
\end{aligned}
\end{equation}}According to Assumptions \ref{ass2} and \ref{ass3} and the definitions of $\tau_1^k$ and $\delta_1$ in \eqref{tauk} and \eqref{delta}, respectively, we have
$$\rho_1\tau_1^k-L_1^{-}\left(x_2^k\right)-L_1^-(x_2^k)\beta_1^k-\alpha_1^k\tau_1^k-\delta_1=(1+\varepsilon)\delta_1-\delta_1= \varepsilon\delta_1$$
and
$$\begin{aligned}&\delta_1-L_1^-(x_2^k)\beta_1^k-\alpha_1^k\tau_1^k\\= &\ \delta_1-\frac{(1+\varepsilon)\delta_1+(1+\beta_1^k)L_1^-(x_2^k)}{\rho_1-\alpha_1^k}\alpha_1^k-L_1^-(x_2^k)\beta_1^k\\
\geq&\ \delta_1-(1-\varepsilon) \delta_1= \varepsilon \delta_1.\end{aligned}$$
Similarly, for the $\tau_2^k$ and $\delta_2$ defined in \eqref{tauk} and \eqref{delta}, we have
$$\rho_2\tau_2^k-L_2^{-}\left(x_1^{k+1}\right)-L_2^-(x_1^{k+1})\beta_2^k-\alpha_2^k\tau_2^k-\delta_2= \varepsilon\delta_2$$
and
$$\begin{aligned}\delta_2-L_2^-(x_1^{k+1})\beta_2^k-\alpha_2^k\tau_2^k =\delta_2-(1-\varepsilon) \delta_2^k\geq \varepsilon\delta_2.\end{aligned}$$
Plugging them into \eqref{proof_eq6}, we have
\begin{equation} 
\begin{aligned}
&\Phi\left(x_1^{k+1},x_2^{k+1}\right)+\frac{\delta_1}{2}\|x_1^{k+1}-x_1^k\|^2 +\frac{\delta_2}{2}\|x_2^{k+1}-x_2^k\|^2
\\
&\leq \Phi\left(x_1^k,x_2^k\right)+\frac{\delta_1}{2}\|x_1^{k}-x_1^{k-1}\|^2 +\frac{\delta_2}{2}\|x_2^{k}-x_2^{k-1}\|^2\\
&\leq 
- \frac{\varepsilon\delta_1}{2} \left(\left\|x_1^{k+1}-x_1^k\right\|^2+\left\| x_1^k-x_1^{k-1}\right\|^2\right)\\&\quad - \frac{\varepsilon\delta_2}{2} \left(\left\|x_2^{k+1}-x_2^k\right\|^2+\left\| x_2^k-x_2^{k-1}\right\|^2\right).
\end{aligned}
\end{equation}
This together the definition of $\Theta_{\delta_1,\delta_2}$ in \eqref{Theta} and $\delta=\frac{\varepsilon}{2}\min\{\delta_1,\delta_2\}$, we obtain \eqref{descent} and the proof is completed.
\end{proof}

\begin{lemma}\label{lemma2}
Let $\left\{\left(x_1^k, x_2^k\right)\right\}_{k \in \mathbb{N}}$ be a sequence generated by Algorithm \ref{algo1}, for all $k \in \mathbb{N}$, $w_1^k=$ $\left(x_1^k, x_2^k\right), w_2^k=\left(x_1^{k-1}, x_2^{k-1}\right)$, and $\mathbf{w}^k=\left(w_1^k, w_2^k\right)$. Suppose that Assumptions \ref{ass1}, \ref{ass2} and \ref{ass3} hold. 
Then, for any $k \geq  1$,
 there exists $\gamma>0$ such that 
 \begin{equation}\label{lemma3_eq1}
 {\rm dist}\left(0,\partial \Theta_{\delta_1,\delta_2}({\bf w}^{k+1})\right)\leq \gamma\left\|{\bf w}^{k+1}-{\bf w}^k\right\|.
 \end{equation}    
\end{lemma}
% The proof of Lemma \ref{lemma2} can be found in Appendix \ref{Plemma2}.
%\section{Proof of Lemma \ref{lemma2}}\label{Plemma2}
\begin{proof}
According to \eqref{bprof}, the optimality condition of the subproblem \eqref{ama1:3} can be read as 
\begin{equation}\label{optimality_x1}
\begin{aligned}
0\in\ &\partial\theta_1(x_1^{k+1})+\nabla_{x_1} h^+(x_1^{k+1},x_2^k)-\nabla_{x_1} h^-(z_1^k,x_2^k)\\&-\tau_1^k(y_1^k-x_1^k)+\tau_1^k(\nabla {\psi_k}(x_1^{k+1})-\nabla{\psi_k}(x_1^k)).
\end{aligned}
\end{equation}
Similarly, the optimality condition of the subproblem \eqref{ama2:3} can be written as
\begin{equation}\label{optimality_x2}
\begin{aligned}
0\in\ &\partial\theta_2(x_2^{k+1})+\nabla_{x_2} h^+(x_1^{k+1},x_2^{k+1})-\nabla_{x_2} h^-(x_1^{k+1},z_2^k)\\&-\tau_2^k(y_2^k-x_2^k)+\tau_2^k(\nabla \varphi(x_2^{k+1})-\nabla\varphi(x_2^k)).
\end{aligned}
\end{equation}
Firstly, from the definition of $\Theta_{\delta_1,\delta_2}$ in \eqref{Theta}, we have
 \begin{equation}\label{lemma3_eq2}
\begin{aligned}
 &\partial_{w_{11}} \Theta_{\delta_1,\delta_2}({\bf w}^{k+1}) \\
& = \partial\theta_1(x_1^{k+1})+\nabla_{x_1} h^+(x_1^{k+1},x_2^{k+1})-\nabla_{x_1} h^-(x_1^{k+1},x_2^{k+1})\\&\quad+\delta_1(x_1^{k+1}-x_1^k)\\
&\ni \nabla_{x_1} h^+(x_1^{k+1},x_2^{k+1})-\nabla_{x_1} h^+(x_1^{k+1},x_2^k)\\&\quad+\nabla_{x_1} h^-(z_1^k,x_2^k)-\nabla_{x_1} h^-(x_1^{k+1},x_2^{k+1})\\
&\quad+\tau_1^k(y_1^k-x_1^k)+\delta_1(x_1^{k+1}-x_1^k)\\
&\quad-\tau_1^k(\nabla {\psi_k}(x_1^{k+1})-\nabla{\psi_k}(x_1^k)),
\end{aligned}
\end{equation}
where the last inclusion follows from \eqref{optimality_x1}.
Secondly, we compute the subgradient of $\Theta_{\delta_1,\delta_2}$ with respect to $w_{12}$,  
 \begin{equation}\label{lemma3_eq3}
\begin{aligned}
& \partial_{w_{12}} \Theta_{\delta_1,\delta_2}({\bf w}^{k+1}) \\
&= \partial\theta_2(x_2^{k+1})+\nabla_{x_2} h^+(x_1^{k+1},x_2^{k+1})-\nabla_{x_2} h^-(x_1^{k+1},x_2^{k+1})\\&\quad+\delta_2(x_2^{k+1}-x_2^k)\\
& \ni \nabla_{x_2} h^-(x_1^{k+1},z_2^k)-\nabla_{x_2} h^-(x_1^{k+1},x_2^{k+1})\\
&\quad+\tau_2^k(y_2^k-x_2^k)+\delta_2(x_2^{k+1}-x_2^k)-\tau_2^k(\nabla \varphi(x_1^{k+1})-\nabla\varphi(x_1^k)),
\end{aligned}
\end{equation}
where the inclusion follows from \eqref{optimality_x2}. By \eqref{Theta}, we obtain 
\begin{equation}\label{lemma3_eq5}
\begin{aligned}
&\nabla_{w_{21}} \Theta_{\delta_1,\delta_2}({\bf w}^{k+1})=\delta_1\left( x_1^k-x_1^{k+1}\right),\\
&\nabla_{w_{22}} \Theta_{\delta_1,\delta_2}({\bf w}^{k+1})=\delta_2\left( x_2^k-x_2^{k+1}\right).
\end{aligned}
\end{equation}
Since $\nabla h$ is Lipschitz continuous on bounded subsets of $\mathbb{R}^{n_1} \times \mathbb{R}^{n_2}$ from Assumption \ref{ass2}, it follows from \eqref{lemma3_eq2} that
$$
\begin{aligned}
&{\rm dist}(0,\partial_{w_{11}} \Theta_{\delta_1,\delta_2}({\bf w}^{k+1})) \\
\leq &  \|\nabla_{x_1} h^+(x_1^{k+1},x_2^{k+1})-\nabla_{x_1} h^+(x_1^{k+1},x_2^k)\|\\&+\|\nabla_{x_1} h^-(z_1^k,x_2^k)-\nabla_{x_1} h^-(x_1^{k+1},x_2^{k+1})\|\\
&+\tau_1^k\|y_1^k-x_1^k\|+\delta_1\|x_1^{k+1}-x_1^k\|\\&+\tau_1^k\|\nabla {\psi_k}(x_1^{k+1})-\nabla{\psi_k}(x_1^k)\| \\
\leq & L^+\|x_2^{k+1}-x_2^k\|+ L^-\|z_1^k-x_1^{k+1}\|+L^-\|x_2^k-x_2^{k+1}\|\\
&+\tau_1^k\alpha_1^k\|x_1^k-x_1^{k-1}\|+\delta_1\|x_1^{k+1}-x_1^k\|+\tau_1^k L_{{\psi_k}}\|x_1^{k+1}-x_1^k\| \\
\leq & L\|x_2^{k+1}-x_2^k\|+(\tau_1^+\bar\alpha_1+\bar\beta_1 L^-)\|x_1^k-x_1^{k-1}\|\\&+(\delta_1+\tau_1^+ L_{\psi})\|x_1^{k+1}-x_1^k\|,
\end{aligned}
$$
where $L=L^++L^-$, $L_{\psi}=\sup\{L_{\psi_k}:k\in\mathbb{N}\}$, the first and second inequalities follow from Assumption \ref{ass2}, \eqref{ama1:1} and \eqref{ama1:2}, the fact the sequence $\left\{\tau_1^k\right\}_{k \in \mathbb{N}}$ is bounded from above by $\tau_1^{+}$, and $\alpha_1^k\leq\bar\alpha_1, \beta_1^k \leq \bar\beta_1$ for all $k \in \mathbb{N}$. 
Moreover, it follows from \eqref{lemma3_eq3} that
$$
\begin{aligned}
&{\rm dist}(0,\partial_{w_{12}} \Theta_{\delta_1,\delta_2}({\bf w}^{k+1})) \\
\leq &  \|\nabla_{x_2} h^-(x_1^{k+1},z_2^k)-\nabla_{x_2} h^-(x_1^{k+1},x_2^{k+1})\|+\tau_2^k\|y_2^k-x_2^k\|\\
&+\delta_2\|x_2^{k+1}-x_2^k\|+\tau_2^k\|\nabla \varphi_k(x_2^{k+1})-\nabla\varphi_k(x_2^k)\| \\
\leq &\lambda_2^+\|z_2^k-x_2^{k+1}\|+\tau_2^k\alpha_2^k\|x_2^k-x_2^{k-1}\|\\
&+\delta_2\|x_2^{k+1}-x_2^k\|+\tau_2^kL_{\varphi_k}\|x_2^{k+1}-x_2^k\| \\
\leq & \lambda_2^+\|x_2^k-x_2^{k+1}\|+(\lambda_2^+\bar\beta_2+\tau_2^+\bar\alpha_2)\|x_2^k-x_2^{k-1}\|
\\
&+(\delta_2+\tau_2^+L_\varphi)\|x_2^{k+1}-x_2^k\|,
\end{aligned}
$$
where $L_{\varphi}=\sup\{L_{\varphi_k}:k\in\mathbb{N}\}$.
Summing up these estimations, we get from \eqref{Theta} that
\begin{equation} \begin{aligned}
 &{\rm dist}\left(0,\partial \Theta_{\delta_1,\delta_2}({\bf w}^{k+1})\right) \\
 &\leq {\rm dist}(0,\partial_{w_{11}} \Theta_{\delta_1,\delta_2}({\bf w}^{k+1}))+{\rm dist}(0,\partial_{w_{12}} \Theta_{\delta_1,\delta_2}({\bf w}^{k+1}))\\
 &\quad+\|\nabla_{w_{21}} \Theta_{\delta_1,\delta_2}({\bf w}^{k+1})\|+\|\nabla_{w_{22}} \Theta_{\delta_1,\delta_2}({\bf w}^{k+1})\| \\
 &\leq  L\|x_2^{k+1}-x_2^k\|+(\tau_1^+\bar\alpha_1+\bar\beta_1 L^-)\|x_1^k-x_1^{k-1}\|\\
 &\quad+(\delta_1+\tau_1^+ L_{{\psi_k}})\|x_1^{k+1}-x_1^k\|\\
 &\quad + \lambda_2^+\|x_2^k-x_2^{k+1}\|+(\lambda_2^+\bar\beta_2+\tau_2^+\bar\alpha_2)\|x_2^k-x_2^{k-1}\|
\\\nonumber
 &\quad+(\delta_2+\tau_2^+L_\varphi)\|x_2^{k+1}-x_2^k\|\\
&\quad +\delta_1\| x_1^k-x_1^{k+1}\|+\delta_2\| x_2^k-x_2^{k+1}\|. 
% \\
% &\leq \tilde\gamma \left(\|x_1^{k+1}-x_1^k\|+\|x_1^k-x_1^{k-1}\|\right)\\&\quad+\tilde\gamma \left(\|x_2^{k+1}-x_2^k\|+\|x_2^k-x_2^{k-1}\| \right)\\
% &\leq 2\sqrt{2}\tilde\gamma\|{\bf w}^{k+1}-{\bf w}^k\|,
\end{aligned} \end{equation}   
Let $\tilde\gamma = \max\{L+\lambda_2^++2\delta_2+\tau_2^+L_\varphi,\tau_1^+\bar\alpha_1+\bar\beta_1 L^-,2\delta_1+\tau_1^+ L_{\psi},\lambda_2^+\bar\beta_2+\tau_2^+\bar\alpha_2 \}$, we have
\begin{equation} \begin{aligned}
 &{\rm dist}\left(0,\partial \Theta_{\delta_1,\delta_2}({\bf w}^{k+1})\right) \\
&\leq \tilde\gamma \left(\|x_1^{k+1}-x_1^k\|+\|x_1^k-x_1^{k-1}\|\right)\\&\quad+\tilde\gamma \left(\|x_2^{k+1}-x_2^k\|+\|x_2^k-x_2^{k-1}\| \right)\\
&\leq 2\sqrt{2}\tilde\gamma\|{\bf w}^{k+1}-{\bf w}^k\|,
\end{aligned} \end{equation}   
This completes the proof with $\gamma=2\sqrt{2}\tilde\gamma$.
\end{proof}

\begin{theorem}\label{theo1}
Let $\left\{\left(x_1^k, x_2^k\right)\right\}_{k \in \mathbb{N}}$ be a sequence generated by Algorithm \ref{algo1}, for all $k \in \mathbb{N}$, $w_1^k=$ $\left(x_1^k, x_2^k\right), w_2^k=\left(x_1^{k-1}, x_2^{k-1}\right)$, and $\mathbf{w}^k=\left(w_1^k, w_2^k\right)$. Suppose that Assumptions \ref{ass1}, \ref{ass2} and \ref{ass3} hold, and $\{\mathbf{w}^k\}$ is assumed to be bounded. Then,
\begin{itemize}
\item [(i)]  it holds that 
$\sum_{k=0}^{\infty}\left\|\mathbf{w}^{k+1}-\mathbf{w}^k\right\|^2<\infty
$, and furthermore
$\lim _{k \rightarrow \infty}\left\|\mathbf{w}^{k+1}-\mathbf{w}^k\right\|=0.$ 
\item[(ii)] The set $z({\bf w}^0)$ of the cluster point of $\left\{{\bf w}^k\right\}_{k\in\mathbb{N}}$ is nonempty and compact, and any point in $z({\bf w}^0)$ is the critical point of \eqref{problem}.
\item[(iii)] The limit $\lim_{k\rightarrow\infty}\Theta_{\delta_1,\delta_2}({\bf w}^k)=\Theta^\infty$ exist, and $\Theta_{\delta_1,\delta_2}({\bf w}^*)=\Theta^\infty$ for all ${\bf w}^*\in z({\bf w}^0)$.
\end{itemize}
\end{theorem}
%The proof of Theorem \ref{theo1} can be found in Appendix \ref{Ptheo1}. 
%\section{Proof of Theorem \ref{theo1}}\label{Ptheo1}
\begin{proof}
We first prove (i).
From Lemma \ref{lemma1}, for any $N \in \mathbb{N}$,  
\begin{equation}\label{proof2_eq1}
 \sum_{k=0}^N\delta\left\|\mathbf{w}^{k+1}-\mathbf{w}^k\right\|^2\leq \Theta_{\delta_1,\delta_2}\left(\mathbf{w}^0\right)-\Theta_{\delta_1,\delta_2}\left(\mathbf{w}^{N+1}\right) .  
\end{equation}
According to Assumption \ref{ass1}, we know that $\Phi$ is bounded from below, and hence  $\Theta_{\delta_1,\delta_2}$ is also bounded from below because $\Theta_{\delta_1,\delta_2}(\cdot,\cdot) \geq \Phi(\cdot)$. Thus, letting $N \rightarrow \infty$ in \eqref{proof2_eq1} yields that
$$   
\sum_{k=0}^{\infty}\delta\left\|\mathbf{w}^{k+1}-\mathbf{w}^k\right\|^2<\infty,
$$
which means that
$$
\lim _{k \rightarrow \infty}\left\|\mathbf{w}^{k+1}-\mathbf{w}^k\right\|=0 .
$$
This fact together with \eqref{descent} implies that $\left\|\mathbf{w}^k\right\| \rightarrow 0$ as $k \rightarrow \infty$.

Now we prove (ii).   
Since $\left\{\mathbf{w}^k\right\}_{k \in \mathbb{N}}$ is assumed to be bounded, the set $z\left(\mathbf{w}^0\right)$ is nonempty. 
Thus the limit point $\mathbf{w}^*=\left(x_1^*, x_2^*, \hat{x}_1, \hat{x}_2\right)$ of the subsequence of $\left\{\mathbf{w}^k\right\}_{k \in \mathbb{N}^*}$ of $\left\{\mathbf{w}^{k_j}\right\}_{j \in \mathbb{N}}$ is exits. 
We will prove that $\mathbf{w}^*$ is a critical point of $\Theta_{\delta_1,\delta_2}$ defined in \eqref{Theta}. 
It follows from \eqref{ama1:3} and $k=k_j-1$ that
\begin{equation}\label{proof3_eq1}\small
\begin{aligned}
&\theta_1\left(x_1^{k_j}\right)+h^+(x_1^{k_j},x_2^{k_j-1}) +\tau_1^{k_j-1}D_{\psi}(x_1^{k_j},x_1^{k_j-1})\\
& \leq\theta_1\left(x_1^*\right)+h^+(x_1^*,x_2^{k_j-1})+\tau_1^{k_j-1}D_{\psi}(x_1^*,x_1^{k_j-1})\\
&\quad-\left\langle x_1^*-x_1^{k_j}, \nabla_{x_1} h^-(z_1^{k_j-1}, x_2^{k_j-1}) 
+\tau_1^{k_j-1}(y_1^{k_j-1}-x_1^{k_j-1})\right\rangle.
\end{aligned}
\end{equation}
Similarly, it follows from \eqref{ama2:3} that
\begin{equation}\label{proof3_eq2}\small
\begin{aligned}
&\theta_2\left(x_2^{k_j}\right)+h^+(x_1^{k_j},x_2^{k_j}) +\tau_2^{k_j-1}D_{\phi}(x_2^{k_j},x_2^{k_j-1})\\
& \leq\theta_2\left(x_2^*\right)+h^+(x_1^{k_j},x_2^*)+\tau_2^{k_j-1}D_{\phi}(x_2^*,x_2^{k_j-1})\\
&\quad-\left\langle x_2^*-x_2^{k_j}, \nabla_{x_2} h^-(x_1^{k_j}, z_2^{k_j-1}) 
+\tau_2^{k_j-1}(y_2^{k_j-1}-x_2^{k_j-1})\right\rangle.
\end{aligned}
\end{equation}
Hence, it follows from the nonnegativity of Bregman distance and the inequalities \eqref{proof3_eq1}
and \eqref{proof3_eq2} that
\begin{equation} 
\begin{aligned}
&\lim_{j\rightarrow\infty}\Theta_{\delta_1,\delta_2}({\bf w}^k) \\
 &=\lim _{j \rightarrow \infty}\Big\{\theta_1\left(x_1^{k_j}\right)+\theta_2\left(x_2^{k_j}\right)+h\left(x_1^{k_j},x_2^{k_j}\right)\\
&\quad+\frac{\delta_1}{2}\left\|x_1^{k_j}-x_1^{k_j-1}\right\|^2+\frac{\delta_2}{2}\left\|x_2^{k_j}-x_2^{k_j-1}\right\|^2\Big\}\\
&\leq\lim _{j \rightarrow \infty}\Big\{\theta_1\left(x_1^{k_j}\right)+\theta_2\left(x_2^{k_j}\right)+h\left(x_1^{k_j},x_2^{k_j}\right)\\
&\quad+\frac{\delta_1}{2}\left\|x_1^{k_j}-x_1^{k_j-1}\right\|^2+\frac{\delta_2}{2}\left\|x_2^{k_j}-x_2^{k_j-1}\right\|^2\\
&\quad+\tau_1^{k_j-1}D_{\psi}(x_1^{k_j},x_1^{k_j-1})+\tau_2^{k_j-1}D_{\psi}(x_2^{k_j},x_2^{k_j-1})\Big\}\\
& \leq \limsup _{j \rightarrow \infty}\Big\{
\theta_1\left(x_1^*\right)+h^+(x_1^*,x_2^{k_j-1})\\
&\quad+\tau_1^{k_j-1}D_{\psi}(x_1^*,x_1^{k_j-1})-h^+(x_1^{k_j},x_2^{k_j-1})\\
&\quad-\left\langle x_1^*-x_1^{k_j}, \nabla_{x_1} h^-(z_1^{k_j-1}, x_2^{k_j-1}) 
+\tau_1^{k_j-1}(y_1^{k_j-1}-x_1^{k_j-1})\right\rangle\\
&\quad+\frac{\delta_1}{2}\left\|x_1^{k_j}-x_1^{k_j-1}\right\|^2+\theta_2\left(x_2^*\right)\\
&\quad+h^+(x_1^{k_j},x_2^*)+\tau_2^{k_j-1}D_{\phi}(x_2^*,x_2^{k_j-1})-h^-(x_1^{k_j},x_2^{k_j})\\
&\qquad-\left\langle x_2^*-x_2^{k_j}, \nabla_{x_2} h^-(x_1^{k_j}, z_2^{k_j-1}) 
+\tau_2^{k_j-1}(y_2^{k_j-1}-x_2^{k_j-1})\right\rangle\\
&\quad+\frac{\delta_2}{2}\left\|x_2^{k_j}-x_2^{k_j-1}\right\|^2\Big\}\\
&=\theta_1(x_1^*)+h(x_1^*,x_2^*)+\theta_2(x_2^*)\leq \Theta_{\delta_1,\delta_2}({\bf w}^*)=\Theta^*.
\end{aligned}
\end{equation}
where the second equality follows from the continuity of $h(\cdot, \cdot)$ and $\lim _{j \rightarrow \infty}\left\|\boldsymbol{w}^{k_j}-\boldsymbol{w}^{k_j-1}\right\|=0$ proved in (i). From Assumption \ref{ass1}, we have
$$
\liminf _{j \rightarrow \infty} \theta_1\left(x_1^{k_j}\right) \geq \theta_1\left(x_1^*\right) \text { and } \liminf _{j \rightarrow \infty} \theta_2\left(x_2^{k_j}\right) \geq \theta_2\left(x_2^*\right).
$$
The above relations yield
$$
\Theta^*\equiv\Theta_{\delta_1,\delta_2}\left(\bf{w}^*\right)\leq \liminf _{j \rightarrow \infty} \Theta_{\delta_1,\delta_2}\left({\bf w}^{k_j}\right)  .
$$
Consequently, $\lim _{j \rightarrow \infty} \Theta_{\delta_1,\delta_2}\left({\bf w}^{k_j}\right)=\Theta^*$. On the other hand, it follows from Lemma \ref{lemma2} that
$$
\lim _{j \rightarrow \infty} \operatorname{dist}\left(0, \partial \Theta_{\delta_1,\delta_2}\left(\bf{w}^{k_j}\right)\right) \leq \lim _{j \rightarrow \infty} \gamma\left\|\bf{w}^{k_j}-\bf{w}^{k_j-1}\right\|=0 .
$$
Now, the closedness property of $\partial \Theta_{\delta_1,\delta_2}$ refers that $0 \in \partial \Psi\left(\mathbf{w}^*\right)$, which proves that $\mathbf{w}^*$ is a critical point of $\Theta_{\delta_1,\delta_2}$.  

Now we prove (iii). The sequence $\{\Theta_{\delta_1,\delta_2}({\bf w}^k)\}$ is bounded below by Assumption \ref{ass1}. The sequence $\{\Theta_{\delta_1,\delta_2}({\bf w}^k)\}$
is nonincreasing
from \eqref{descent}. Thus, the limit $\lim_{k\rightarrow\infty}\Theta_{\delta_1,\delta_2}({\bf w}^k)=\Theta^\infty$ exists.
Taking ${\bf w}^*\in z({\bf w}^0)$ as an accumulation point of $\{{\bf w}^k\}$, then there exists a subsequence $\{{\bf w}^{k_j}\}$  satisfying  $\lim_{j\rightarrow\infty}{\bf w}^{k_j}={\bf w}^*$. From Items (i) and (ii), we have
$$\Theta_{\delta_1,\delta_2}({\bf w}^*)=\lim_{j\rightarrow\infty}\Theta_{\delta_1,\delta_2}({\bf w}^{k_j})=\lim_{k\rightarrow\infty}\Theta_{\delta_1,\delta_2}({\bf w}^{k})=\Theta^\infty.$$
We complete the proof by the arbitrariness of ${\bf w}^*$ in $ z({\bf w}^0)$.
\end{proof}

\begin{theorem}\label{theo2}
Let $\left\{\left(x_1^k, x_2^k\right)\right\}_{k \in \mathbb{N}}$ be a sequence generated by Algorithm \ref{algo1}, for all $k \in \mathbb{N}$, $w_1^k=$ $\left(x_1^k, x_2^k\right), w_2^k=\left(x_1^{k-1}, x_2^{k-1}\right)$, and $\mathbf{w}^k=\left(w_1^k, w_2^k\right)$. Suppose that Assumptions \ref{ass1}, \ref{ass2} and \ref{ass3} hold, and $\{\mathbf{w}^k\}$ is assumed to be bounded. If $\Phi$ in \eqref{problem} is a KL function. Then, 
the sequence $\left\{{\bf w}^k\right\}_{k \geq 1}$ has finite length, that is,  
$
 \sum_{k = 1}^\infty\left\|{\bf w}^{k+1}-{\bf w}^k\right\|<+\infty.
$
Hence, the whole sequence $\{{\bf w}^k\}_{k\geq 1} $ is convergent.
\end{theorem}
%The proof of Theorem \ref{theo2} can be found in Appendix \ref{Ptheo2}. 
%\section{Proof of Theorem \ref{theo2}}\label{Ptheo2}
To prove Theorem \ref{theo2}, we need to introduce a uniformized KL property which was established in \cite{bolte2014alternating} in the following, it will be useful for further convergence analysis.

\begin{lemma}\label{Lem2.1}{\rm\cite{bolte2014alternating}} {\rm (Uniformized KL property)} Let $\Omega$ be a compact set and $f:\mathbb{R}^{n}\rightarrow(-\infty,+\infty]$ be a proper and lower semicontinuous function. Assume that $f$ is a constant on $\Omega$ and satisfies the KL property at each point of $\Omega$. Then, there exist $\varsigma>0,~\eta>0$ and $\varphi\in \Phi_{\eta}$ such that \begin{equation}
\varphi'(f(x)-f(\bar x)){\rm dist}(0,\partial f(x))\geq1,
\end{equation}
 for all $\bar x\in \Omega$ and each $x$ satisfying
${\rm dist}(x,\Omega)<\varsigma$ and $f(\bar x) < f(x) < f(\bar x)+\eta.$
\end{lemma}

Now we begin to prove Theorem \ref{theo2}.
\begin{proof}
 It follows from Theorem \ref{theo1} that for any ${\bf w}^*\in z({\bf w}^0)$, there exists a subsequence $\{{\bf w}^{k_j}\}$ of $\{{\bf w}^k\}$ converging to ${\bf w}^*$, and
 $\lim_{k\rightarrow \infty}  \Theta_{\delta_1,\delta_2} ({\bf w}^k)=\Theta_{\delta_1,\delta_2} ({\bf w}^*)$.
If there exists an integer $\bar k$ such that $\Theta_{\delta_1,\delta_2} ({\bf w}^k)=\Theta_{\delta_1,\delta_2} ({\bf w}^*)$, then from Lemma \ref{lemma1}, we have
$$\begin{aligned}
\delta\left\|\mathbf{w}^{k+1}-\mathbf{w}^k\right\|^2&\leq \Theta_{\delta_1,\delta_2} ({\bf w}^k)-\Theta_{\delta_1,\delta_2} ({\bf w}^{k+1})\\&\leq \Theta_{\delta_1,\delta_2} ({\bf w}^{\bar k})-\Theta_{\delta_1,\delta_2} ({\bf w}^*)=0, ~~ \forall k>\bar k. 
\end{aligned}
$$
Hence, we have ${\bf w}^{k+1}={\bf w}^k$  for any $k>\bar k$. Thus the assertion $\sum_{k = 1}^\infty\left\|{\bf w}^{k+1}-{\bf w}^k\right\|<+\infty$ holds trivially.
Otherwise, since $\Theta_{\delta_1,\delta_2} ({\bf w}^k)$ is nonincreasing from Lemma \ref{lemma1}, we have $\Theta_{\delta_1,\delta_2} ({\bf w}^k)>\Theta_{\delta_1,\delta_2} ({\bf w}^*)$ for all $k$. Again from $\lim_{k\rightarrow \infty}  \Theta_{\delta_1,\delta_2} ({\bf w}^k)=\Theta_{\delta_1,\delta_2} ({\bf w}^*)$, we know that for any $\eta>0$, there exists a nonnegative integer $k_0$ such that $\Theta_{\delta_1,\delta_2} ({\bf w}^k)<\Theta_{\delta_1,\delta_2}({\bf w}^*)+\eta$ for any $k>k_0$. In addition, for any $\varsigma>0$ there exists a positive integer $k_1$ such that ${\rm dist}({\bf w}^{k},z({\bf w}^0))<\varsigma$ for all $k>k_1$. Consequently, for any $\eta,~\varsigma>0$, when $k>k_2:=\max\{k_0,k_1\}$, we have
$${\rm dist}({\bf w}^k,z({\bf w}^0))<\varsigma \quad \hbox{and} \quad \Theta_{\delta_1,\delta_2} ({\bf w}^k)<\Theta_{\delta_1,\delta_2}({\bf w}^*)+\eta.$$
Since $z({\bf w}^0)$ is a nonempty and compact set, and $\Theta_{\delta_1,\delta_2}$ is a constant on $z({\bf w}^0)$, we can apply Lemma \ref{Lem2.1} with $\Omega:=z({\bf w}^0)$.  Therefore, for any $k>k_2$, we have
\begin{equation}\label{var}
{\varphi_k}'(\Theta_{\delta_1,\delta_2}({\bf w}^k)-\Theta_{\delta_1,\delta_2}({\bf w}^*)){\rm dist}(0,\partial \Theta_{\delta_1,\delta_2}({\bf w}^k))\geq 1.
\end{equation}
From the concavity of ${\varphi_k}$, we have
{\footnotesize
$$\begin{aligned}
&{\varphi_k} (\Theta_{\delta_1,\delta_2}({\bf w}^k)-\Theta_{\delta_1,\delta_2}({\bf w}^*))-{\varphi_k} (\Theta_{\delta_1,\delta_2}({\bf w}^{k+1})-
\Theta_{\delta_1,\delta_2}({\bf w}^*))\\&\geq{\varphi_k}'(\Theta_{\delta_1,\delta_2}({\bf w}^k)-\Theta_{\delta_1,\delta_2}({\bf w}^*))( \Theta_{\delta_1,\delta_2}({\bf w}^k)-\Theta_{\delta_1,\delta_2}({\bf w}^{k+1})).
\end{aligned}$$}
For convenience, for all $p,q\in\mathbb{ N}$, we define
$$\begin{aligned}
\zeta_{p,q}:=&{\varphi_k}\big(\Theta_{\delta_1,\delta_2}(u^p)-\Theta_{\delta_1,\delta_2}({\bf w}^*)\big)\\
&-{\varphi_k}\big(\Theta_{\delta_1,\delta_2}(u^q)-\Theta_{\delta_1,\delta_2}({\bf w}^*)\big).   
\end{aligned}$$
Then, associated with ${\varphi_k}'(\Theta_{\delta_1,\delta_2}({\bf w}^k)-\Theta_{\delta_1,\delta_2}({\bf w}^*))>0$, ${\rm dist}(0,\partial \Theta_{\delta_1,\delta_2}({\bf w}^k))\leq \gamma\|{\bf w}^k-{\bf w}^{k-1}\|$ in Lemma \ref{lemma2}, and \eqref{var}, we get
\begin{equation*}
\begin{split}
&\Theta_{\delta_1,\delta_2}({\bf w}^k)-\Theta_{\delta_1,\delta_2}({\bf w}^{k+1})\leq\frac{\zeta_{k,k+1}}{{\varphi_k}'(\Theta_{\delta_1,\delta_2}({\bf w}^k)-\Theta_{\delta_1,\delta_2}({\bf w}^*))}\\
&\leq \gamma\|{\bf w}^k-{\bf w}^{k-1}\|\zeta_{k,k+1}.
\end{split}
\end{equation*}
Combining Lemma \ref{lemma1} and the above relation, it yields that for any $k>k_2$,
$$\begin{aligned} 
&\delta\left\|\mathbf{w}^{k+1}-\mathbf{w}^k\right\|^2\leq \Theta_{\delta_1,\delta_2} ({\bf w}^k)-\Theta_{\delta_1,\delta_2} ({\bf w}^{k+1})
\\&\leq  \gamma\|{\bf w}^k-{\bf w}^{k-1}\|\zeta_{k,k+1}.
\end{aligned}$$
This implies that
$$ \begin{aligned} &\left\|\mathbf{w}^{k+1}-\mathbf{w}^k\right\| \leq  \sqrt{\frac{\gamma}{\delta}\zeta_{k,k+1}\|{\bf w}^k-{\bf w}^{k-1}\|}
\\&\leq \frac{\gamma}{2\delta}\zeta_{k,k+1}+\frac{1}{2}\|{\bf w}^k-{\bf w}^{k-1}\|,\end{aligned}$$
which implies that $\sum_{k = 1}^\infty\left\|{\bf w}^{k+1}-{\bf w}^k\right\|<+\infty$. 
Thus, $\{{\bf w}^k\}_{k\geq 1}$ is a Cauchy sequence and hence it is convergent, which also implies that $\{(x_1^k,x_2^k)\}_{k\in\mathbb{N}}$ is convergent. Suppose that $\lim_{k\rightarrow\infty}(x_1^k,x_2^k)=(x_1^*,x_2^*)$. Then, we give that $0\in\partial\Phi(x_1^*,x_2^*)$. Hence, 
$ w_1^*=\left(w_{11}^*, w_{12}^*\right)=(x_1^*,x_2^*), w_2^*=\left(w_{21}^*, w_{22}^*\right) =(x_1^*,x_2^*)$, and $\mathbf{w}^*=\left(w_1^*, w_2^*\right)$.
$0\in\Theta_{\delta_1,\delta_2}({\bf w}^*)$ implies that
$$ \begin{aligned} 0\in (&\partial_{w_{11}}\Phi(w_1^*)+\delta_1(w_{11}^*-w_{12}^*),\partial_{w_{12}}\Phi(w_1^*)+\delta_2(w_{21}^*-w_{22}^*),\\&\delta_1(w_{12}^*-w_{11}^*),\delta_2(w_{22}^*-w_{21}^*) ),\end{aligned}$$
which means that
$$0\in (\partial_{w_{11}}\Phi(w_1^*),\partial_{w_{12}}\Phi(w_1^*))=\partial \Phi(x_1^*,x_2^*).$$
Therefore, $(x_1^*,x_2^*)$ is a critical point of \eqref{problem}. This completes the proof. 
\end{proof}

\begin{figure}[t!]
\setlength{\abovecaptionskip}{0in}
\subfigcapskip=-0.05in
\centering
\begin{minipage}{0.2\linewidth}
\centerline{\includegraphics[width=0.86in]{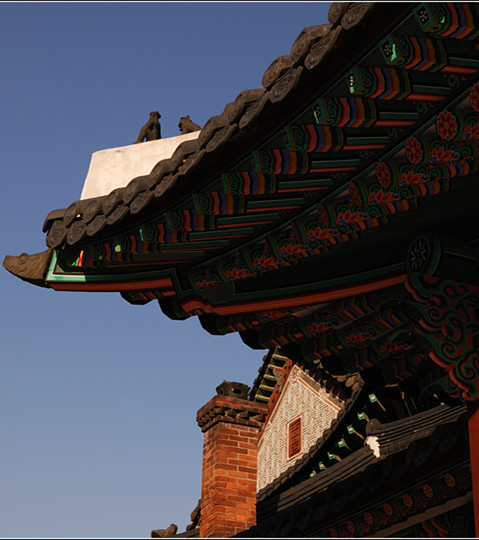}}\vspace{-0.03in}
\centerline{}
\end{minipage}\hspace{0.15in}
\begin{minipage}{0.2\linewidth}
\centerline{\includegraphics[width=0.86in]{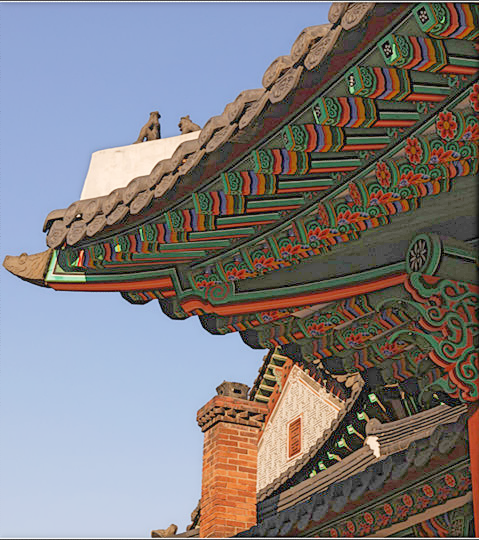}}\vspace{-0.03in}
\centerline{}
\end{minipage}\hspace{0.15in}
\begin{minipage}{0.2\linewidth}
\centerline{\includegraphics[width=0.86in]{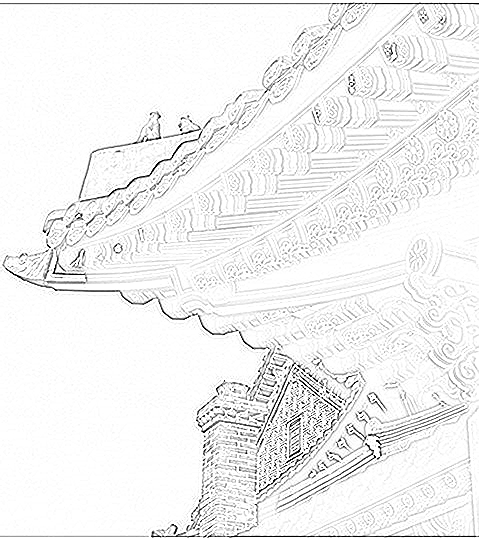}}\vspace{-0.03in}
\centerline{}
\end{minipage}\hspace{0.15in}
\begin{minipage}{0.2\linewidth}
\centerline{\includegraphics[width=0.86in]{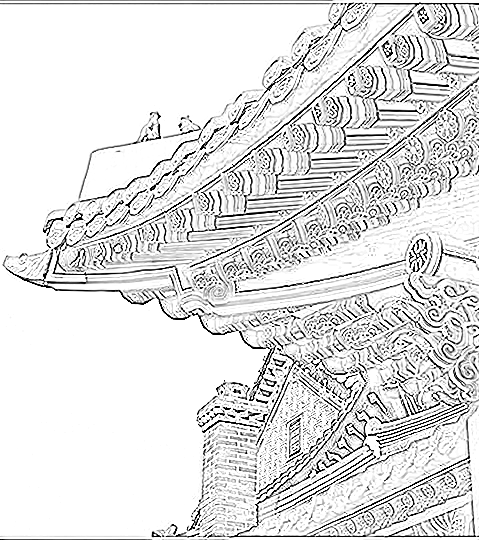}}\vspace{-0.03in}
\centerline{}
\end{minipage}\vspace{-0.1in}		

\begin{minipage}{0.2\linewidth}
\centerline{\includegraphics[width=0.86in]{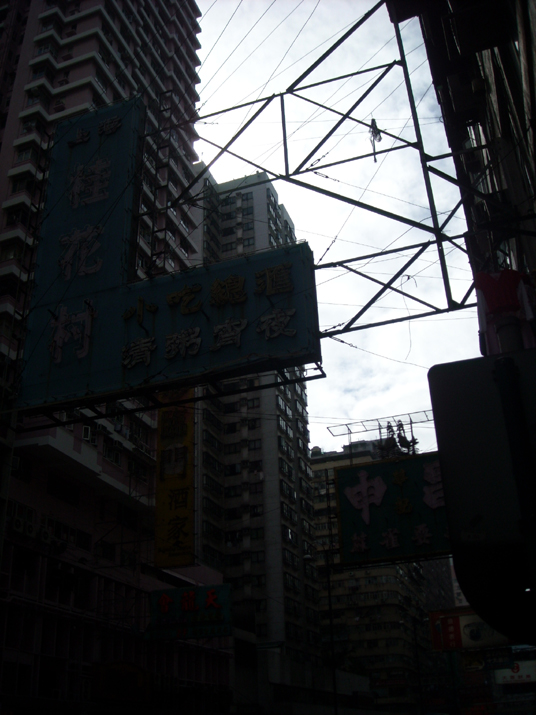}}\vspace{-0.03in}
\centerline{}
\end{minipage}\hspace{0.15in}
\begin{minipage}{0.2\linewidth}
\centerline{\includegraphics[width=0.86in]{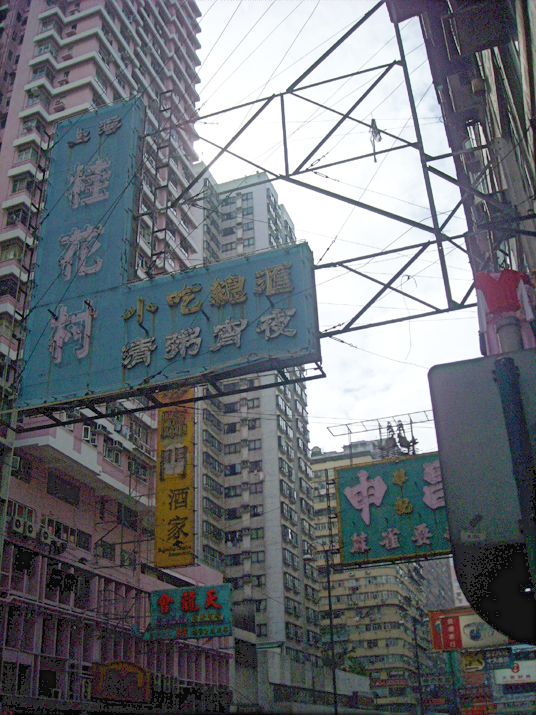}}\vspace{-0.03in}
\centerline{}
\end{minipage}\hspace{0.15in}
\begin{minipage}{0.2\linewidth}
\centerline{\includegraphics[width=0.86in]{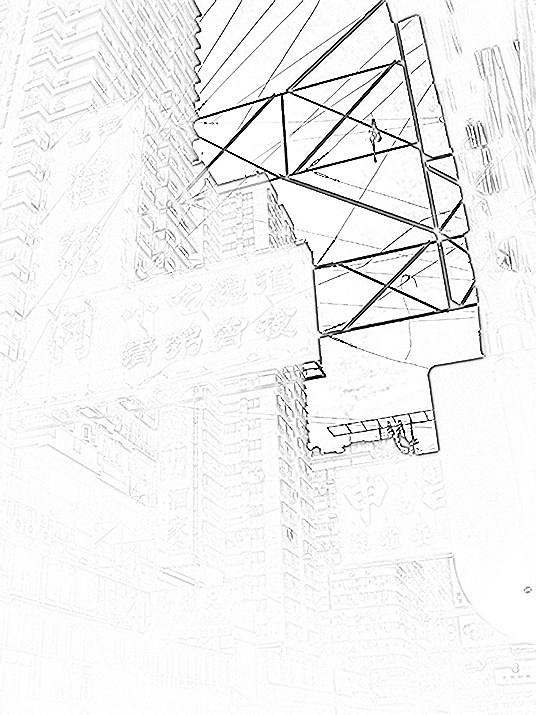}}\vspace{-0.03in}
\centerline{}
\end{minipage}\hspace{0.15in}
\begin{minipage}{0.2\linewidth}
\centerline{\includegraphics[width=0.86in]{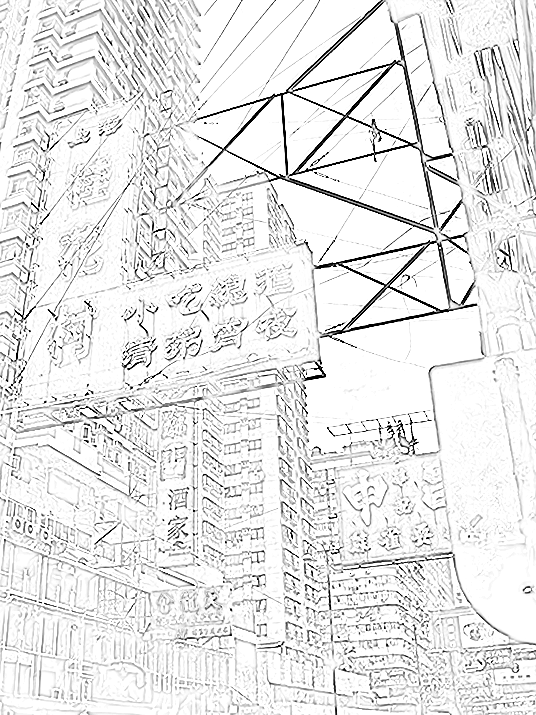}}\vspace{-0.03in}
\centerline{}
\end{minipage}
\caption{\y Visual comparison of different edge extraction methods in non-reference Fusion dataset \cite{Chen2018Retinex}. The first column is low-light images; the second column is our enhanced results; the third and the last columns are the edges obtained by using Eq. \eqref{eq3} and the learned edge by our strategy, respectively. }
\label{fig:edge1}
\end{figure}

\begin{figure}[b!]
\setlength{\abovecaptionskip}{0in}
\subfigcapskip=-0.05in
\subfigure[\tiny{(a) Input (9.13/36.84)
}]{
\zoomincludgraphic{0.80in}{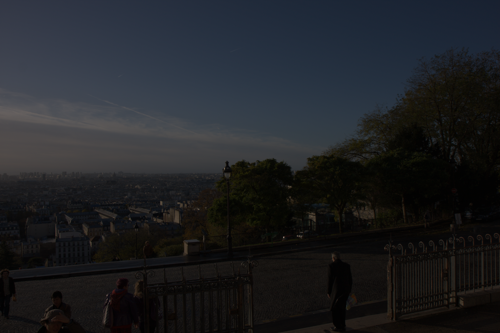}{0.79}{0.06}{0.89}{0.23}{2.5}{help_grid_off}{up_left}{line_connection_off}{2}{blue}{1}{red}}\hspace{-0.29in}
\subfigure[\tiny{(b) SMG (19.61/81.57)}]{
\zoomincludgraphic{0.80in}{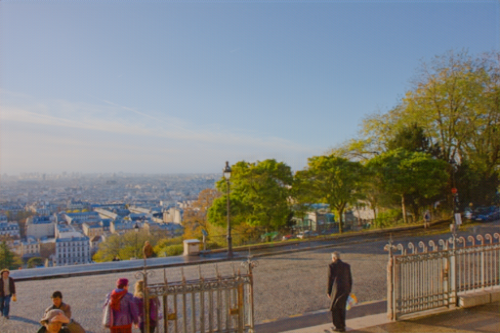}{0.79}{0.06}{0.89}{0.23}{2.5}{help_grid_off}{up_left}{line_connection_off}{2}{blue}{1}{red} }
\hspace{-0.38in}
\subfigure[\tiny{(c) Ours (22.94/87.68)}]{
\zoomincludgraphic{0.80in}{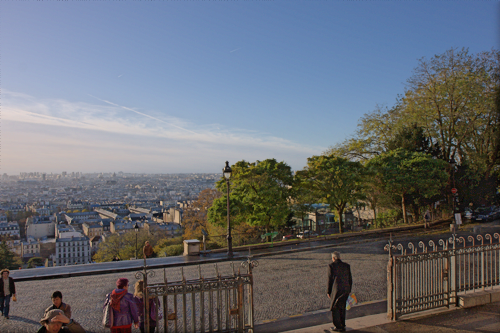}{0.79}{0.06}{0.89}{0.23}{2.5}{help_grid_off}{up_left}{line_connection_off}{2}{blue}{1}{red} }
\hspace{-0.38in}
\subfigure[\tiny{(d) GT }]{
\zoomincludgraphic{0.80in}{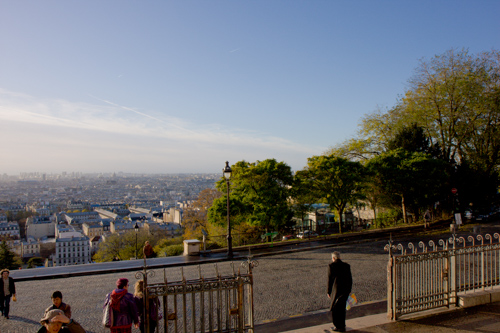}{0.79}{0.06}{0.89}{0.23}{2.5}{help_grid_off}{up_left}{line_connection_off}{2}{blue}{1}{red} }

\subfigure[\tiny{(e) edge of (a)
}]{
\zoomincludgraphic{0.80in}{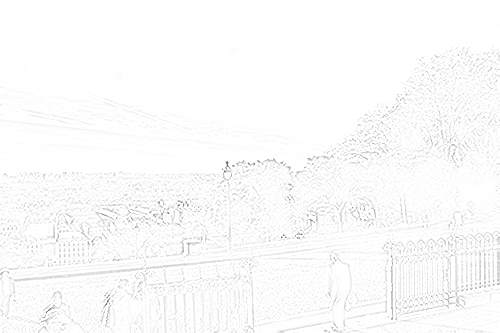}{0.79}{0.06}{0.89}{0.23}{2.5}{help_grid_off}{up_left}{line_connection_off}{2}{blue}{1}{red}}\hspace{-0.29in}
\subfigure[\tiny{(f) SMG}]{
\zoomincludgraphic{0.80in}{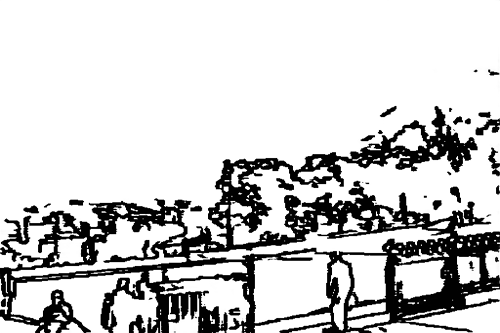}{0.79}{0.06}{0.89}{0.23}{2.5}{help_grid_off}{up_left}{line_connection_off}{2}{blue}{1}{red} }
\hspace{-0.38in}
\subfigure[\tiny{(g) Ours}]{
\zoomincludgraphic{0.80in}{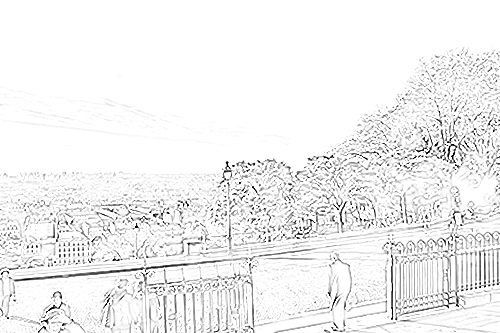}{0.79}{0.06}{0.89}{0.23}{2.5}{help_grid_off}{up_left}{line_connection_off}{2}{blue}{1}{red} }
\hspace{-0.38in}
\subfigure[\tiny{(h) GT }]{
\zoomincludgraphic{0.80in}{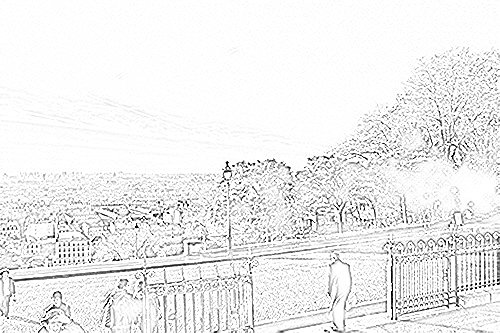}{0.79}{0.06}{0.89}{0.23}{2.5}{help_grid_off}{up_left}{line_connection_off}{2}{blue}{1}{red} }

\subfigure[\tiny{(a) Input (12.77/67.42)
}]{
\zoomincludgraphic{0.80in}{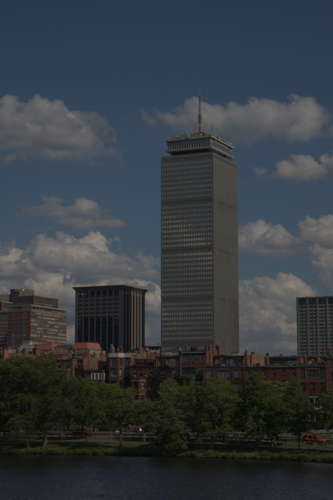}{0.48}{0.32}{0.63}{0.39}{2.5}{help_grid_off}{up_left}{line_connection_off}{2}{blue}{1}{red}}\hspace{-0.29in}
\subfigure[\tiny{(b) SMG (18.30/83.93)}]{
\zoomincludgraphic{0.80in}{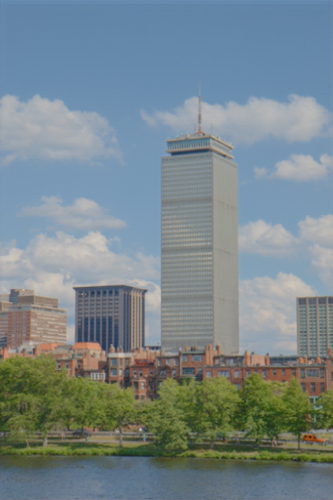}{0.48}{0.32}{0.63}{0.39}{2.5}{help_grid_off}{up_left}{line_connection_off}{2}{blue}{1}{red} }
\hspace{-0.38in}
\subfigure[\tiny{(c) Ours (21.12/91.16)}]{
\zoomincludgraphic{0.80in}{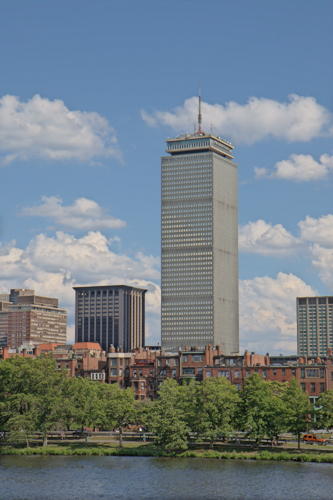}{0.48}{0.32}{0.63}{0.39}{2.5}{help_grid_off}{up_left}{line_connection_off}{2}{blue}{1}{red} }
\hspace{-0.38in}
\subfigure[\tiny{(d) GT }]{
\zoomincludgraphic{0.80in}{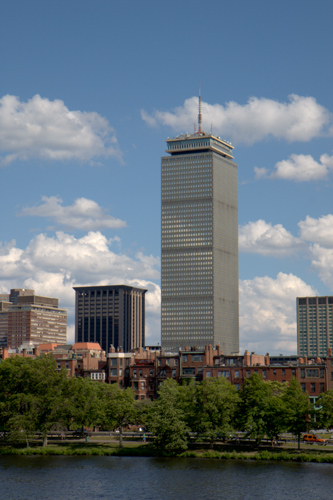}{0.48}{0.32}{0.63}{0.39}{2.5}{help_grid_off}{up_left}{line_connection_off}{2}{blue}{1}{red} }

\subfigure[\tiny{(e) edge of (a)
}]{
\zoomincludgraphic{0.80in}{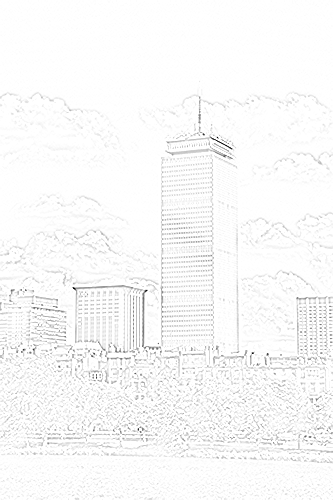}{0.48}{0.32}{0.63}{0.39}{2.5}{help_grid_off}{up_left}{line_connection_off}{2}{blue}{1}{red}}\hspace{-0.29in}
\subfigure[\tiny{(f) SMG }]{
\zoomincludgraphic{0.80in}{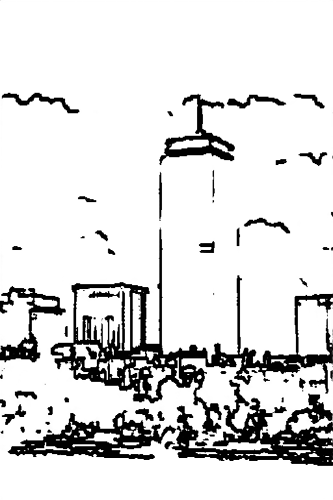}{0.48}{0.32}{0.63}{0.39}{2.5}{help_grid_off}{up_left}{line_connection_off}{2}{blue}{1}{red} }
\hspace{-0.38in}
\subfigure[\tiny{(g) Ours}]{
\zoomincludgraphic{0.80in}{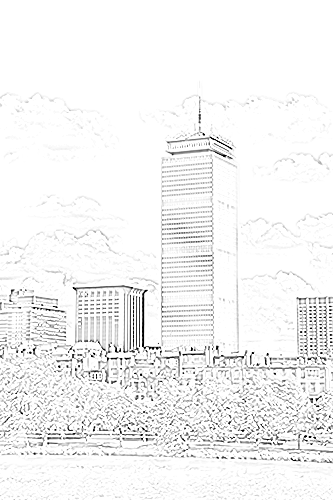}{0.48}{0.32}{0.63}{0.39}{2.5}{help_grid_off}{up_left}{line_connection_off}{2}{blue}{1}{red} }
\hspace{-0.38in}
\subfigure[\tiny{(d) GT }]{
\zoomincludgraphic{0.80in}{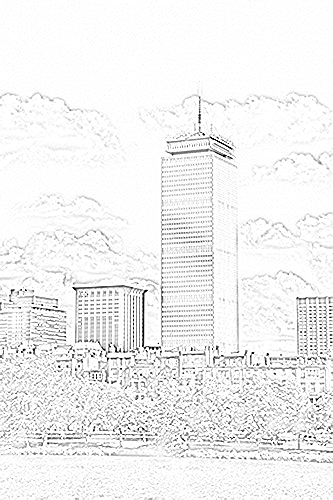}{0.48}{0.32}{0.63}{0.39}{2.5}{help_grid_off}{up_left}{line_connection_off}{2}{blue}{1}{red} }
\caption{\y Image enhancement results with index PSNR (dB) and SSIM (\%). (a) is the input low-light image from dataset Fivek \cite{bychkovsky2011learning}; (b) and (c) are enhancement results by SMG \cite{xu2023low} and ours; (d) is the ground truth; (e) is the edge of (a) extracted by the Laplace operator; (f) and (g) are corresponding edges generated by SMG \cite{xu2023low} and ours; (h) is the ground truth.}\label{Fig_edge_fivek}
\end{figure}

\section{Numerical experiments}\label{experiment}
To evaluate the effectiveness of our proposed scheme, we first analyze the proposed edge extractor using some ablation studies. Secondly, the parameters of the proposed model and algorithm are studied. Then, the synthesis and real-world low-light datasets are tested to illustrate the robustness of our method. Moreover, some state-of-the-art methods are compared to demonstrate the superiority of the proposed method.  

\subsection{Implement}
In this work, we apply Eq. \eqref{eq3} on 485 training images of the LOL dataset \cite{Chen2018Retinex} to obtain their corresponding edges. As illustrated in Fig. \ref{fig:retinex}, we extract the edge label from the high-contrast image. 
Our edge extractor is implemented using the PyTorch framework, and we update it using the Adam optimizer. The initial learning rate is set to $10^{-4}$ and halved every $200$ epochs. Once the edge network is learned, we integrate it into the proposed framework to enhance the performance of the image Retinex model.

\begin{figure}[t!]
\setlength{\abovecaptionskip}{0in}
    \centering
    \begin{minipage}{0.28\linewidth}\centerline{\includegraphics[width=2in]{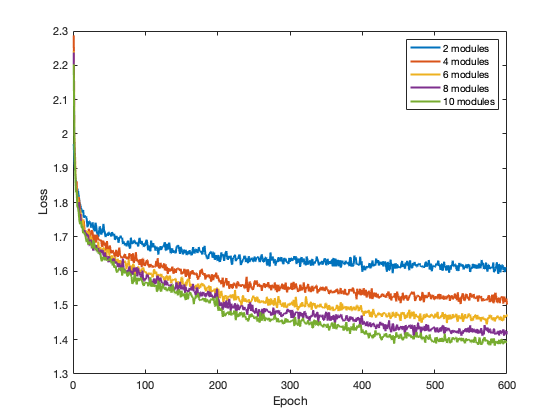}}
    \centerline{}
    \end{minipage}\hspace{0.8in}
    \begin{minipage}{0.28\linewidth}\centerline{\includegraphics[width=2in]{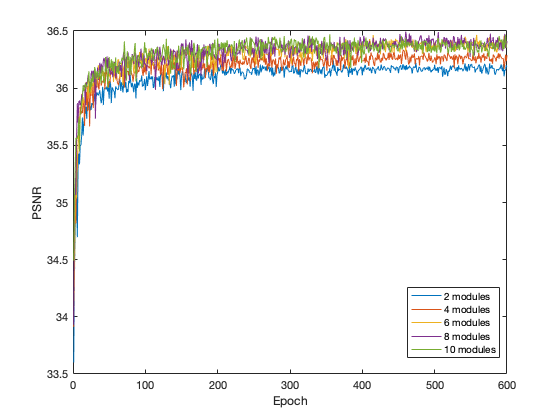}}
    \centerline{}
    \end{minipage}   
    \caption{Study on the number of multi-scale residual block
(MSRB) modules. Considering the cost and the performance, we set the number of modules as $6$ in our experiment. }
    \label{fig:loss}
\end{figure}

\begin{figure}[b!]
   \setlength{\abovecaptionskip}{0in}
    \centering
    \begin{minipage}{0.27\linewidth}\centerline{\includegraphics[width=1.9in]{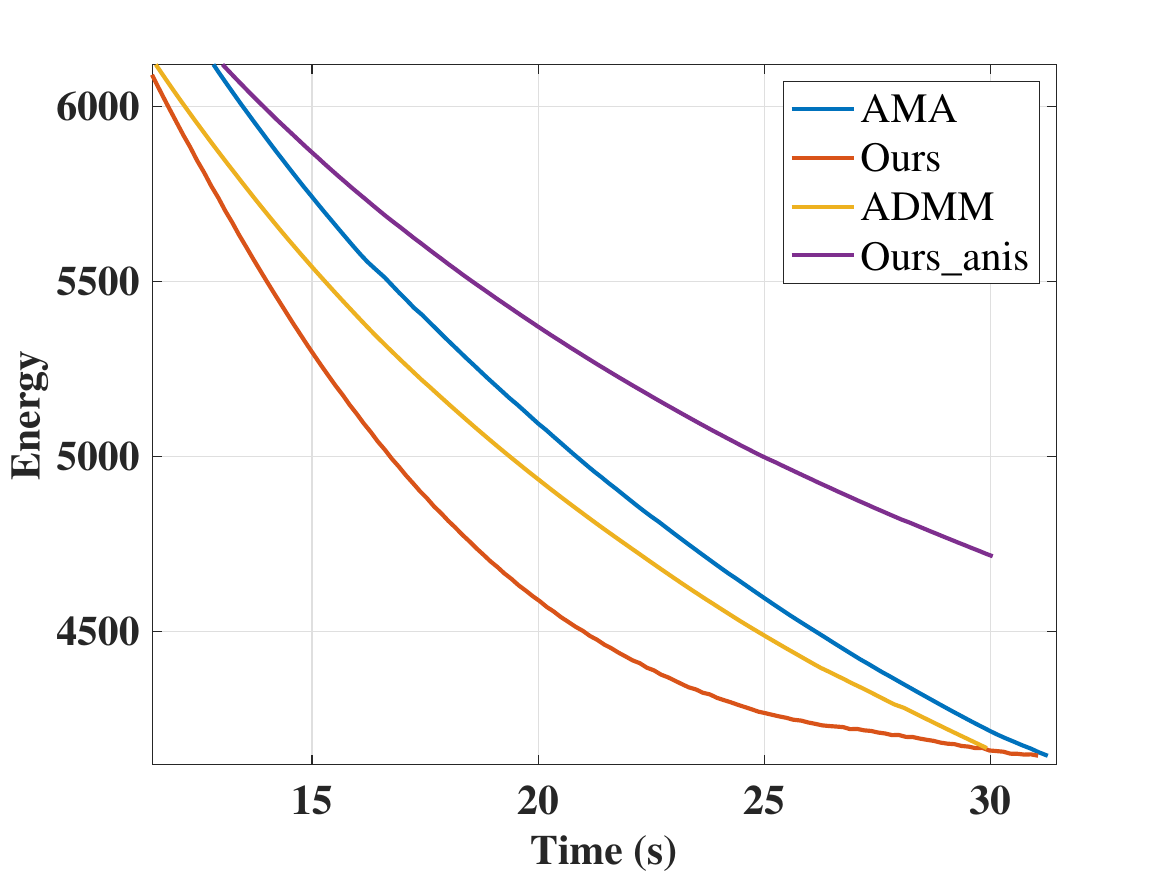}}
    \centerline{}
    \end{minipage}\hspace{0.8in}
    \begin{minipage}{0.27\linewidth}\centerline{\includegraphics[width=1.9in]{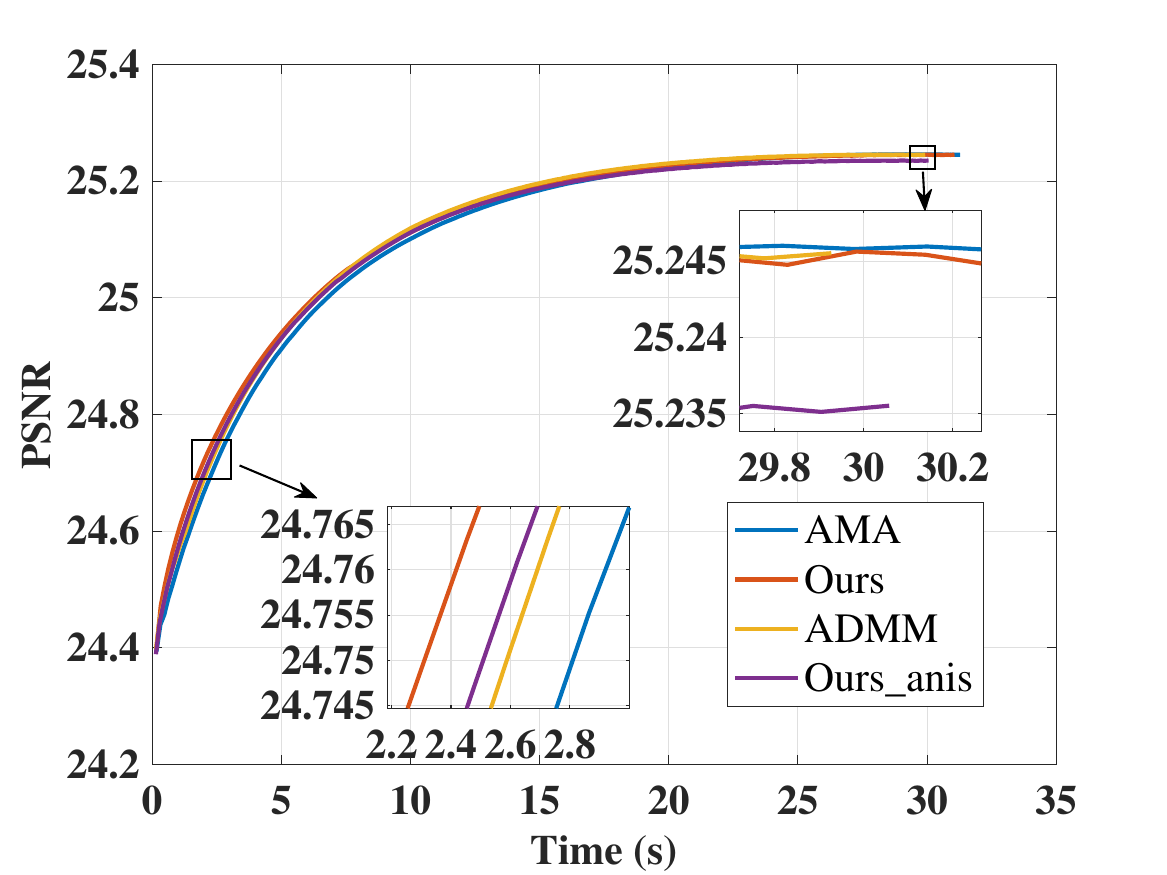}}
    \centerline{}
    \end{minipage}
    \caption{\y Comparison of the AMA, ADMM, the proposed scheme, and the proposed scheme with anisotropic TV \eqref{anis}, referred to as Ours\_anis.}
    \label{fig:energy}
\end{figure}

\subsection{Effectiveness of edge extractor}
We propose a simple but effective edge extractor, which is comprised of the convolutional layer, multi-scale residual block (MSRB), upsample module, and
bottleneck layer. 
{\y To illustrate the effectiveness of the proposed edge extractor, Fig. \ref{fig:edge1} presents the visual comparison of the learned edges and input edges of the non-reference Fusion dataset \cite{Chen2018Retinex}. For better visualization, all edges are presented by using ``$1-\mbox{G(S)}$''. 
The enhancement results of our proposed scheme are also provided. One can recognize the detailed structure from our enhanced images easily. Besides, we also present the comparison of the paired FiveK dataset \cite{bychkovsky2011learning} in} {\y Fig. \ref{Fig_edge_fivek}. Compared to the proposed scheme, the edge-based deep learning method SMG \cite{xu2023low} fails to learn the detailed structure of low-light images. While our learned edge is closer to the ground truth, and our enhanced image also approaches the ground truth. 
}
For the training, we study the effects of the number of MSRB modules. Theoretically, as the number of MSRB modules increases, the performance of the edge extractor gets better. In Fig. \ref{fig:loss}, the loss and PSNR curves show that the learned edges are closer to the labels with $6$ MSRB modules. 
Considering the cost and the performance, in the experiments, we set the number of MSRB as $6$ to capture the edge information directly from the input low-light images.  
{\y 
It takes about $8.5$ hours to train the edge extractor with $6$ MSRB modules on an NVIDIA RTX A6000 GPU. 
}

\begin{figure}[t!]
\setlength{\abovecaptionskip}{0in}
    \centering
    \begin{minipage}{0.28\linewidth}\centerline{\includegraphics[width=2in]{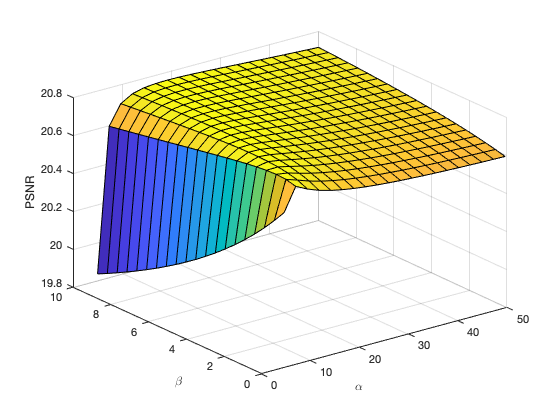}}
    \centerline{}
    \end{minipage}
    \hspace{0.8in}
    \begin{minipage}{0.28\linewidth}\centerline{\includegraphics[width=2in]{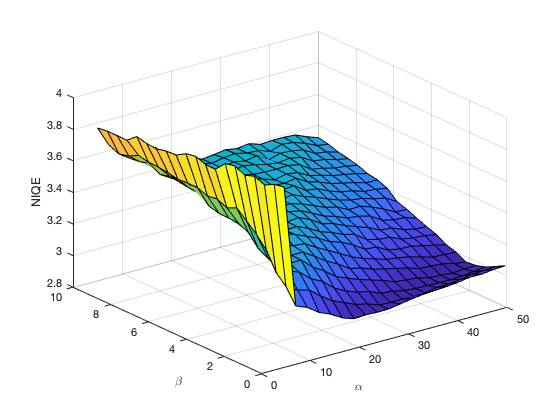}}
    \centerline{}
    \end{minipage}
    \caption{PSNR and NIQE results along with model parameters $\alpha$ and $\beta$. }
    \label{fig:modelpar}
\end{figure}

\begin{figure}[t!]
\setlength{\abovecaptionskip}{0in}
    \centering
    \begin{minipage}{0.28\linewidth}\centerline{\includegraphics[width=1.8in]{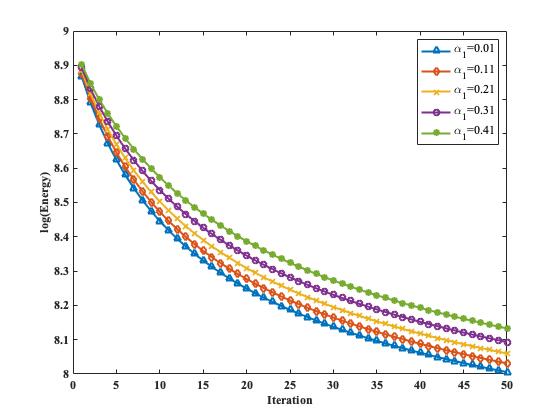}}
    \centerline{}
    \end{minipage}
    \hspace{0.6in}
    \begin{minipage}{0.28\linewidth}\centerline{\includegraphics[width=1.8in]{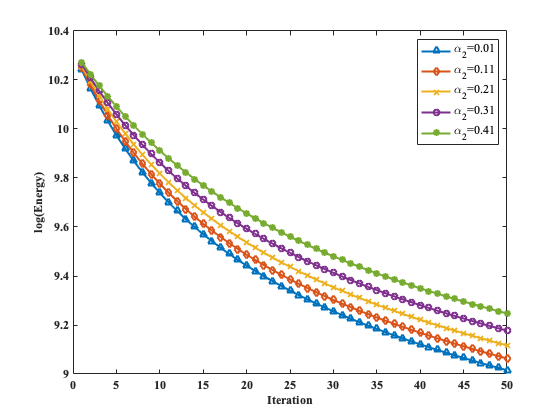}}
    \centerline{}
    \end{minipage}
    \caption{Energy with log of algorithm parameters $\alpha_1$ and $\alpha_2$. }
    \label{fig:algpar1}
\end{figure}

\begin{figure}[t!]
\setlength{\abovecaptionskip}{0in}
    \centering
    % \hspace{0.3in}
    \begin{minipage}{0.28\linewidth}\centerline{\includegraphics[width=1.8in]{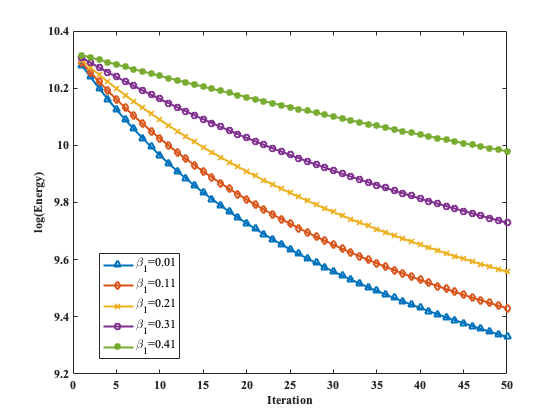}}
    \centerline{}
    \end{minipage} \hspace{0.6in}
    \begin{minipage}{0.28\linewidth}\centerline{\includegraphics[width=1.8in]{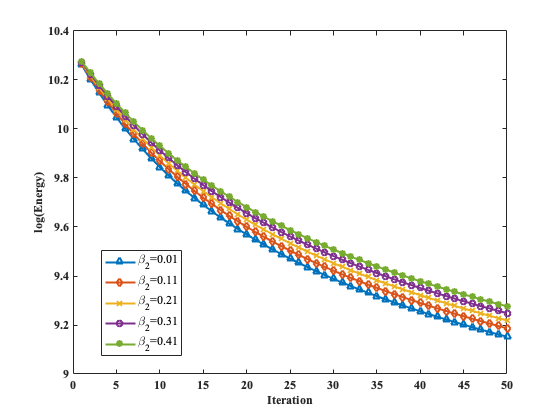}}
    \centerline{}
    \end{minipage}
    \caption{Energy with log of algorithm parameters $\beta_1$ and $\beta_2$. }
    \label{fig:algpar2}
\end{figure}

\subsection{Algorithm and model analysis}
In this subsection, we analyze the efficiency of the proposed model \eqref{model} with ROF regularizer \eqref{rof}, i.e.,  $\Psi(l)=\Vert\nabla l\Vert_1$, referred to as Ours. 
{\y First of all, we compare the proposed algorithm with the alternating minimization algorithm (AMA) and ADMM. 
The energy and PSNR curves presented in Fig. \ref{fig:energy} illustrate that with the same model parameter setting, the proposed algorithm converges faster and more accurately.
Moreover, we also compare our model with the anisotropic TV \eqref{anis}, referred to as Ours\_{anis}. This implies that the proposed algorithm can handle different regularizers robustly. 
}  

Secondly, the algorithm parameters $\alpha_1$, $\alpha_2$, $\beta_1$, $\beta_2$, and $\tau_1$, $\tau_2$ are studied. According to the Assumption \ref{ass3}, $\alpha_i^k\in[0,\bar\alpha_i]$ with $0 < \bar\alpha_i < \frac{(1-\varepsilon)\rho_i}{2}$ and $\beta_i^k\in[0,\bar\beta_i]$ for some $\bar\beta_i\in (0,1)$, $i=1,2$. In experiments, we set strongly convex modulus $\rho_1$ and $\rho_2$ to be $1$, then we have $\alpha_i\in[0,0.5)$ and $\beta_i\in[0,1)$.

On the other hand, according to Assumption \ref{ass2}, $\lambda_1^+$ and $\lambda_2^+$ are set to be the supremum of $L_1^-(x_2^k)$ and $L_2^-(x_1^k)$, respectively. Note that $L_1^{-}\left(x_2^k\right)$ and $L_2^{-}\left(x_1^{k}\right)$ are the Lipschitz constants of $\nabla_{x_1} h^{-}\left(x_1, x_2^k\right)$ at $x_1^k$ and $\nabla_{x_2} h^{-}\left(x_1^{k}, x_2\right)$ at $x_2^k$, respectively. 
Since in the proposed model \eqref{model}, $h^-(x_1,x_2)=\frac{\beta}{2}\Vert x_2-s-x_1\Vert^2$, we have 
$\Vert \nabla_{x_1}h^-(u,x_2)-\nabla_{x_1}h^-(v,x_2) \Vert=\beta\Vert u-v \Vert\leq L_1^-(x_2)\Vert u-v \Vert$, hence $L_1^-(x_2)\in[0,\beta]$. Similarly, we have $L_2^-(x_1)\in[0,\beta]$. To this end, we know that for the proposed Retinex model \eqref{model}, $\lambda_1^+$ and $\lambda_2^+$ are set to be $\beta$. 

In the experiments, by setting $\bar \beta_i=1-\epsilon$, $\bar \alpha_i=\frac{1}{2}-\epsilon$, $\rho_i=1$, and $\lambda_i^+=\beta$, we get $\tau_i=\frac{\frac{1+\epsilon}{\epsilon}(2.5+\beta_i-2\epsilon)\beta}{1-\alpha_i}$.  
Since $\alpha_i$ and $\beta_i$ are extrapolation parameters, we first fix them as $0$ to find $\tau_i$. As analyzed before, $\tau_i$ can be decided by $\alpha_i$, $\beta_i$, and $\beta$. The remaining algorithm parameters are $\tau_i$, $i=1,2$, and decided by model parameter $\beta$. Hence, we first study the model parameters $\alpha$ and $\beta$ with fixed $\alpha_i$ and $\beta_i$. After the optimized $\alpha$ and $\beta$ are obtained, we will focus on the algorithm parameters $\alpha_i$ and $\beta_i$. 
% Thirdly, the model parameters $\alpha$ and $\beta$ are studied. 
We first study $\alpha$ and $\beta$ in interval $[1, 200]$ with stepsize $10$. We find that our model is quite robust with parameter $\beta$ while the curve is turbulent alone $\alpha$. Hence we narrow the interval to $[5, 50]$ and $[0.1, 10]$ use the linespace to find $20$ numbers, respectively. 
Fig. \ref{fig:modelpar} plots the PSNR and NIQE results along the model parameters $\alpha\in[5, 50]$ and $\beta\in[0.1, 10]$. 
The surface indicates that the proposed model is stable to parameters $\alpha$ and $\beta$ in the given interval. Since curves become quite flat as the parameter $\alpha$ from $3$ to $5$ along the $\alpha$ axis, which implies choosing $\alpha$ in the interval $[10, 40]$ ensures the best result. Similarly, the interval of model parameter $\beta$ is chosen at $[0.1, 5]$.

With the obtained optimized model parameter $\beta$, the algorithm parameter $\tau_i$ then controlled by $\alpha_i$ and $\beta_i$, $i=1, 2$. Hence, only $\alpha_i\in[0, 0.5)$ and $\beta_i\in[0, 1)$ are studied in the algorithm parameter. For these four algorithm parameters, we choose to tune one parameter in its interval while others are fixed as $0$. More specifically, we set the model parameters $\alpha_i\in[0.01: 0.1: 0.5]$ and $\beta_i\in[0.01: 0.1: 1]$. From Fig. \ref{fig:algpar1} and Fig. \ref{fig:algpar2}, we know that extrapolation parameters speed up the whole algorithm with convergence.

\begin{table*}[t!]
    \centering
    \caption{\y Quantitative results with PSNR (dB), SSIM (\%), NIQE, FADE, and SSEQ values on datasets Fivek, LSRW, and SYD. 
    The \rr{BEST} result is in red whereas the \rb{SECOND BEST} one is in blue. 
    }
    \label{tab:realblur_paried}
    \resizebox{0.95\hsize}{!}{
    \tabcolsep=0.in
    \begin{tabular}{c|ccccccccccccc}
    \hline
        ~Dataset~&~Index~~& ~~Input~~~ & RetinexNet \cite{Chen2018Retinex}& ~GLADNet \cite{wang2018gladnet}~~&URetinex \cite{wu2022uretinex}&~~EFINet \cite{liu2022efinet}~~~&~~~~RQ \cite{liu2023low}~~~~& ~~~SMG \cite{xu2023low}~~~ & ~~Bread \cite{guo2023low}~~~& ~~~Ours~~~ \\ \hline
        \multirow{5}{*}{Fivek} %600 images 
        &PSNR&13.42&12.59&16.46&13.22&\rb{17.24}&14.79&16.31&15.94&\rr{17.82}\\
        &SSIM&55.24&59.96&65.98&62.18&\rb{68.27}&55.10&65.14&65.28&\rr{70.33}\\
        &NIQE&3.832&3.860&\rb{3.450}&3.700&3.655&\rr{3.405}&7.704&4.097&3.761\\
        &FADE&1.922&\rr{1.731}&2.275&2.335&1.922&\rb{1.762}&2.023&2.105&1.799\\
        &SSEQ&26.28&25.16&27.03&28.11&26.26&35.41&21.38&39.26&\rr{19.95}\\
        \hline
        \multirow{5}{*}{LSRW}
        &PSNR&9.16&15.48&\rb{16.64}&15.99&15.29&10.64&16.47&16.34&\rr{16.74}\\
        &SSIM&19.09&48.61&52.92&52.50&53.69&29.44&58.20&\rb{59.60}&\rr{62.72}\\
        &NIQE&3.278&4.312&3.364&3.804&\rr{3.035}&3.531&6.702&3.443&\rb{3.142}\\
        &FADE&1.636&\rb{1.618}&1.974&2.251&1.929&2.271&1.802&2.435&\rr{1.432}\\
        &SSEQ&27.22&27.74&17.80&19.54&30.96&31.09&\rr{13.05}&33.79&\rb{16.17}\\
        \hline
        \multirow{5}{*}{SYD}
        &PSNR&9.13&14.42&\rb{16.72}&16.57&14.78&16.64&16.18&14.30&\rr{16.82}\\
        &SSIM&30.33&68.41&\rb{70.50}&69.97&64.74&\rr{70.72}&68.12&68.00&66.06\\
        &NIQE&4.503&6.211&4.245&4.252&4.043&4.121&5.348&4.128&5.209\\
        &FADE&0.951&\rb{0.894}&1.072&0.961&1.200&1.055&1.243&1.172&\rr{0.873}\\
        &SSEQ&15.45&20.181&15.39&16.91&16.41&23.09&\rr{12.68}&20.91&\rb{15.36}\\
        \hline
    \end{tabular}}
\end{table*}

\subsection{Image enhancement}
This subsection compares our method with some state-of-the-art image enhancement models, including the traditional methods (LR3M \cite{9056796}, QSPE \cite{huang2022quaternion}, 
DRE \cite{jia2024low}, RMR \cite{jia2024variational})
and the deep learning methods
(RetinexNet \cite{Chen2018Retinex}, {\y GLADNet \cite{wang2018gladnet},  
URetinex \cite{wu2022uretinex}, EFINet \cite{liu2022efinet}, 
RQ \cite{liu2023low}, SMG \cite{xu2023low}, Bread \cite{guo2023low}}). 
{\y
Note that LR3M, DRE, RMR, RetinexNet, and URetinex are the methods under the Retinex framework. 
In particular, SMG introduces an end-to-end neural network that leverages edge information to assist with low-light image enhancement.
}

\begin{table}[t!]
    \centering
    \caption{Quantitative results with PSNR (dB) and SSIM (\%) values on dataset Realblur15. 
    The \rr{BEST} result is in red whereas the \rb{SECOND BEST} one is in blue. 
    }
    \label{tab:realblur}
    \resizebox{0.95\hsize}{!}{
    \tabcolsep=0.0in
    \begin{tabular}{c|c|ccccccccccccc}
    \hline
         Image &Index~&~Input~ & RetiNet  & URetinex  & ~EFINet~ & ~RQ~  & ~~LR3M~~ & ~~QSPE~~  & ~~DRE~~  & ~~RWR~~  & ~~Ours~~ \\ \hline
        \multirow{2}{*}{Img1} &PSNR& 8.91 &16.93& 19.69 & 16.60 & 16.56 & 15.70 & 12.53 & 16.35 & \rb{19.92} & \rr{20.36} \\ 
        &SSIM&24.66 &70.78 & \rb{86.65}&83.72&79.48&79.26&67.27&77.80&84.98&\rr{87.23}\\
        \multirow{2}{*}{Img2} &PSNR& 8.47 & 14.41& \rr{25.16} & 15.36 & 13.12 & 14.33 & 11.52 & 15.35 & \rb{19.13} & 18.61 \\ 
        &SSIM&29.65&68.72&\rr{87.29}&79.33&64.81&73.68&68.03&70.98&83.52&\rb{85.96}\\
        \multirow{2}{*}{Img3} &PSNR& 9.49 & 16.82 & \rr{23.76} & 16.34 & 15.36 & 15.98 & 13.22 & 16.75 & 17.85 & \rb{20.33} \\ 
        &SSIM&23.97&68.97&\rr{88.83}&80.38&76.21&79.34&67.77&82.22&80.16&\rb{87.02}\\
        \multirow{2}{*}{Img4} &PSNR& 13.09 & 15.07 & 13.63 & \rr{21.76} & 13.56 & 19.19 & 15.83 & 13.99 & 15.85 & \rb{19.26} \\ 
        &SSIM&23.90&44.98&58.00&\rr{82.39}&57.15&\rb{72.36}&53.78&51.71&59.85&68.01\\
        \multirow{2}{*}{Img5} &PSNR& 10.16 & 15.25 & 17.13 & 16.14 & 17.16 & \rb{18.80} & 13.29 & 14.83 & 14.14 & \rr{19.01} \\ 
        &SSIM&23.81&72.64&81.59&86.83&81.58&\rb{87.60}&61.29&81.71&75.39&\rr{88.26}\\
        \multirow{2}{*}{Img6} &PSNR& 6.71 & 14.01 & \rr{21.61} & 14.82 & 10.72 & 12.20 & 9.22 & 11.71 & 15.05 & \rb{15.98} \\ 
        &SSIM&34.67&64.51&\rr{88.86}&74.78&63.95&70.66&62.39&66.63&\rb{83.54}&83.07\\
        \multirow{2}{*}{Img7} &PSNR& 10.44 & 15.28 & 15.56 & 19.54 & \rr{25.13} & 14.09 & 22.66 & \rb{24.97} & 13.97 & 21.84 \\ 
        &SSIM&20.75&59.45&76.64&\rr{88.35}&86.08&72.14&75.05&67.55&69.18&\rb{86.77}\\
        \multirow{2}{*}{Img8} &PSNR& 8.56 & 17.30 & 18.74 & 16.78 & \rr{26.03} & 14.78 & 19.41 & \rb{23.02} & 11.32 & 22.75 \\ 
        &SSIM&21.37&79.30&89.19&89.19&\rb{92.00}&86.82&59.35&87.93&53.43&\rr{95.74}\\
        \multirow{2}{*}{Img9} &PSNR& 10.38 & 16.05 & 14.82 & 18.56 & \rr{22.70} & 18.23 & \rb{21.62} & 19.20 & 13.91 & 21.51 \\       &SSIM&19.08&68.07&75.70&\rb{87.78}&\rr{88.32}&\rr{88.32}&62.47&70.80&67.87&88.01\\
        \multirow{2}{*}{Img10} &PSNR& 10.43 & 15.82 & 14.95 & \rb{20.89} & 18.04 & 15.33 & 20.38 & 17.12 & 13.64 & \rr{22.80} \\ 
        &SSIM&21.86&69.82&81.26&\rr{93.46}&83.16&80.19&50.57&83.73&58.81&\rb{91.31}\\
        \multirow{2}{*}{Img11} &PSNR& 10.58 & 14.66 & 14.89 & 18.06 & \rr{21.91} & 15.97 & 16.69 & 18.70 & 17.75 & \rb{19.86} \\ 
        &SSIM&24.15&57.93&70.24&83.65&\rr{88.94}&81.08&76.29&72.02&\rb{86.43}&83.31\\
        \multirow{2}{*}{Img12} &PSNR& 10.90 & 16.69 & 14.17 & 18.64 & \rr{25.52} & 19.39 & 17.51 & \rb{21.27} & 13.74 & 20.54 \\ 
        &SSIM&22.56&58.48&70.69&\rb{86.78}&\rr{89.80}&79.47&50.01&65.21&62.64&79.87\\
        \multirow{2}{*}{Img13} &PSNR& 11.87 & 15.57 & 14.03 & 20.33 & 16.95 & 21.76 & 17.54 & \rr{23.34} & 14.89 & \rb{22.94} \\ 
        &SSIM&15.96&63.84&73.77&\rr{87.34}&76.62&83.36&46.19&81.98&50.67&\rb{85.86}\\
        \multirow{2}{*}{Img14} &PSNR& 9.93 & 14.44 & 18.84 & 17.28 & 17.10 & 16.68 & 14.87 & 17.49 & \rr{20.39} & \rb{18.85} \\ 
        &SSIM&26.65&56.29&77.12&\rb{80.79}&76.08&76.56&65.58&76.38&\rr{82.60}&77.55\\
        \multirow{2}{*}{Img15} &PSNR& 9.30 & 14.86 & \rb{17.67} & 16.29 & 16.05 & 15.99 & 12.36 & 15.42 & 16.25 & \rr{18.55} \\
        &SSIM& 27.24&63.73&\rb{79.25}&83.10&78.05& 78.36&65.64&78.02&77.67&\rr{84.29}\\\hline
        \multirow{2}{*}{Ave.} &PSNR&9.95&	15.54&		17.64	&17.83	&	\rb{18.39}&		16.56	&		15.91&		17.97&	15.85&	\rr{20.21}\\ 
        &SSIM&24.02&64.50&79.00&\rb{84.52}&78.81&79.28&62.11&74.31&71.79&\rr{84.82}\\
        \hline
    \end{tabular}}
\end{table}

{\y Three paired datasets Fivek \cite{bychkovsky2011learning}, LSRW \cite{hai2023r2rnet}, and SYD \cite{lv2021attention} are tested to verify the effectiveness of the proposed method. We report average numerical image enhancement results of $600$ images in Fivek, $965$ images in SYD, and $50$ images in LSRW with PSNR, SSIM, NIQE, FADE, and SSEQ indexes in Table \ref{tab:realblur_paried}.}
Besides, $15$ low-light images in the dataset RealBlur \cite{rim2020real} with paired bright-light images are tested to demonstrate the validity of the proposed approach. Here we indicate these real low-light images as RealBlur15. 
We present the PSNR and SSIM results of dataset Realblur15 in Table \ref{tab:realblur}. 
The best and the second best values are marked as red and blue, respectively. From numerical results, the proposed method with the learned prior is better than the methods with traditional priors in most images. While comparing to the deep learning methods, some results fail to achieve the best one, most of our results still get the second best index. 
Our method achieves the best averaged overall performance among all the methods, which shows the effectiveness of our method on real low-light images. 

\begin{figure*}[t!]
\setlength{\abovecaptionskip}{0in}
\subfigcapskip=-0.05in
\subfigure[\scriptsize{(a) Input (7.27/32.59)}]{
\zoomincludgraphic{1.33in}{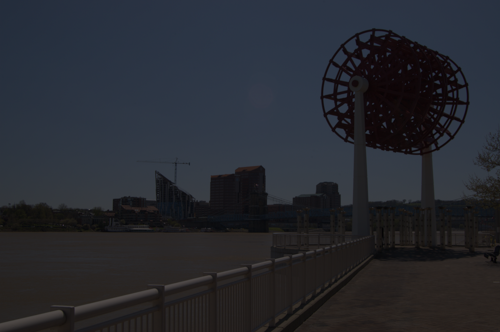}{0.4}{0.3}{0.55}{0.53}{2}{help_grid_off}{up_left}{line_connection_off}{2}{blue}{1}{red}}\hspace{-0.25in}
\subfigure[\scriptsize{(b) RetinexNet (15.89/84.70)}]{
\zoomincludgraphic{1.33in}{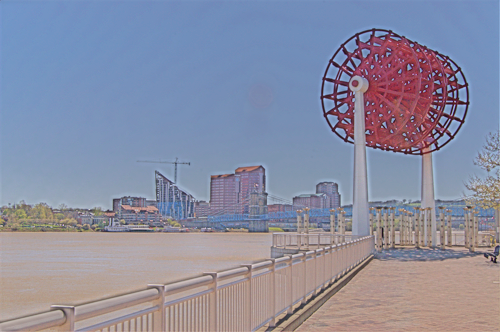}{0.4}{0.3}{0.55}{0.53}{2}{help_grid_off}{up_left}{line_connection_off}{2}{blue}{1}{red} }\hspace{-0.3in}
\subfigure[\scriptsize{(c) GLADNet (20.16/79.30)}]{
\zoomincludgraphic{1.33in}{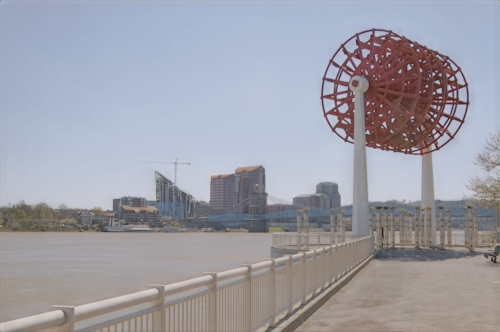}{0.4}{0.3}{0.55}{0.53}{2}{help_grid_off}{up_left}{line_connection_off}{2}{blue}{1}{red} }
\hspace{-0.35in}
\subfigure[\scriptsize{(d) URetinexNet (19.07/83.12)}]{
\zoomincludgraphic{1.33in}{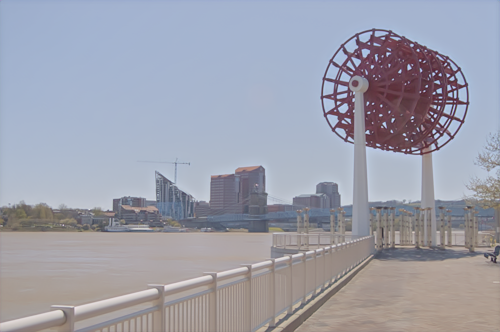}{0.4}{0.3}{0.55}{0.53}{2}{help_grid_off}{up_left}{line_connection_off}{2}{blue}{1}{red} }
\hspace{-0.35in}
\subfigure[\scriptsize{(e) EFINet (19.43/89.01)}]{
\zoomincludgraphic{1.33in}{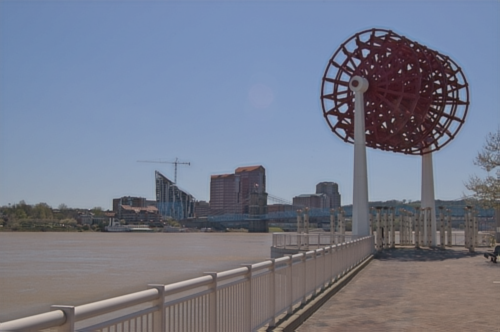}{0.4}{0.3}{0.55}{0.53}{2}{help_grid_off}{up_left}{line_connection_off}{2}{blue}{1}{red} }
\hspace{-0.3in}\vspace{-0.1in}

\subfigure[\scriptsize{(f) RQ (21.51/76.72)}]{
\zoomincludgraphic{1.33in}{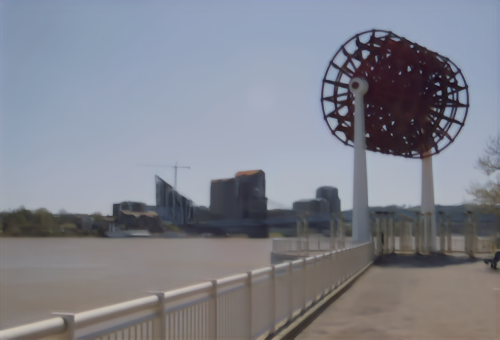}{0.4}{0.3}{0.55}{0.53}{2}{help_grid_off}{up_left}{line_connection_off}{2}{blue}{1}{red} }\hspace{-0.3in}
\subfigure[\scriptsize{(g) SMG (16.35/84.87)}]{
\zoomincludgraphic{1.33in}{forResub/fivek/SMG/00118_SMG.png}{0.4}{0.3}{0.55}{0.53}{2}{help_grid_off}{up_left}{line_connection_off}{2}{blue}{1}{red} }\hspace{-0.3in}
\subfigure[\scriptsize{(h) Bread (11.75/66.87)}]{
\zoomincludgraphic{1.33in}{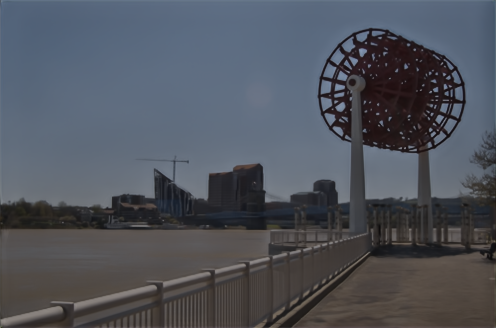}{0.4}{0.3}{0.55}{0.53}{2}{help_grid_off}{up_left}{line_connection_off}{2}{blue}{1}{red} }\hspace{-0.3in}
\subfigure[\scriptsize{(i) Ours (21.65/90.12)}]{
\zoomincludgraphic{1.33in}{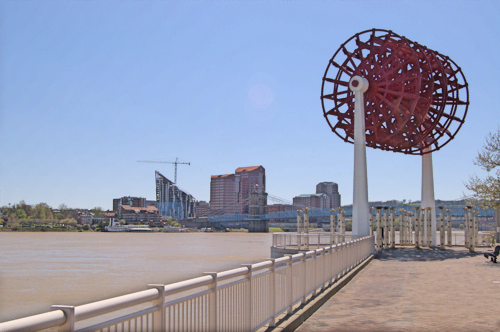}{0.4}{0.3}{0.55}{0.53}{2}{help_grid_off}{up_left}{line_connection_off}{2}{blue}{1}{red} }\hspace{-0.3in}
\subfigure[\scriptsize{(j) ExpertC}]{
\zoomincludgraphic{1.33in}{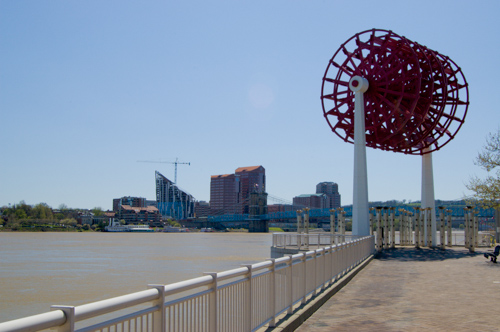}{0.4}{0.3}{0.55}{0.53}{2}{help_grid_off}{up_left}{line_connection_off}{2}{blue}{1}{red} }\hspace{-0.3in}
\caption{\y Image enhancement results with index PSNR (dB) and SSIM (\%). (a) is the input low-light image from dataset FiveK; the enhanced results of (b) RetinexNet \cite{Chen2018Retinex}, (c) GLADNet \cite{wang2018gladnet}, (d) URetinexNet \cite{wu2022uretinex}, (e) EFINet \cite{liu2022efinet}, (f) RQ \cite{liu2023low}, (g) SMG \cite{xu2023low}, (h) Bread \cite{guo2023low}, and (i) Ours; (j) is the reference image renditions by expert C in FiveK dataset. }\label{fivek1}
\end{figure*}

\begin{figure*}[t!]
\setlength{\abovecaptionskip}{0in}
\subfigcapskip=-0.05in
% \centering
\subfigure[\scriptsize{(a) Input (10.35/40.65)}]{
\zoomincludgraphic{1.33in}{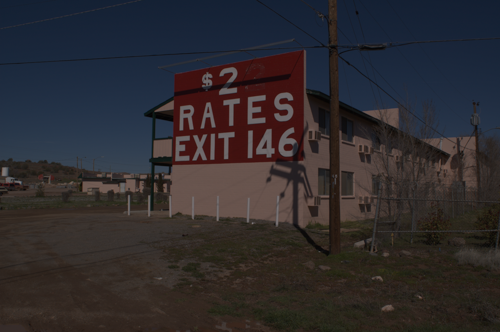}{0.295}{0.49}{0.385}{0.6}{3}{help_grid_off}{bottom_left}{line_connection_off}{2}{blue}{1}{red}}\hspace{-0.25in}
\subfigure[\scriptsize{(b) RetinexNet (14.64/78.98)}]{
\zoomincludgraphic{1.33in}{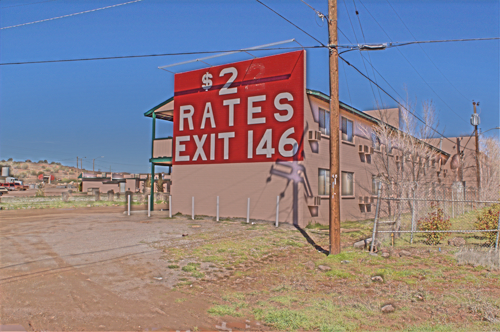}{0.295}{0.49}{0.385}{0.6}{3}{help_grid_off}{bottom_left}{line_connection_off}{2}{blue}{1}{red} }\hspace{-0.3in}
\subfigure[\scriptsize{(c) GLADNet (14.46/72.81)}]{
\zoomincludgraphic{1.33in}{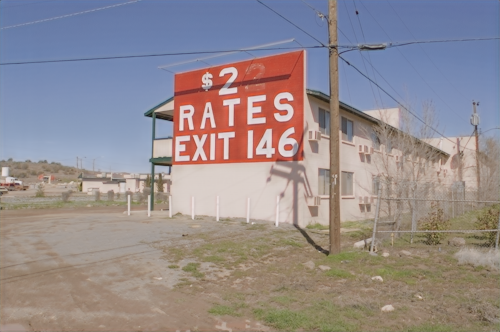}{0.295}{0.49}{0.385}{0.6}{3}{help_grid_off}{bottom_left}{line_connection_off}{2}{blue}{1}{red} }
\hspace{-0.35in}
\subfigure[\scriptsize{(d) URetinexNet (12.43/73.42)}]{
\zoomincludgraphic{1.33in}{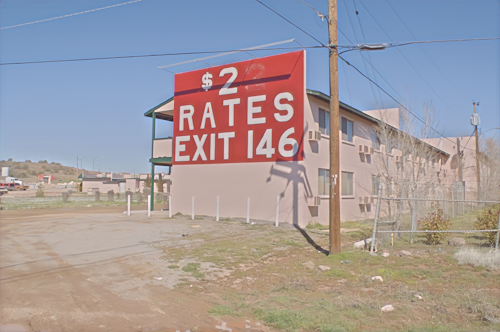}{0.295}{0.49}{0.385}{0.6}{3}{help_grid_off}{bottom_left}{line_connection_off}{2}{blue}{1}{red} }
\hspace{-0.35in}
\subfigure[\scriptsize{(e) EFINet (17.77/80.96)}]{
\zoomincludgraphic{1.33in}{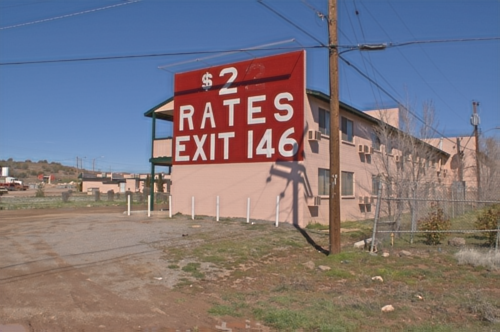}{0.295}{0.49}{0.385}{0.6}{3}{help_grid_off}{bottom_left}{line_connection_off}{2}{blue}{1}{red} }
\hspace{-0.3in}\vspace{-0.1in}

\subfigure[\scriptsize{(f) RQ (16.01/70.50)}]{
\zoomincludgraphic{1.33in}{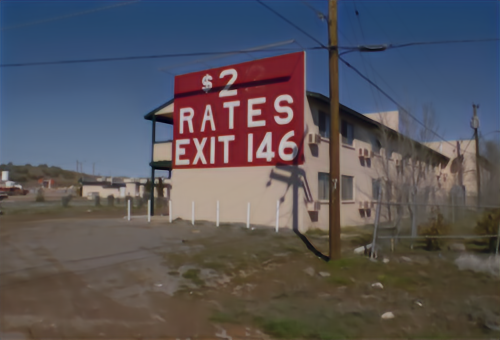}{0.295}{0.49}{0.385}{0.6}{3}{help_grid_off}{bottom_left}{line_connection_off}{2}{blue}{1}{red} }\hspace{-0.3in}
\subfigure[\scriptsize{(g) SMG (17.80/79.63)}]{
\zoomincludgraphic{1.33in}{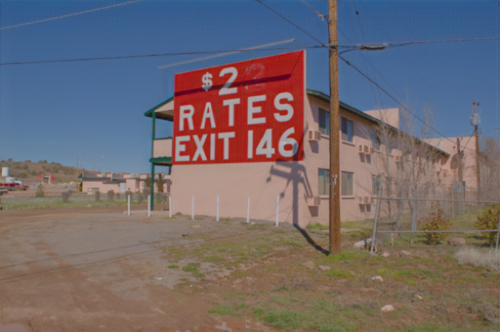}{0.295}{0.49}{0.385}{0.6}{3}{help_grid_off}{bottom_left}{line_connection_off}{2}{blue}{1}{red} }\hspace{-0.3in}
\subfigure[\scriptsize{(h) Bread (13.73/61.83)}]{
\zoomincludgraphic{1.33in}{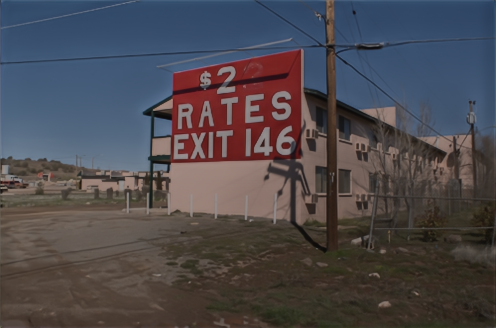}{0.295}{0.49}{0.385}{0.6}{3}{help_grid_off}{bottom_left}{line_connection_off}{2}{blue}{1}{red} }\hspace{-0.3in}
\subfigure[\scriptsize{(i) Ours (17.86/83.64)}]{
\zoomincludgraphic{1.33in}{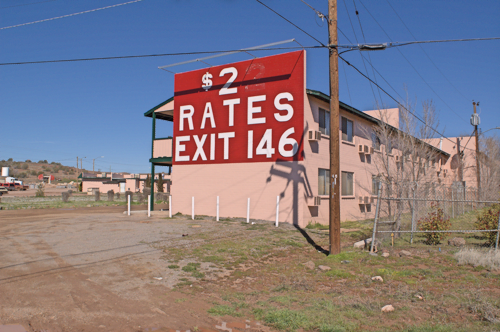}{0.295}{0.49}{0.385}{0.6}{3}{help_grid_off}{bottom_left}{line_connection_off}{2}{blue}{1}{red} }\hspace{-0.3in}
\subfigure[\scriptsize{(j) ExpertC}]{
\zoomincludgraphic{1.33in}{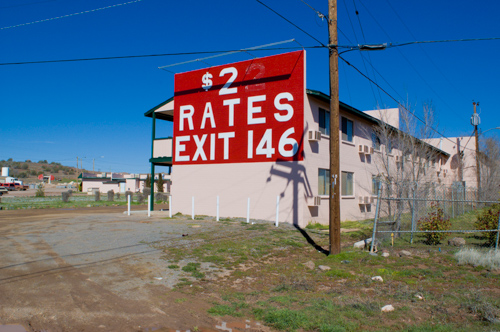}{0.295}{0.49}{0.385}{0.6}{3}{help_grid_off}{bottom_left}{line_connection_off}{2}{blue}{1}{red} }\hspace{-0.3in}
\caption{\y Image enhancement results with index PSNR (dB) and SSIM (\%). (a) is the input low-light image from dataset FiveK; the enhanced results of (b) RetinexNet \cite{Chen2018Retinex}, (c) GLADNet \cite{wang2018gladnet}, (d) URetinexNet \cite{wu2022uretinex}, (e) EFINet \cite{liu2022efinet}, (f) RQ \cite{liu2023low}, (g) SMG \cite{xu2023low} (h) Bread \cite{guo2023low}, and (i) Ours; (j) is the reference image renditions by expert C in FiveK dataset. }\label{fivek2}
\end{figure*}

\begin{figure}[t!]
\setlength{\abovecaptionskip}{0in}
\subfigcapskip=-0.09in
\hspace{-0.245in}
\subfigure[\tiny{(a) Input (10.16/23.81)}]{
\zoomincludgraphic{0.095\textwidth}{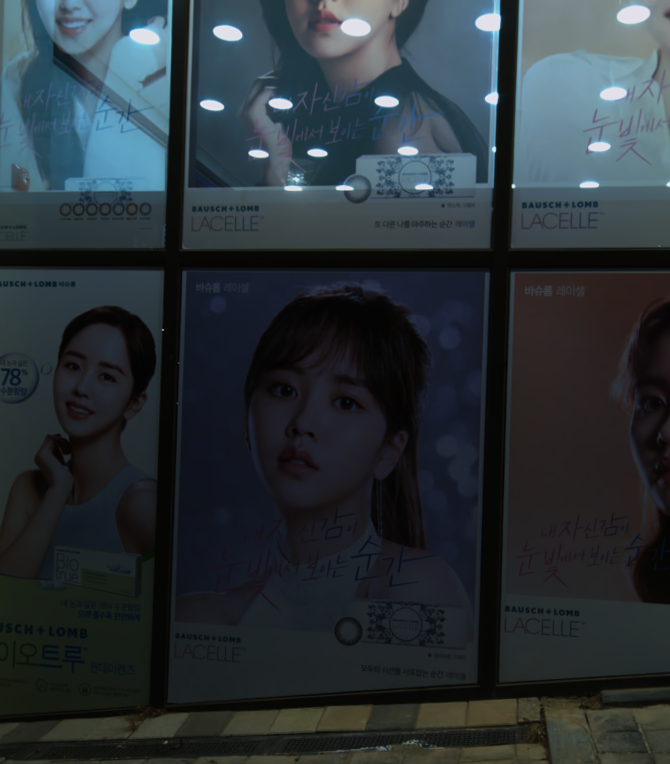}{0.07}{0.42}{0.24}{0.6}{2.6}{help_grid_off}{up_right}{line_connection_off}{1.2}{blue}{1}{red}}\hspace{-0.236in}
\subfigure[\tiny{(b) RetiNet (15.25/72.64)~~~~~}]
{\zoomincludgraphic{0.095\textwidth}{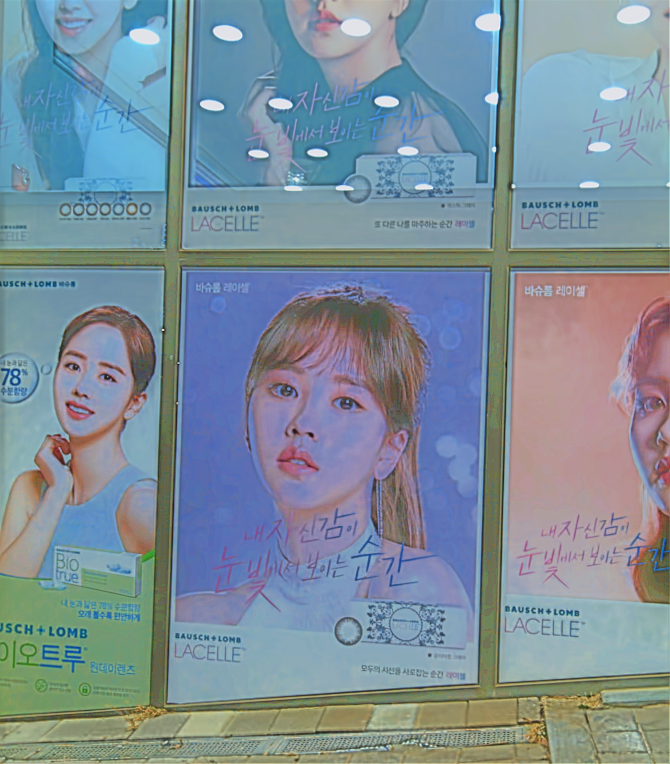}{0.07}{0.42}{0.24}{0.6}{2.6}{help_grid_off}{up_right}{line_connection_off}{1.2}{blue}{1}{red} }\hspace{-0.333in}
\subfigure[\tiny{(c) EFINet (16.14/86.83)}]{
\zoomincludgraphic{0.095\textwidth}{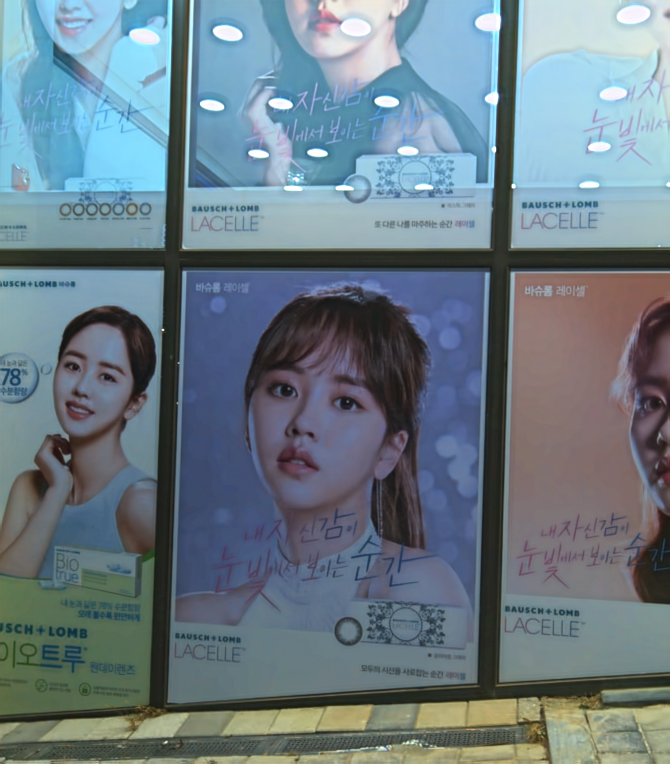}{0.07}{0.42}{0.24}{0.6}{2.6}{help_grid_off}{up_right}{line_connection_off}{1.2}{blue}{1}{red} }\hspace{-0.334in}
\subfigure[\tiny{(d)  RQ (17.16/81.58)}]{
\zoomincludgraphic{0.095\textwidth}{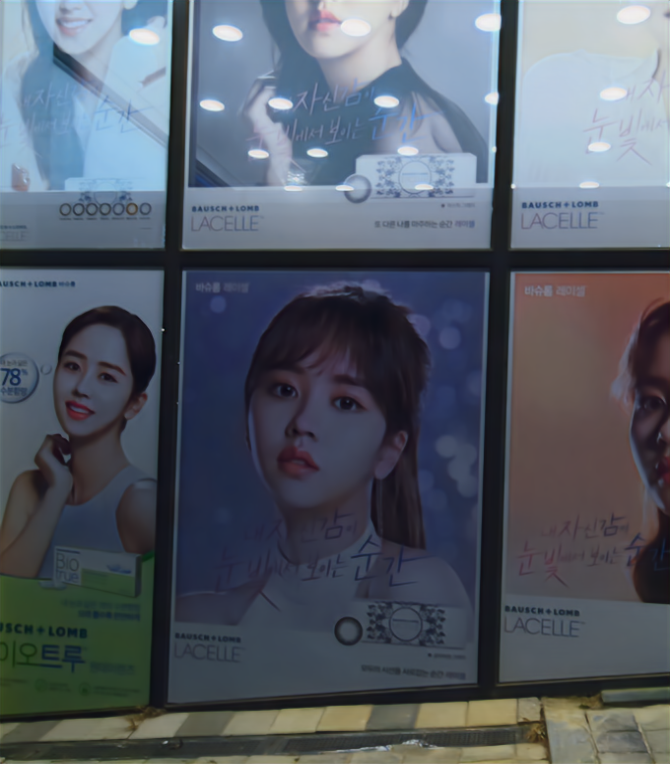}{0.07}{0.42}{0.24}{0.6}{2.6}{help_grid_off}{up_right}{line_connection_off}{1.2}{blue}{1}{red} }
\hspace{-0.37in}
\subfigure[\tiny{(e) LR3M (18.80/87.60)}]{
\zoomincludgraphic{0.095\textwidth}{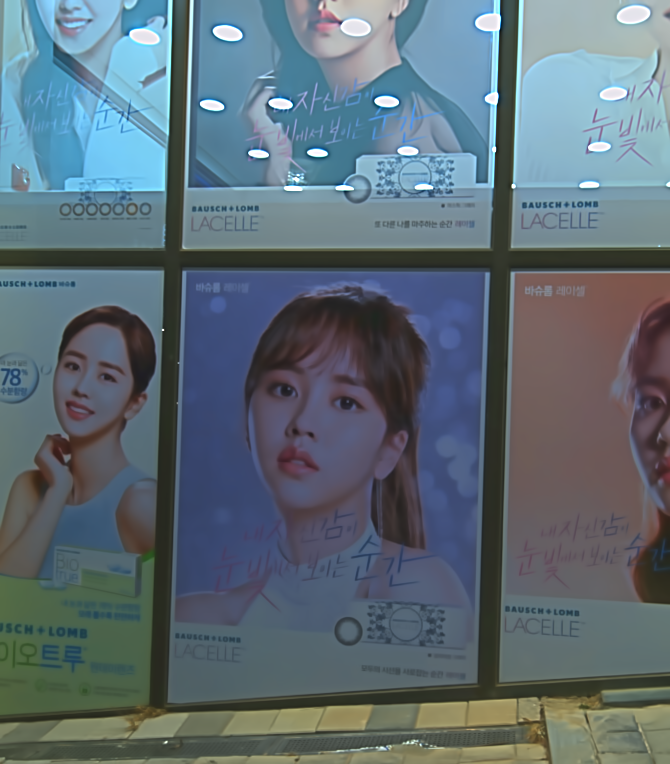}{0.07}{0.42}{0.24}{0.6}{2.6}{help_grid_off}{up_right}{line_connection_off}{1.2}{blue}{1}{red} }
\vspace{-0.05in}

\hspace{-0.256in}
\subfigure[\tiny{(f) QSPE (13.29/61.29)}]{
\zoomincludgraphic{0.095\textwidth}{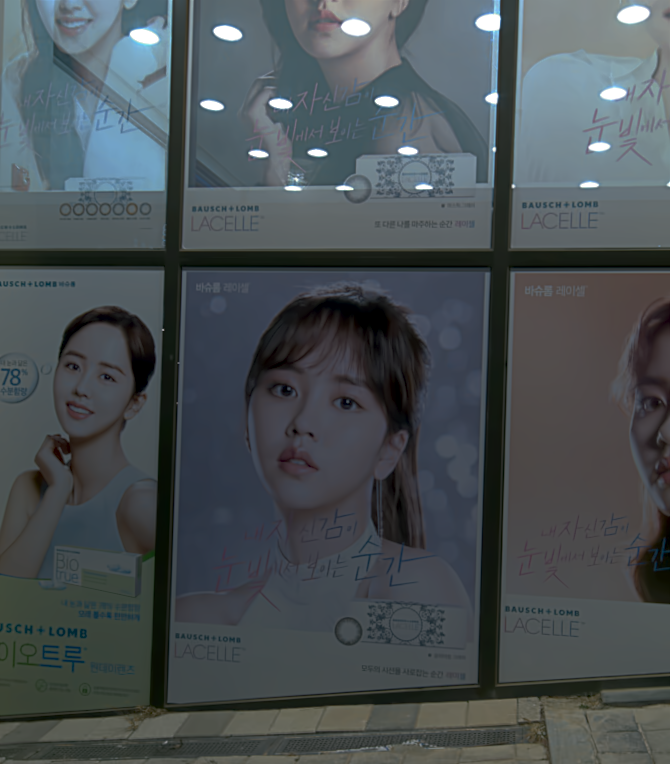}{0.07}{0.42}{0.24}{0.6}{2.6}{help_grid_off}{up_right}{line_connection_off}{1.2}{blue}{1}{red} }\hspace{-0.345in}
\subfigure[\tiny{(g) DRE (14.83/81.71)}]{
\zoomincludgraphic{0.095\textwidth}{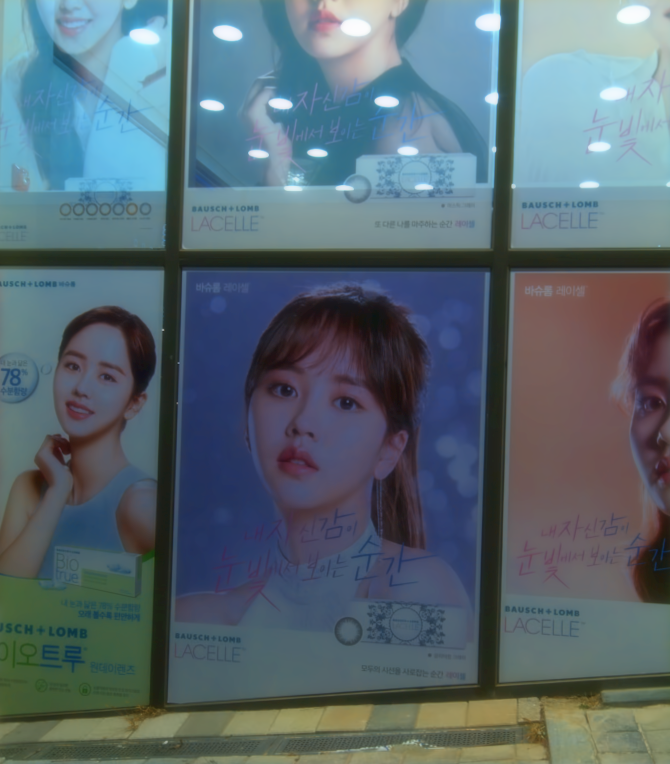}{0.07}{0.42}{0.24}{0.6}{2.6}{help_grid_off}{up_right}{line_connection_off}{1.2}{blue}{1}{red} }\hspace{-0.345in}
\subfigure[\tiny{(h) RWR (14.14/75.39)}]{
\zoomincludgraphic{0.095\textwidth}{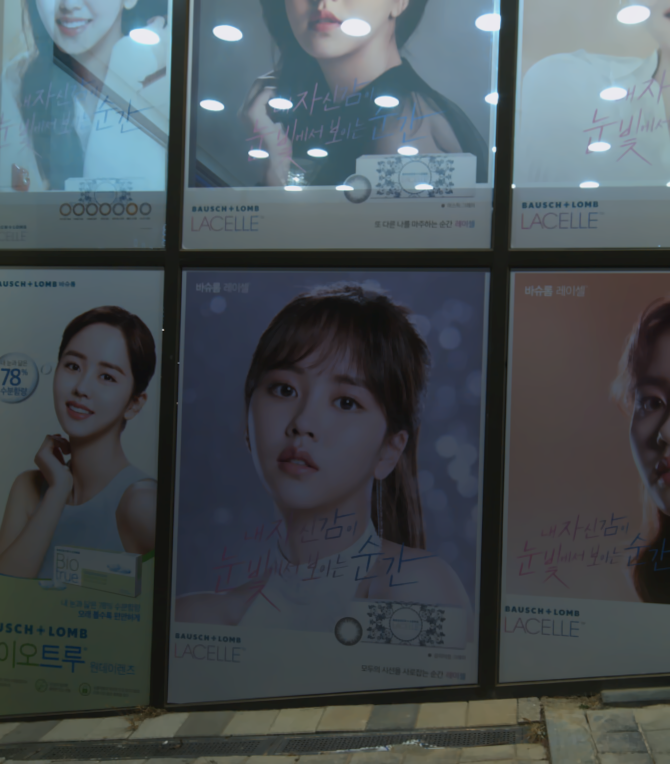}{0.07}{0.42}{0.24}{0.6}{2.6}{help_grid_off}{up_right}{line_connection_off}{1.2}{blue}{1}{red} }\hspace{-0.346in}
\subfigure[\tiny{(i) Ours (19.01/88.26)}]{
\zoomincludgraphic{0.095\textwidth}{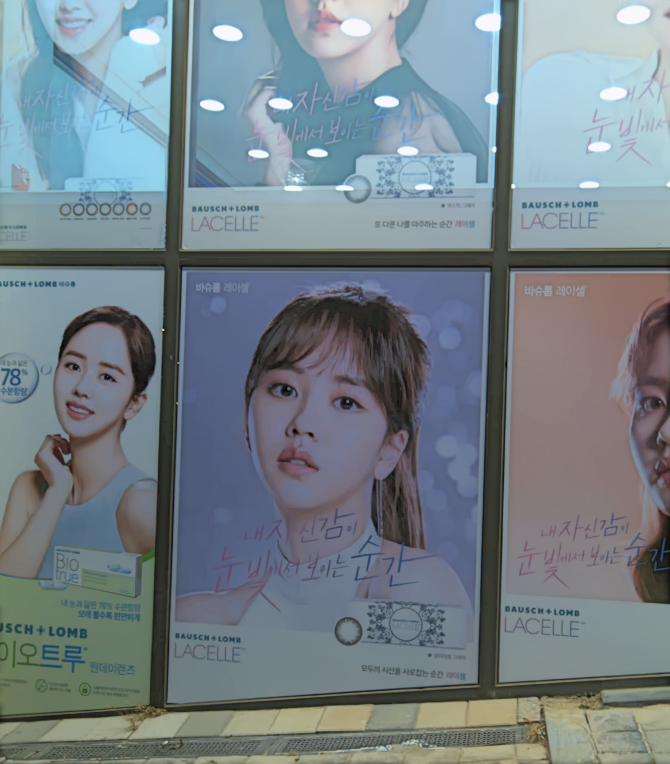}{0.07}{0.42}{0.24}{0.6}{2.6}{help_grid_off}{up_right}{line_connection_off}{1.2}{blue}{1}{red} }\hspace{-0.345in}
\subfigure[\tiny{(j) GT}]{
\zoomincludgraphic{0.095\textwidth}{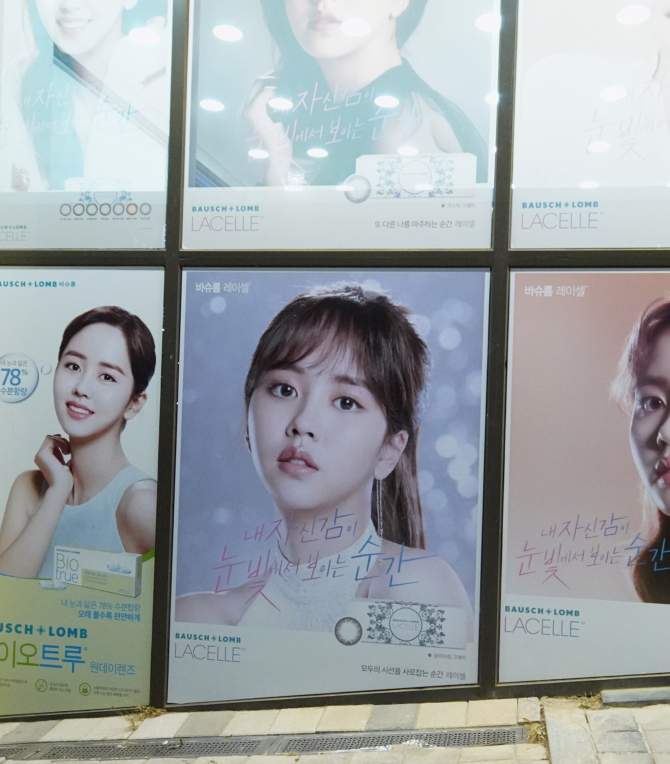}{0.07}{0.42}{0.24}{0.6}{2.6}{help_grid_off}{up_right}{line_connection_off}{1.2}{blue}{1}{red}}
\caption{Image enhancement results with indexes PSNR (dB) and SSIM (\%). (a) is the observed low-light image Img1; the enhanced results of (b) RetinexNet \cite{Chen2018Retinex}, (c) EFINet \cite{liu2022efinet}, (d) RQ \cite{liu2023low}, (e) LR3M \cite{9056796}, (f) QSPE \cite{huang2022quaternion}, (g) DRE \cite{jia2024low}, (h) RWR \cite{jia2024variational}, and (i) Ours; (j) is the ground truth (GT). }\label{Fig:realblur}
\end{figure}

{\y
Besides the numerical results, visual performance is presented in Figs. \ref{fivek1}, \ref{fivek2}. With the guidance of our learned edge extractor, the enhanced results in our (i) showed better structural preserving ability.  While the SMG \cite{xu2023low} is also guided by the edge module, some detailed information is not satisfactory. From both the numerical and visual results, the proposed method generates competitive performance. 
}On the other hand, the visual results of Img1 from Realblur15 are presented in Fig \ref{Fig:realblur}. The real scene image with non-uniformed light and small spark lights is hard to handle. As shown in Fig. \ref{Fig:realblur} (b), (c), and (e), we can see that the small spark lights affect the image enhancement results and make the lights abnormal.  Compared to the paired ground-truth image (j), Fig. \ref{Fig:realblur} (b), (d), (f), and (g) have a color bias due to the non-uniformed lights. Although the proposed approach generates the best results among the presented images, the color of (i) is still not exactly the same as the ground-truth image (j). Therefore, image enhancement for real-world images is still a challenging task. 

\subsection{Non-reference quality assessment}
In this subsection, five classical real-world non-reference datasets, DICM \cite{lee2013contrast}, LIME \cite{guo2016lime}, MEF \cite{ma2015perceptual}, {\y Fusion \cite{Chen2018Retinex}, and NPE \cite{wang2013naturalness}} are tested to verify the robustness of the proposed scheme. The state-of-the-art methods EFINet \cite{liu2022efinet}, Bread \cite{guo2023low}, RQ \cite{liu2023low}, LR3M \cite{9056796}, DRE \cite{jia2024low}, RMR \cite{jia2024variational} are compared. Due to the non-reference of the images, we conduct the non-reference metric ARISM, NIQE, and FADE as our index. 
{\y In Table \ref{tab:results}, the average numerical results of five datasets are listed. With the help of the learned edge extractor, the proposed method shows better performance compared to the approaches with the traditional priors. For most circumstances,} {\y the proposed method has a competitive result compared to the deep learning-based methods.
Notably, our method achieved the highest ARISM score, which further demonstrates its effectiveness in enhancing real-world low-light images.}

\begin{table}[t!]
    \centering
    \tabcolsep=0.03in
        \caption{\y Average image enhancement results on non-reference datasets DICM, LIME, MEF, NPE, and Fusion. 
        We highlighted the \rr{BEST} result in red and the \rb{SECOND BEST} one in blue.}
    \label{tab:results}
    \begin{tabular}{ccccccccccccc}\hline
         Datasets&Index&EFINet&Bread&RQ& LR3M &DRE  &RWR  &Ours \\\hline
        \multirow{3}{*}{DICM} 
        &ARISM$\uparrow$&1.243&1.223&1.208&1.235&1.207&\rb{1.255}&\rr{1.264}\\
        &NIQE$\downarrow$&\rr{2.516}&3.087&3.382&3.580& 3.325  &{3.051}&\rb{3.042}\\
        &FADE$\downarrow$&1.660&1.884&1.731&1.664&1.717&\rb{1.641}&\rr{1.604}\\
        \hline
        \multirow{3}{*}{LIME} &ARISM$\uparrow$&{1.243}&1.218&1.197&1.218&1.205&\rb{1.259}&\rr{1.263}\\
        &NIQE$\downarrow$&3.991&4.132&4.108&4.569&3.845&\rb{3.723}&\rr{3.556}\\
        &FADE$\downarrow$&\rb{1.436}&1.612&1.472&1.504&1.669&1.442&\rr{1.435}\\\hline
        \multirow{3}{*}{MEF} &ARISM$\uparrow$&1.203&1.217&1.184&1.177&1.188&1.160&\rr{1.229}\\
        &NIQE$\downarrow$&3.651&\rb{3.527}&3.888&4.567&3.772&4.123&\rr{3.526}\\
        &FADE$\downarrow$&1.721&2.009&1.723&1.773&\rb{1.708}&1.824&\rr{1.546}\\\hline
        \multirow{3}{*}{Fusion} 
        &ARISM$\uparrow$ &1.254&1.229&1.208&1.230&1.226&\rb{1.273}&\rr{1.277}\\
        &NIQE$\downarrow$&\rb{3.021}&3.226&3.620&4.032&\rr{2.872}&3.569&3.267\\
        &FADE$\downarrow$&\rb{1.694}&1.729&1.733&\rr{1.616}&2.039&1.807&1.725\\\hline
        \multirow{3}{*}{NPE} 
        &ARISM$\uparrow$ &1.2674&1.230&1.223&\rb{1.263}&1.246&1.235&\rr{1.304}\\
        &NIQE$\downarrow$&3.305&3.269&3.531&3.987&\rr{3.076}&3.828&\rb{3.302}\\
        &FADE$\downarrow$&\rb{1.747}&1.770&1.842&\rr{1.728}&2.006&1.971&\rr{1.728}\\\hline
    \end{tabular}
\end{table}

\begin{figure*}[t!]
\setlength{\abovecaptionskip}{0in}
\subfigcapskip=-0.05in
%\centering
\subfigure[\scriptsize{(a) Input}]{
\zoomincludgraphic{1.225in}{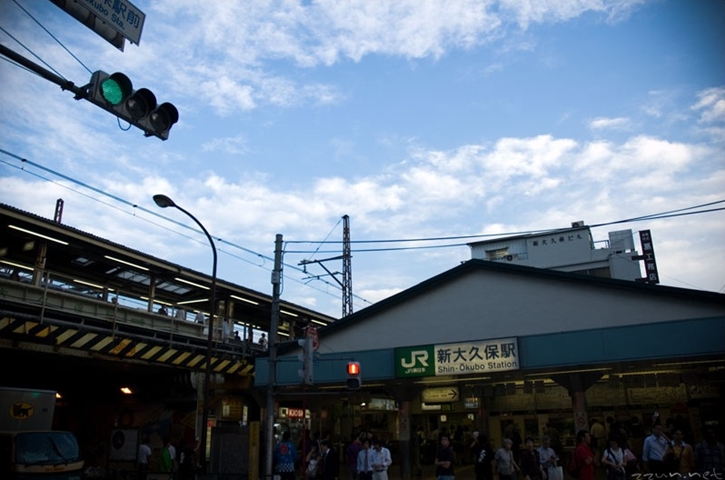}{0.15}{0.001}{0.001}{0.4}{2.42}{help_grid_off}{up_right}{line_connection_off}{2}{blue}{1}{red}}\hspace{-0.25in}
\subfigure[\scriptsize{(b) EFINet (2.247)}]{
\zoomincludgraphic{1.225in}{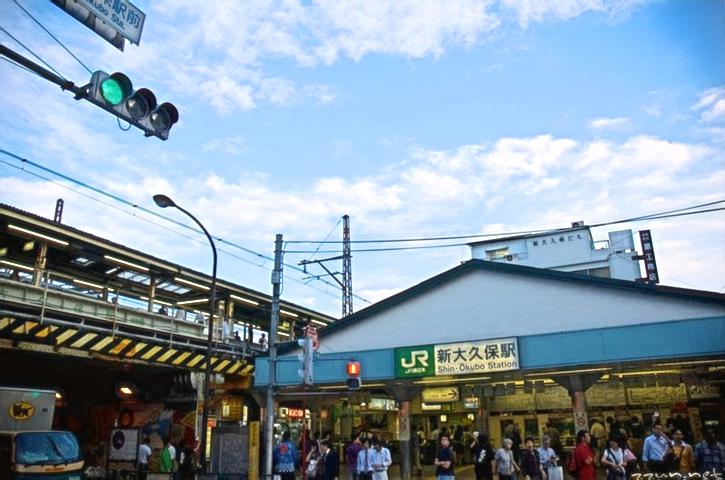}{0.15}{0.001}{0.001}{0.4}{2.42}{help_grid_off}{up_right}{line_connection_off}{2}{blue}{1}{red} }\hspace{-0.3in}
\subfigure[\scriptsize{(c) Bread (2.535)}]{
\zoomincludgraphic{1.225in}{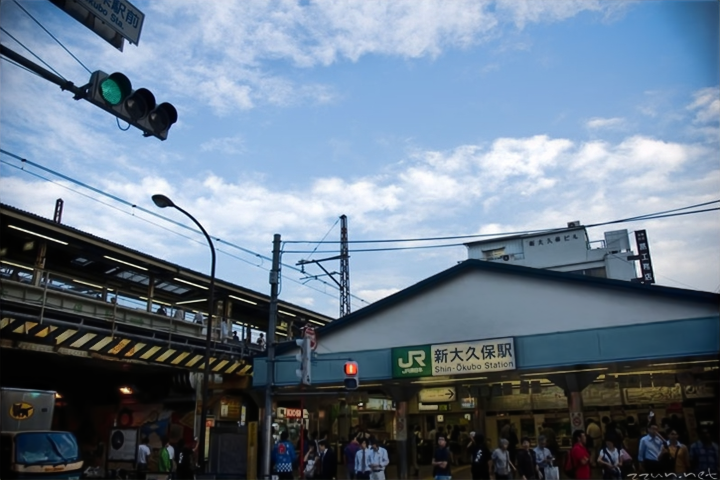}{0.15}{0.001}{0.001}{0.4}{2.42}{help_grid_off}{up_right}{line_connection_off}{2}{blue}{1}{red} }
\hspace{-0.35in}
\subfigure[\scriptsize{(d) RQ (3.144)}]{
\zoomincludgraphic{1.225in}{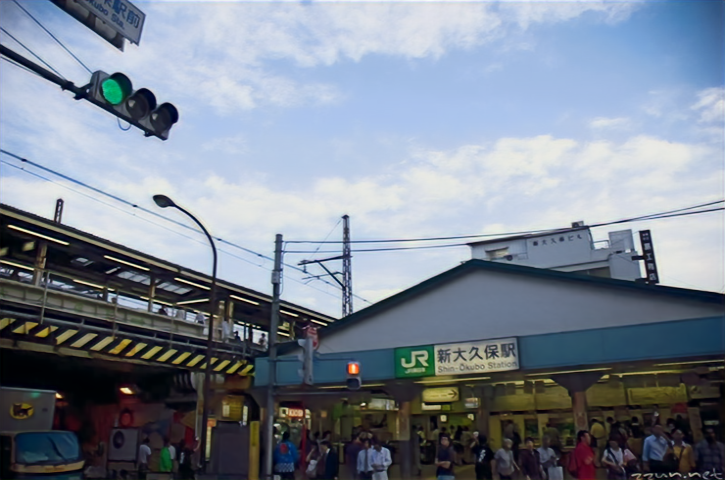}{0.15}{0.001}{0.001}{0.4}{2.42}{help_grid_off}{up_right}{line_connection_off}{2}{blue}{1}{red} }
\hspace{-0.3in}\vspace{-0.1in}

\subfigure[\scriptsize{(e) LR3M (3.403)}]{
\zoomincludgraphic{1.225in}{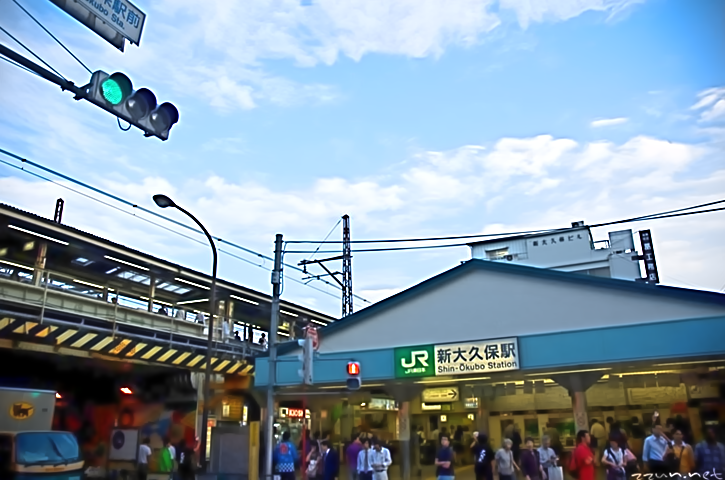}{0.15}{0.001}{0.001}{0.4}{2.42}{help_grid_off}{up_right}{line_connection_off}{2}{blue}{1}{red} }
\hspace{-0.35in}
\subfigure[\scriptsize{(f) DRE (2.144)}]{
\zoomincludgraphic{1.225in}{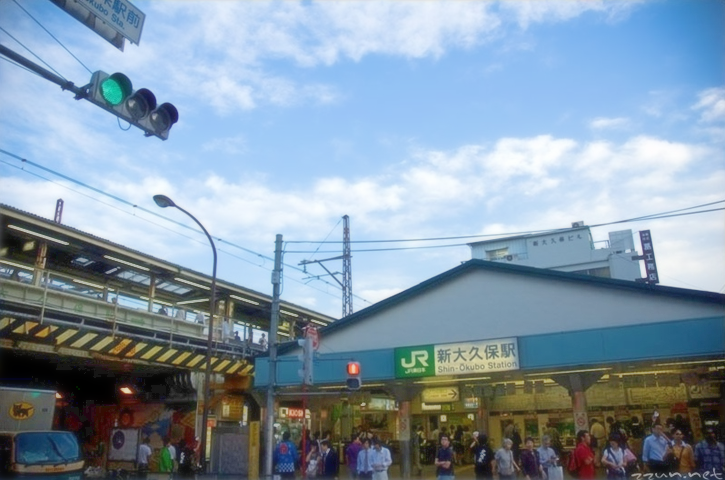}{0.15}{0.001}{0.001}{0.4}{2.42}{help_grid_off}{up_right}{line_connection_off}{2}{blue}{1}{red} }\hspace{-0.3in}
\subfigure[\scriptsize{(g) RWR (3.365)}]{
\zoomincludgraphic{1.225in}{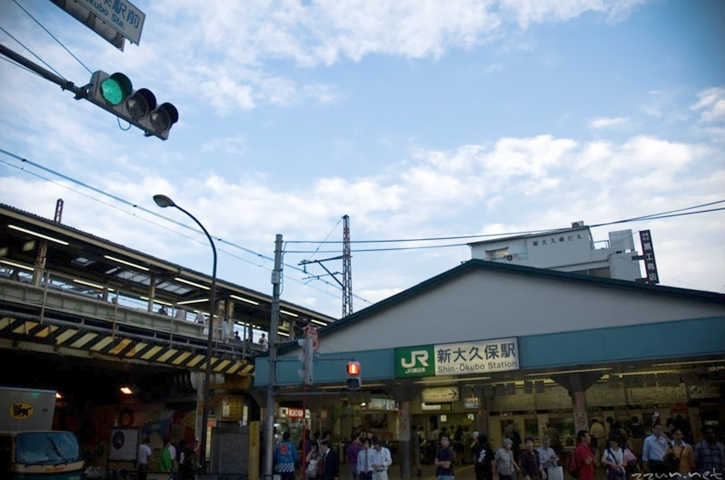}{0.15}{0.001}{0.001}{0.4}{2.42}{help_grid_off}{up_right}{line_connection_off}{2}{blue}{1}{red} }\hspace{-0.3in}
\subfigure[\scriptsize{(h) Ours (2.139)}]{
\zoomincludgraphic{1.225in}{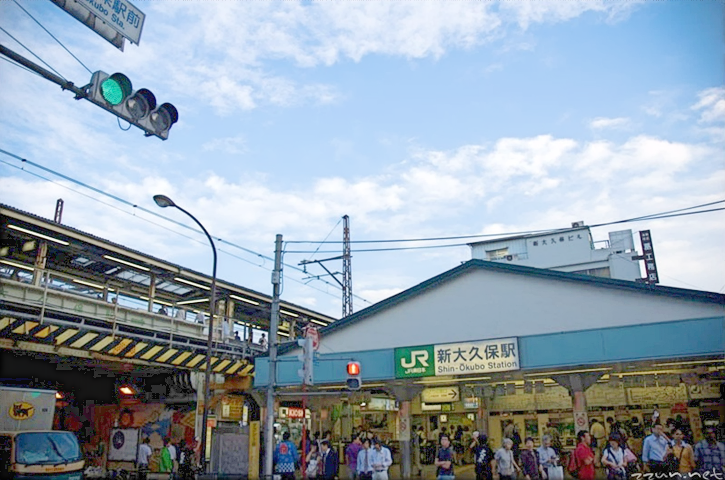}{0.15}{0.001}{0.001}{0.4}{2.42}{help_grid_off}{up_right}{line_connection_off}{2}{blue}{1}{red} }\hspace{-0.3in}
\caption{Image enhancement results with index NIQE$\downarrow$. (a) is the input low-light image form dataset DICM; the enhanced results of (b) EFINet \cite{liu2022efinet}, (c) Bread \cite{guo2023low}, (d) RQ \cite{liu2023low}, (e) LR3M \cite{9056796}, (f) DRE \cite{jia2024low}, (g) RWR \cite{jia2024variational}, and (h)  Ours.}\label{Fig:dicm}
\end{figure*}

\begin{figure*}[t]
\setlength{\abovecaptionskip}{0in}
\subfigcapskip=-0.05in
\subfigure[\scriptsize{(a) Input}]{
\zoomincludgraphic{1.02in}{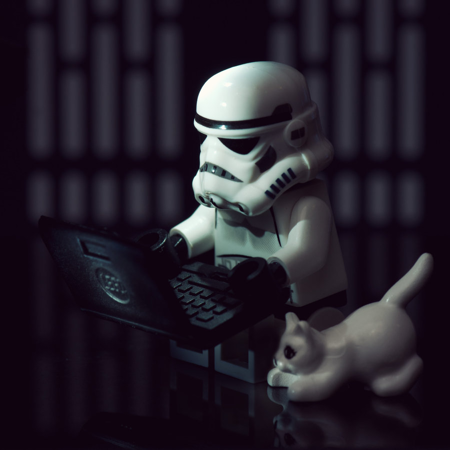}{0.65}{0.12}{0.37}{0.55}{2.27}{help_grid_off}{up_right}{line_connection_off}{2}{blue}{1}{red}}\hspace{-0.25in}
\subfigure[\scriptsize{(b) EFINet (4.220)}]{
\zoomincludgraphic{1.02in}{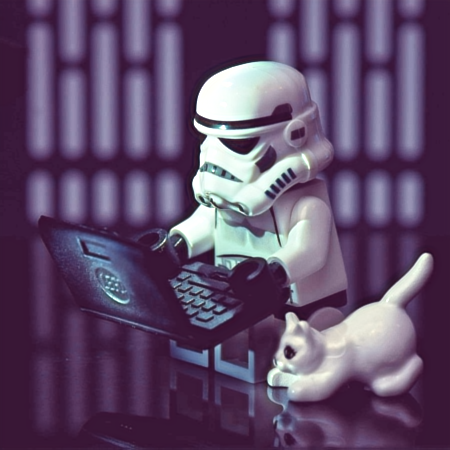}{0.65}{0.12}{0.37}{0.55}{2.27}{help_grid_off}{up_right}{line_connection_off}{2}{blue}{1}{red} }\hspace{-0.3in}
\subfigure[\scriptsize{(c) Bread (5.132)}]{
\zoomincludgraphic{1.02in}{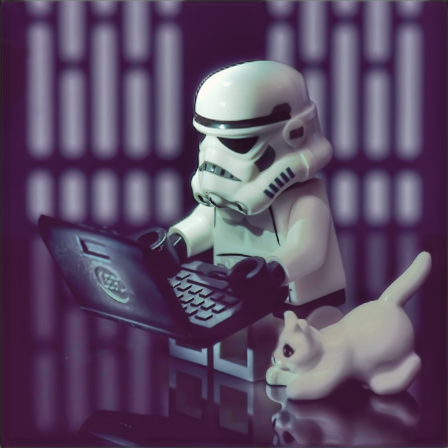}{0.65}{0.12}{0.37}{0.55}{2.27}{help_grid_off}{up_right}{line_connection_off}{2}{blue}{1}{red} }
\hspace{-0.35in}
\subfigure[\scriptsize{(d) RQ (4.424)}]{
\zoomincludgraphic{1.02in}{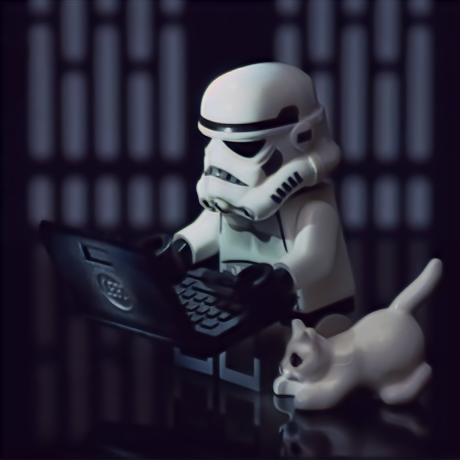}{0.65}{0.12}{0.37}{0.55}{2.27}{help_grid_off}{up_right}{line_connection_off}{2}{blue}{1}{red} }
\hspace{-0.3in}\vspace{-0.1in}

\subfigure[\scriptsize{(e) LR3M (4.801)}]{
\zoomincludgraphic{1.02in}{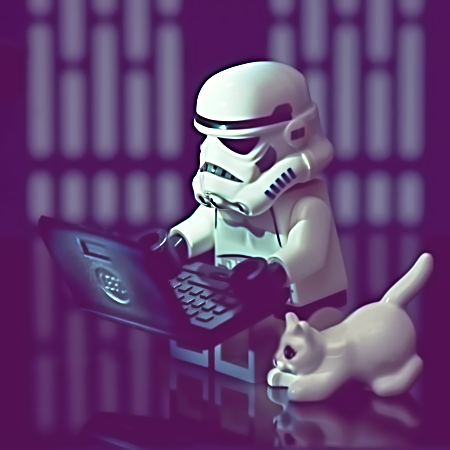}{0.65}{0.12}{0.37}{0.55}{2.27}{help_grid_off}{up_right}{line_connection_off}{2}{blue}{1}{red} }
\hspace{-0.35in}
\subfigure[\scriptsize{(f) DRE (4.586)}]{
\zoomincludgraphic{1.02in}{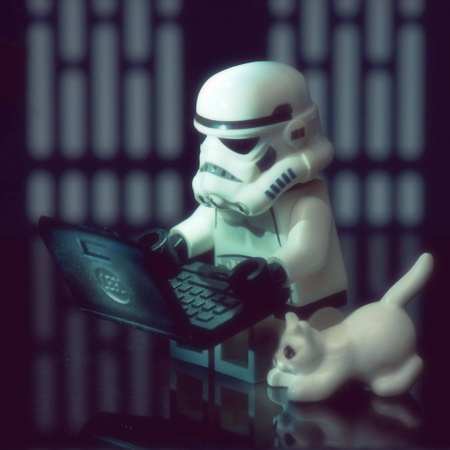}{0.65}{0.12}{0.37}{0.55}{2.27}{help_grid_off}{up_right}{line_connection_off}{2}{blue}{1}{red} }\hspace{-0.3in}
\subfigure[\scriptsize{(g) RWR (5.244)}]{
\zoomincludgraphic{1.02in}{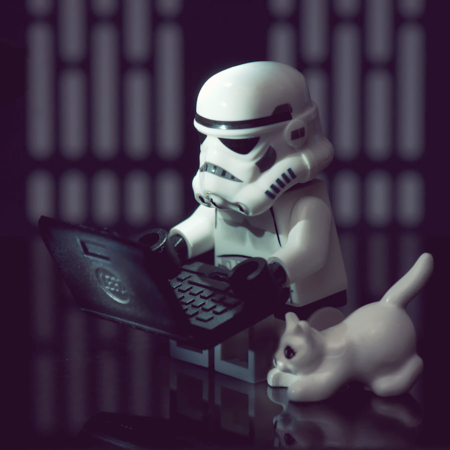}{0.65}{0.12}{0.37}{0.55}{2.27}{help_grid_off}{up_right}{line_connection_off}{2}{blue}{1}{red} }\hspace{-0.3in}
\subfigure[\scriptsize{(h) Ours (3.519)}]{
\zoomincludgraphic{1.02in}{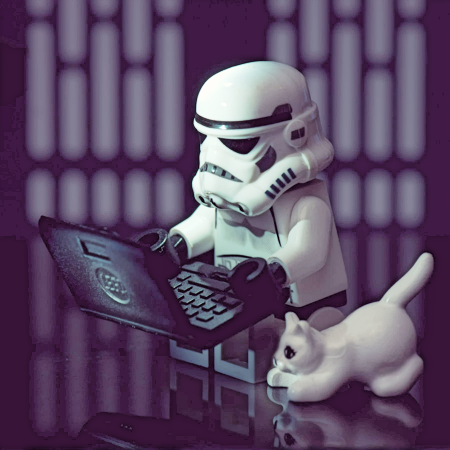}{0.65}{0.12}{0.37}{0.55}{2.27}{help_grid_off}{up_right}{line_connection_off}{2}{blue}{1}{red} }\hspace{-0.3in}
\caption{Image enhancement results with index NIQE$\downarrow$. (a) is the input low-light image form dataset LIME; the enhanced results of (b) EFINet \cite{liu2022efinet}, (c) Bread \cite{guo2023low}, (d) RQ \cite{liu2023low}, (e) LR3M \cite{9056796}, (f) DRE \cite{jia2024low}, (g) RWR \cite{jia2024variational}, and (h)  Ours. }\label{Fig:lime}
\end{figure*}

\begin{figure*}[t!]
\setlength{\abovecaptionskip}{0in}
\subfigcapskip=-0.05in
\subfigure[\scriptsize{(a) Input}]{
\zoomincludgraphic{1.1in}{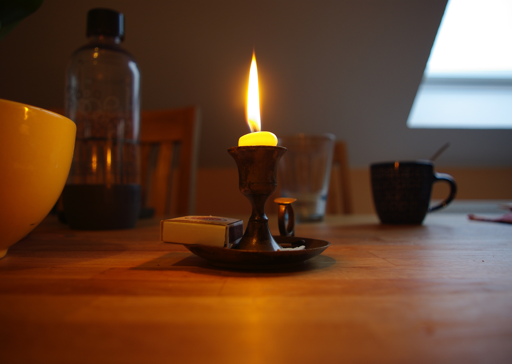}{0.89}{0.32}{0.71}{0.65}{2.94}{help_grid_off}{up_right}{line_connection_off}{2}{blue}{1}{red}}\hspace{-0.25in}
\subfigure[\scriptsize{(b) EFINet (3.333)}]{
\zoomincludgraphic{1.1in}{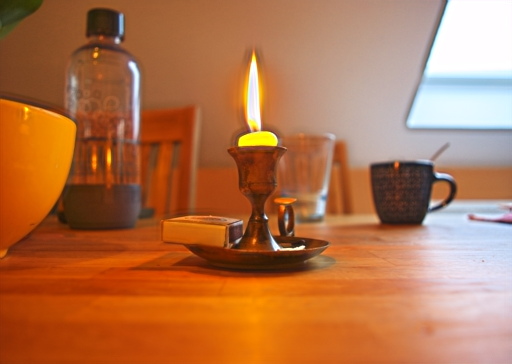}{0.89}{0.32}{0.71}{0.65}{2.94}{help_grid_off}{up_right}{line_connection_off}{2}{blue}{1}{red} }\hspace{-0.3in}
\subfigure[\scriptsize{(c) Bread (3.779)}]{
\zoomincludgraphic{1.1in}{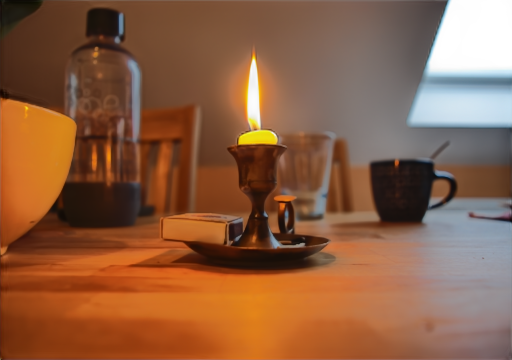}{0.89}{0.32}{0.71}{0.65}{2.94}{help_grid_off}{up_right}{line_connection_off}{2}{blue}{1}{red} }
\hspace{-0.35in}
\subfigure[\scriptsize{(d) RQ (4.161)}]{
\zoomincludgraphic{1.1in}{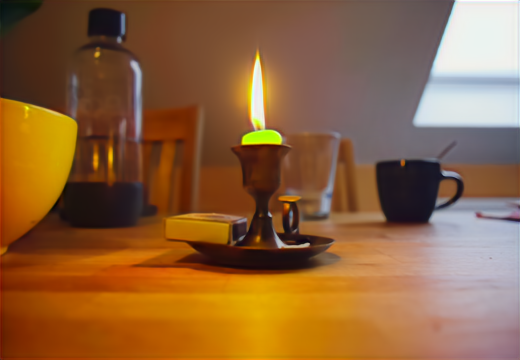}{0.89}{0.32}{0.71}{0.65}{2.94}{help_grid_off}{up_right}{line_connection_off}{2}{blue}{1}{red} }
\hspace{-0.3in}\vspace{-0.1in}

\subfigure[\scriptsize{(e) LR3M (3.935)}]{
\zoomincludgraphic{1.1in}{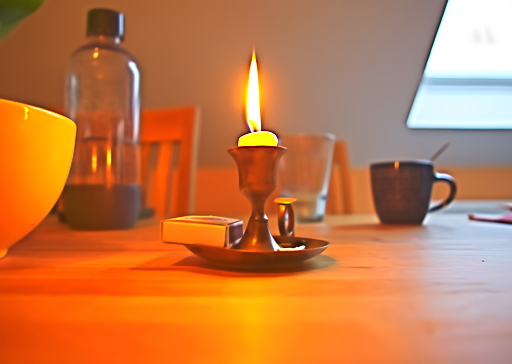}{0.89}{0.32}{0.71}{0.65}{2.94}{help_grid_off}{up_right}{line_connection_off}{2}{blue}{1}{red} }
\hspace{-0.35in}
\subfigure[\scriptsize{(f) DRE (3.597)}]{
\zoomincludgraphic{1.1in}{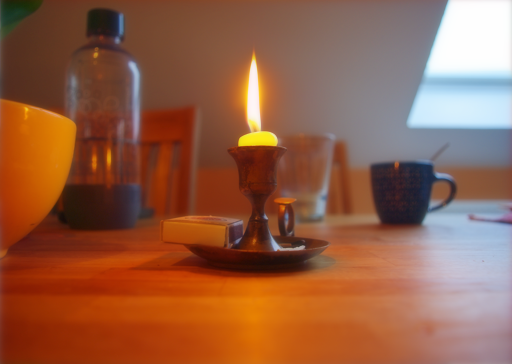}{0.89}{0.32}{0.71}{0.65}{2.94}{help_grid_off}{up_right}{line_connection_off}{2}{blue}{1}{red} }\hspace{-0.3in}
\subfigure[\scriptsize{(g) RWR (4.054)}]{
\zoomincludgraphic{1.1in}{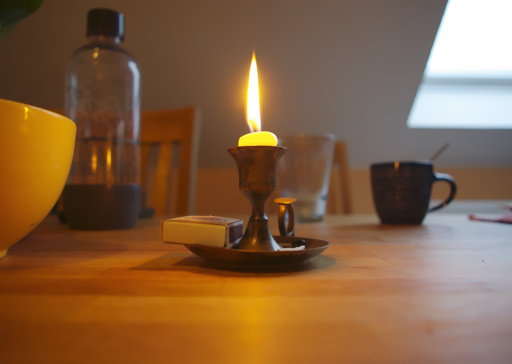}{0.89}{0.32}{0.71}{0.65}{2.94}{help_grid_off}{up_right}{line_connection_off}{2}{blue}{1}{red} }\hspace{-0.3in}
\subfigure[\scriptsize{(h) Ours (3.224)}]{
\zoomincludgraphic{1.1in}{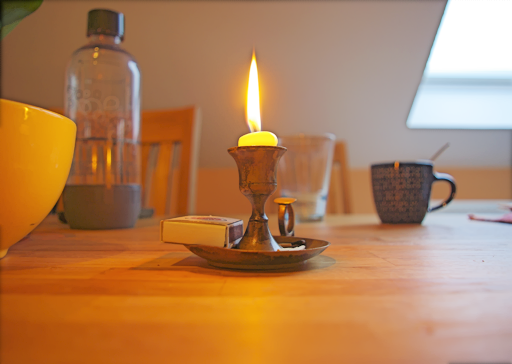}{0.89}{0.32}{0.71}{0.65}{2.94}{help_grid_off}{up_right}{line_connection_off}{2}{blue}{1}{red} }\hspace{-0.3in}
\caption{Image enhancement results with index NIQE$\downarrow$. (a) is the input low-light image form dataset MEF; the enhanced results of (b) EFINet \cite{liu2022efinet}, (c) Bread \cite{guo2023low}, (d) RQ \cite{liu2023low}, (e) LR3M \cite{9056796}, (f) DRE \cite{jia2024low}, (g) RWR \cite{jia2024variational}, and (h) Ours. }\label{Fig:mef}
\end{figure*}

In Fig. \ref{Fig:dicm}, we present the image enhancement results of the image in dataset DICM. The input image (a) is quite simple to handle compared to other extreme dark images. However, the zoom part of results (c), (d), and (g) remain dark. Compared to other methods, our approach generates the enhanced image (h) with a normal light. 
An image from dataset LIME with edge feature is presented in Fig. \ref{Fig:lime}. For better understanding, we zoom the texture part, which is a keyboard. Our result (h) has clear details of the keyboard in the enhanced image. From the zoom part of the input image (a), the keyboard is under the non-uniformed light, hence (b), (c), (e), and (f) are with the color bias. For (d) and (g), though the color remains, the details cannot be well preserved.  
In Fig. \ref{Fig:mef}, we present the image enhancement results of the image in dataset MEF. Similar color bias happens in (d), (e), and (g). For better visuality, we zoom in on the mug in the image. Due to the help of the proposed edge network, our result (h) has a better enhancement performance while others still have low contrast. 
In conclusion, the proposed method achieved competitive results in enhancing real-world low-light images, as evidenced by both numerical and visual evaluations. The design of the learned edge extractor makes the proposed method particularly effective in handling texture images.

\section{Conclusion}
This paper proposed an edge-guided Retinex model with a novel inertial Bregman alternating linearized minimization algorithm for low-light image enhancement. With the help of deep learning, we trained the edge extractor with multi-scale residual blocks to obtain the edge regularizer. The learned edge extractor allows us to directly capture edge features from the low-light image, which helps the image enhancement. Additionally, we designed an efficient inertial Bregman alternating linearized minimization algorithm to tackle the proposed Retinex model. The effectiveness of the proposed algorithm was demonstrated through both theoretical and experimental analysis. To better illustrate the superiority of the proposed scheme, some state-of-the-art models on real-world datasets were compared. Based on the presented experimental results, the proposed method with learned prior and solved by the proposed algorithm achieves competitive performance in enhancing the low-light images. However, we acknowledge that our results still exhibit a color bias issue when dealing with non-uniform real-world low-light images. To address this, we plan to design an efficient model in the future to improve the performance of non-uniformed low-light image enhancement.

\bibliographystyle{IEEEtran}
\bibliography{references}
\vfill

\begin{IEEEbiography}[{\includegraphics[width=1in,height=1.25in,clip,keepaspectratio]{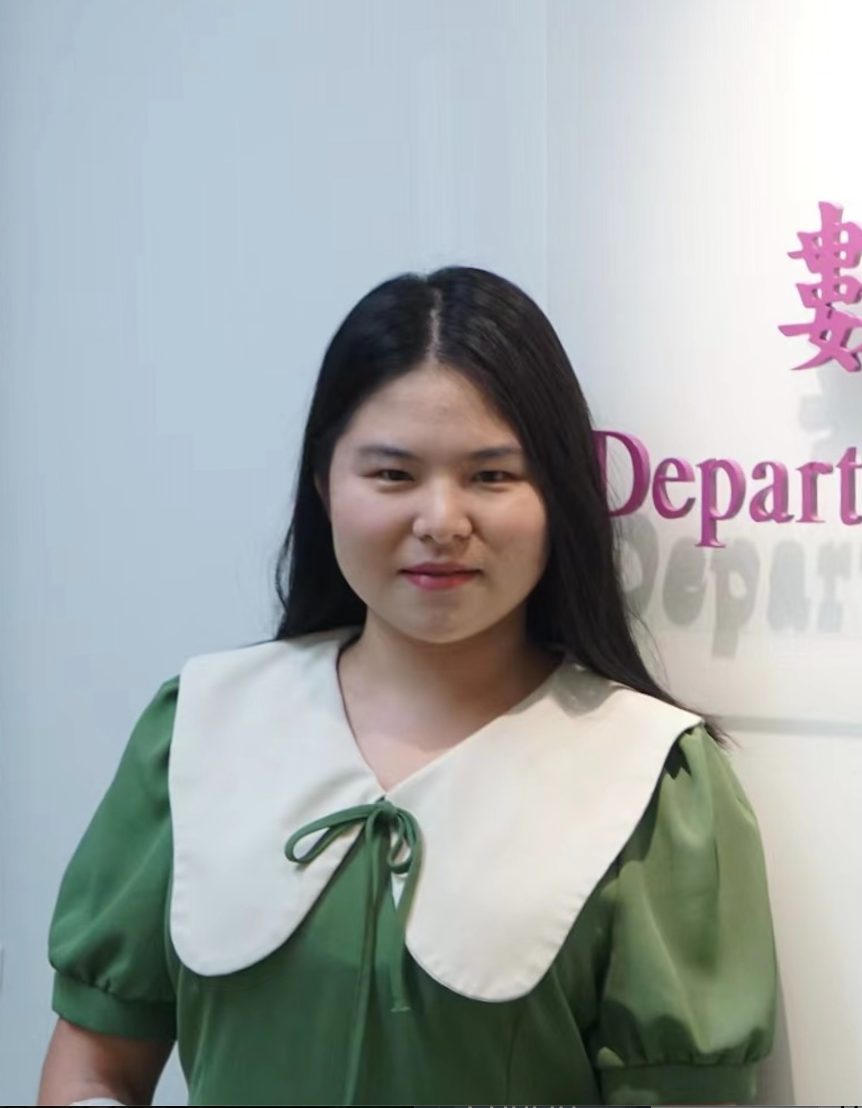}}]{Chaoyan Huang} received the M.S. degree from the School of Science, Nanjing University of Posts and Telecommunications, Nanjing, China, in 2022. She is currently a PhD candidate in the Department of Mathematics, The Chinese University of Hong Kong (CUHK). Her research interests include image processing, inverse problems, deep learning, and optimization.   
\end{IEEEbiography}

\begin{IEEEbiography}[{\includegraphics[width=1in,height=1.25in,clip,keepaspectratio]{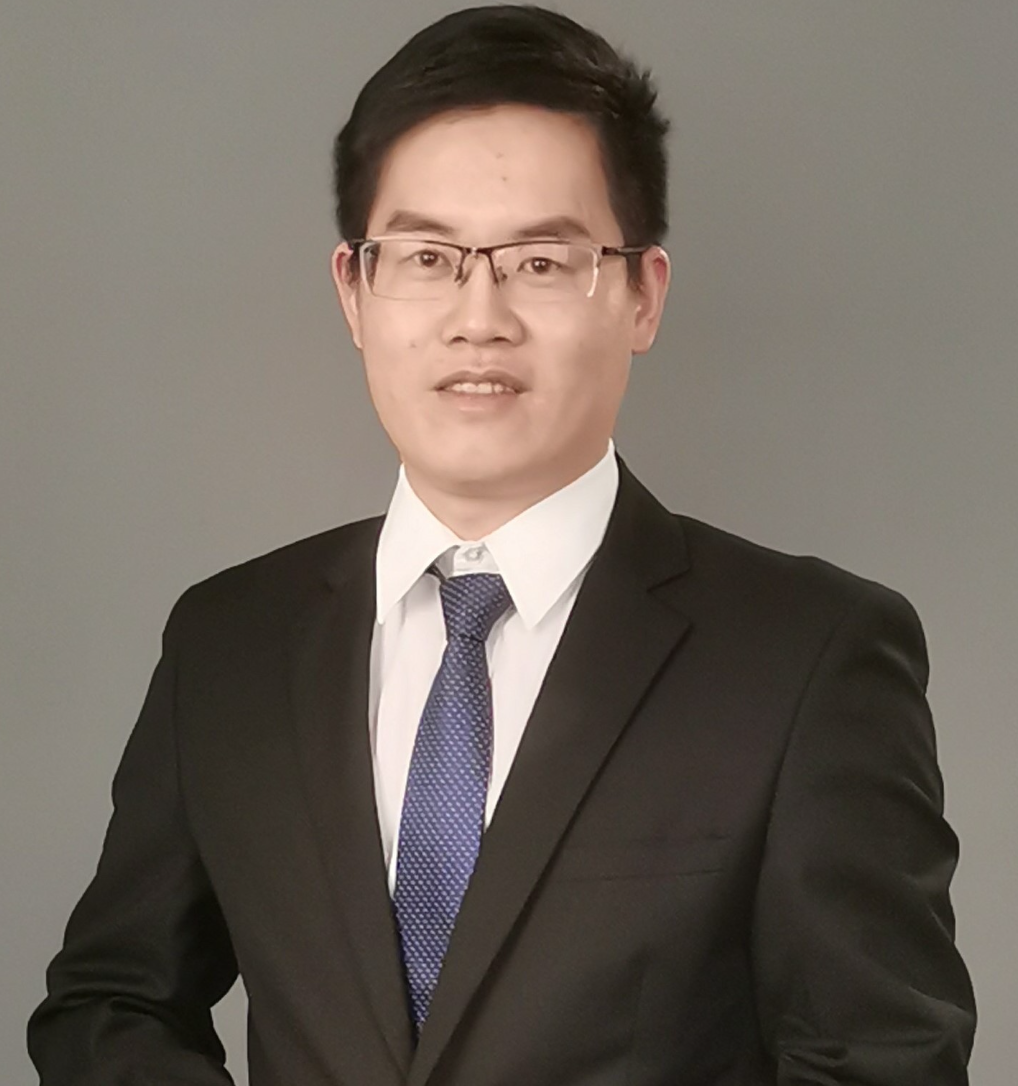}}]{Zhongming Wu} received the Ph.D. degree in management science and engineering from the School of Economics and Management, Southeast University, in 2019. From 2023 to 2025, he was a Postdoctoral Researcher with the Department of Mathematics, The Chinese University of Hong Kong (CUHK). He is currently a Professor with the School of Management Science and Engineering, Nanjing University of Information Science and Technology. His research interests include optimization theories and applications.   
\end{IEEEbiography}

\begin{IEEEbiography}[{\includegraphics[width=1in,height=1.25in,clip,keepaspectratio]{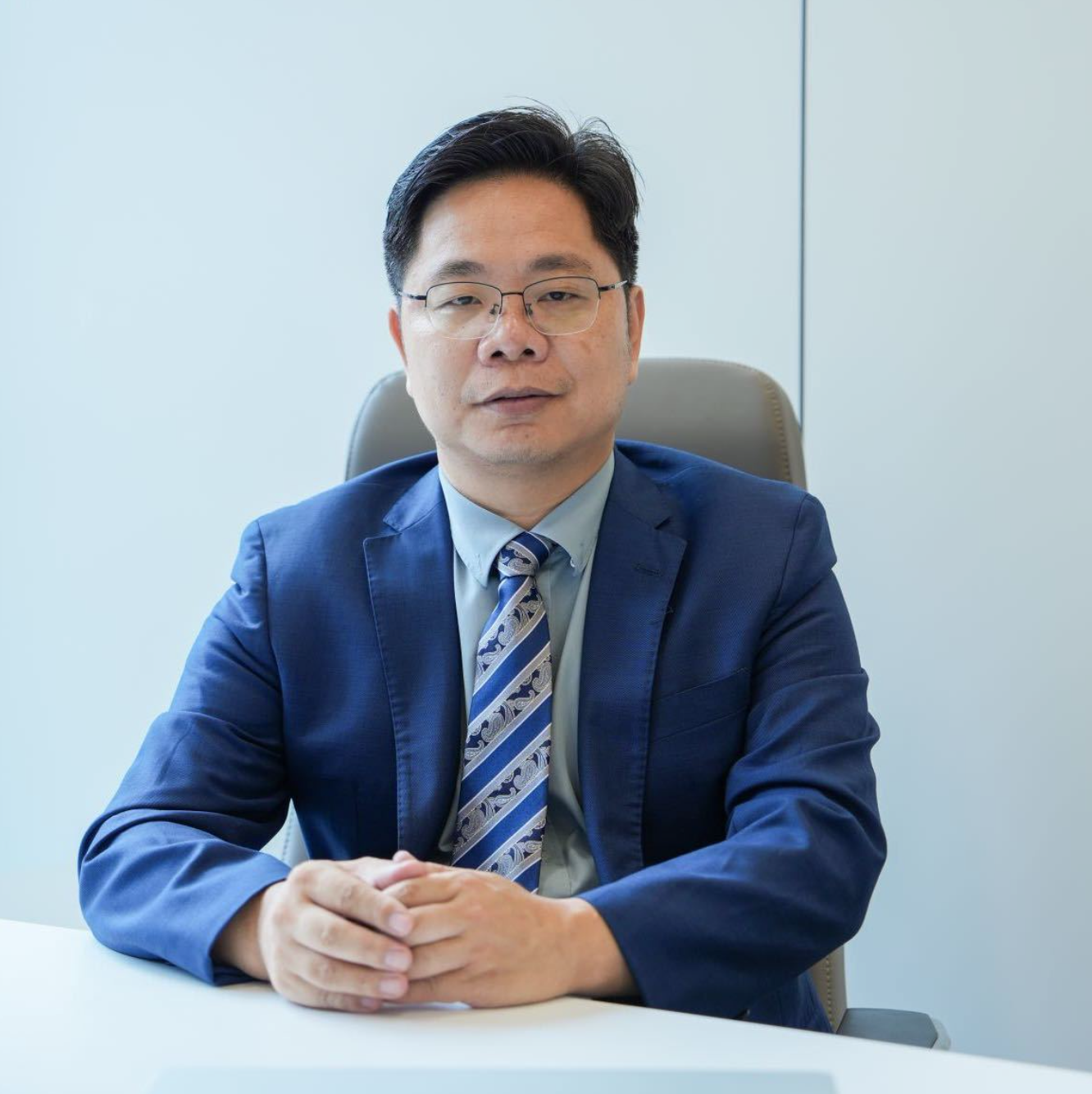}}]{Tieyong Zeng} received the B.S. degree from Peking University, Beijing, China, in 2000, the M.S. degree from École Polytechnique, Palaiseau, France, in 2004, and the Ph.D. degree from the University of Paris XIII, Paris, France, in 2007. From 2007 to 2008, he was a Postdoctoral Researcher with ENS de Cachan, Cachan, France, and an Assistant/Associate Professor with Hong Kong Baptist University, Hong Kong, from 2008 to 2018. He is currently a Professor with the Department of Mathematics, The Chinese University of Hong Kong, Hong Kong. His research interests include image processing, machine learning, and scientific computing.
\end{IEEEbiography}

\end{document}

%% file: imZoomConfig.tex
\usepackage{graphicx}
\usepackage{tikz}
\usepackage{subfigure}
\usetikzlibrary{positioning}
\usetikzlibrary{calc} % Only for Imagemagick workaround
\usepackage{ifthen}
\usepackage{xparse} % also loads expl3

 \ExplSyntaxOn
\NewDocumentCommand{\aspectratio}{smo}
 {% #2 is the image file
  \hbox_set:Nn \l_tmpa_box {\includegraphics{#2}}
  \IfNoValueTF{#3}
   {
    \__student_aspectratio:nn { \box_wd:N \l_tmpa_box } { \box_ht:N \l_tmpa_box }
   }
   {
    \IfBooleanTF{#1}{ \tl_gset:Nx } { \tl_set:Nx } #3
     {
      \__student_aspectratio:nn { \box_wd:N \l_tmpa_box } { \box_ht:N \l_tmpa_box }
     }
   }
 }

\cs_new:Nn \__student_aspectratio:nn
 {
  \fp_eval:n {round( #1 / #2 , 5)}
 }
\ExplSyntaxOff

\newcommand{\neworrenewcommand}[1]{\providecommand{#1}{}\renewcommand{#1}}

\newcommand{\zoomincludgraphic}[9]{
    \neworrenewcommand{\ffoo}[5]{
\begin{tikzpicture}[x=#1, y=#1, font=\footnotesize]
\aspectratio{#2}[\imsizeratio] % the ratio of width and height of input image
  \node[anchor = south east, inner sep=0] (image) at (1,0) {\includegraphics[width=#1]{#2}};
	    \coordinate (viewport lower left) at (#3,#4/\imsizeratio);  
	    \coordinate(viewport upper right) at (#5,#6/\imsizeratio);  
        \draw[##5, line width = ##4 pt] (viewport lower left) rectangle (viewport upper right);
     \pgfmathsetmacro{\multone}{#5-#3}
     \pgfmathsetmacro{\multtwo}{#6/\imsizeratio-#4/\imsizeratio}
     
     \ifthenelse{\equal{#9}{bottom_left} }{ 
	      \node[anchor=north, draw= ##3, inner sep=0pt, line width = ##2 pt,outer sep=0pt] (zoomPart) at (\multone*#7/2+##2/345*1.333, \multtwo*#7+##2/345*1.333) {
	       \scalebox{#7}{\tikz{
	         \clip (#3,#4/\imsizeratio) rectangle (#5,#6/\imsizeratio);
	           
	         \node[anchor=south east, inner sep=0] at (1,0) {\includegraphics[width=#1]{#2}}; 
	         }}};
	   \ifthenelse{\equal{##1}{line_connection_on} }{ 
		  \draw[red, dashed] (viewport upper right|-viewport lower left) -- (zoomPart.north east); 
		  \draw[red, dashed] (viewport lower left) -- (zoomPart.north west);
		   }{}
	       
	 }{}

     \ifthenelse{\equal{#9}{bottom_right} }{ 
	      \node[anchor=north, draw= ##3, inner sep=0pt, line width = ##2 pt,outer sep=0pt] (zoomPart) at (1-\multone*#7/2-##2/345*1.333, \multtwo*#7 + ##2/345*1.333) {
	       \scalebox{#7}{\tikz{
			 \clip (#3,#4/\imsizeratio) rectangle (#5,#6/\imsizeratio);
	         \node[anchor=south east, inner sep=0] at (1,0) {\includegraphics[width=#1]{#2}}; 
	         }}%
	       };
		\ifthenelse{\equal{##1}{line_connection_on} }{ 
			  \draw[red, dashed] (viewport upper right|-viewport lower left) -- (zoomPart.south west); %red, dashed, ->
			  \draw[red, dashed] (viewport upper right) -- (zoomPart.north west);
		   }{}
     }{}  
       
    \ifthenelse{\equal{#9}{up_right} }{  %-0.015
	      \node[anchor=north, draw= ##3, inner sep=0pt, line width = ##2pt, outer sep=0pt] (zoomPart) at (1-\multone*#7/2-##2/345*1.333,1/\imsizeratio-##2/345*1.333) {
	       \scalebox{#7}{\tikz{
	          \clip (#3,#4/\imsizeratio) rectangle (#5,#6/\imsizeratio);
	          \node[anchor=south east, inner sep=0] at (1,0) {\includegraphics[width=#1]{#2}}; 
	         }}%
	         
	       };
	   \ifthenelse{\equal{##1}{line_connection_on} }{ 
		  \draw[red, dashed] (viewport lower left|-viewport upper right) -- (zoomPart.south west);
		  \draw[red, dashed] (viewport upper right) -- (zoomPart.south east);
		   }{}
     }{}

     \ifthenelse{\equal{#9}{up_left} }{ 
	      \node[anchor=north, draw= ##3, inner sep=0pt, line width = ##2pt,outer sep=0pt] (zoomPart) at (\multone*#7/2+##2/345*1.333, 1/\imsizeratio-##2/345*1.333) {
	       \scalebox{#7}{\tikz{
	         \clip (#3,#4/\imsizeratio) rectangle (#5,#6/\imsizeratio);
	           
	          \node[anchor=south east, inner sep=0] at (1,0) {\includegraphics[width=#1]{#2}}; 
	         }}};
		   \ifthenelse{\equal{##1}{line_connection_on} }{ 
			  \draw[red, dashed] (viewport lower left|-viewport upper right) -- (zoomPart.south west);
			  \draw[red, dashed] (viewport upper right) -- (zoomPart.south east);
			   }{}
	     }{}

  	\ifthenelse{\equal{#8}{help_grid_on} }{ 
           \begin{scope}[
                x={(image.south east)},
                y={(image.north west)},
                font=\footnotesize,
                help lines,
                overlay
            ]
            
            \draw[help lines, xstep=.1,ystep=.1,overlay] (0,0) grid (1,1);
            \foreach \x in {0,1,...,9} { 
                \node[anchor=north] at (\x/10,0) {0.\x}; %{\tiny0.\x}
            }
            \foreach \y in {0,1,...,9} {
                \node[anchor=east] at (0,\y/10) {0.\y};
            }
        \end{scope}    
	}{}  
   
\end{tikzpicture}

    }
    \ffoo
}